\documentclass{article}

\PassOptionsToPackage{numbers,compress}{natbib}
\usepackage[preprint]{neurips_2026}

\usepackage[utf8]{inputenc} 
\usepackage[T1]{fontenc}    
\usepackage{hyperref}       
\usepackage{url}            
\usepackage{booktabs}       
\usepackage{amsfonts}       
\usepackage{nicefrac}       
\usepackage{microtype}      
\usepackage{xcolor}         
\usepackage{amsmath}
\usepackage{amssymb}
\usepackage{graphicx}
\usepackage{algorithm}
\usepackage{algorithmic}
\usepackage{subcaption}
\usepackage{caption}
\usepackage{wrapfig}
\usepackage{float} 
\usepackage{enumitem}
\usepackage[table]{xcolor}
\usepackage{siunitx}
\usepackage{longtable}
\usepackage{booktabs}
\usepackage{tabularx}
\usepackage{ltablex}
\usepackage{threeparttable}
\keepXColumns
\definecolor{gmcolor}{RGB}{88,86,214}

\usepackage{titletoc}
\titlecontents{section}[1.5em]
    {\addvspace{6pt}\bfseries}
    {\contentslabel{1.5em}}
    {\hspace*{-1.5em}}
    {\titlerule*[0.5pc]{.}\contentspage}
\titlecontents{subsection}[3.8em]
    {}
    {\contentslabel{2.3em}}
    {\hspace*{-2.3em}}
    {\titlerule*[0.5pc]{.}\contentspage}
\hypersetup{linktoc=page}

\title{Beyond Patching: Consistent Circuit Discovery via Gradient Spectral Analysis}

\author{%
  Yujie Zhang$^{*}$ \quad Tianyi Zhao$^{*}$ \quad  Guanghui Min \quad Chen Chen\\
  Department of Computer Science\\
  University of Virginia\\
  Charlottesville, VA 22904 \\
  \texttt{\{qrk9jt, abs4dj, jjm8vr, zrh6du\}@virginia.edu} \\
  {\small $^{*}$Equal contribution}
  }
  
\begin{document}

\maketitle

\vspace{-5mm}
\begin{abstract}


Mechanistic interpretability seeks to reverse-engineer neural networks by identifying small, faithful subgraphs---circuits---that implement specific behaviors. The dominant paradigm scores each unit by replacing its clean activation with a corrupted counterpart as causal intervention.  While effective, this approach rests on a brittle prerequisite: the structure prompt pairs are task-specific, expensive to engineer, and silently couple
the discovered circuit to prompt-design choices. 
In this work, we propose \textbf{Spectral Circuit Analysis (SCA)}, a gradient-spectral pipeline that achieves \emph{consistent} circuit discovery along both axes: stable across input formulations of the same task, and natively exposing structure shared across tasks.
Our contributions are threefold. 
First, we provide a counterfactual-free reframing of circuit discovery: causally relevant directions are recoverable from the spectral structure of forward gradients across paraphrases, with leading directions concentrating task signal and trailing ones absorbing paraphrase noise—yielding circuits tied to the task rather than the prompt. Second, we introduce SCA, a block-wise SVD pipeline that extracts a task subspace per architectural block and redistributes spectral energy onto constituent units via a consistency-weighted share, producing per-unit rankings competitive with attribution-patching baselines under strictly less supervision. Finally, on twelve circuit-discovery tasks in GPT-2 Small and Pythia-160M, SCA performs comparably or exceeds EAP-family methods under the standard recovery-score protocol and recovers IOI canonical heads with $\mathrm{Precision}@K = 1.0$ in top-$10$ slices on GPT-2 Small;
the same machinery extends to a unified cross-task analysis classifying units as universal, family-shared, or task-specific and exposing compositional structure invisible to per-task analyses.
Our code and data are available at \url{https://github.com/YujieZhangzz/Spectral-Circuit-Analysis}.


\end{abstract}

\vspace{-5mm}
\section{Introduction}\label{sec:intro}

Mechanistic interpretability seeks to reverse-engineer neural networks by 
identifying small, faithful subgraphs—\emph{circuits}—that implement 
specific behaviors~\citep{olah2020zoom, elhage2021mathematical}. Case studies on indirect object 
identification~\citep{wang2022interpretability}, factual 
recall~\citep{meng2022locating, meng2022mass}, and in-context 
learning~\citep{olsson2022context} share a precise geometric premise: high-level task variables admit low-dimensional, often near-linear 
representations in transformer activation 
space~\citep{park2023linear, marks2023geometry, geiger2024finding, 
todd2023function}, with inter-layer communication itself routed through 
low-rank channels~\citep{merullo2024talking}. Circuit discovery, at its 
core, is the recovery of low-rank task-relevant structure.

The dominant operationalization casts this recovery as a problem of 
\emph{causal intervention}. Beginning with causal mediation 
analysis~\citep{vig2020investigating} and matured through automated 
discovery~\citep{conmy2023towards}, first-order 
approximations~\citep{syed2024attribution, hanna2024have, kramar2024atp}, 
and differentiable mask optimization~\citep{bhaskar2024finding}, this 
paradigm scores each unit by replacing its clean activation with a 
corrupted counterpart, the best-practice studies mapping the space~\citep{heimersheim2024use, zhang2023towards}. Across this lineage, 
however, the recovered circuit depends sensitively on the chosen contrast: 
corruption choices change recovered components in unpredictable 
ways~\citep{zhang2023towards}, circuits discovered on a fixed template do 
not always generalize across paraphrases~\citep{nainani2024adaptive}, and distinct counterfactual designs for 
the same task can induce systematically different 
circuits~\citep{haklay2025position, franco2026finding}. As task suites 
grow, this coupling between counterfactual design and recovered circuit 
motivates complementary methods that operate without contrast 
construction—pursued recently via information-flow 
attribution~\citep{ferrando2024information}, contextual 
decomposition~\citep{hsu2024efficient}, and sparse-feature 
circuits~\citep{marks2024sparse}. Independently, cross-task 
investigations~\citep{merullo2023circuit, gurnee2024universal} suggest that 
a substantial fraction of circuit machinery is shared across tasks, yet 
existing pipelines discover circuits one task at a time and only compare 
post hoc.


We pursue \emph{consistent} circuit discovery along both axes: a single 
procedure stable across input formulations of the same task and natively exposing structure shared across tasks. Our key observation 
is that \emph{the information identifying causally relevant units is 
already encoded in the spectrum of the forward task gradient}, recoverable 
from semantically related prompts alone. Two properties make this 
spectrum a natural target. First, task-relevant computation in 
transformers is intrinsically low-rank: rank-one MLP edits install factual associations~\citep{meng2022locating}, singular vectors 
of weight matrices align with token 
features~\citep{millidge2022singular}, and modest-dimensional per-head subspaces capture in-context arithmetic~\citep{hu2025understanding}. 
The forward gradient inherits this geometry—its dominant directions form a low-rank subspace not generally aligned with per-unit coordinate axes, calling for spectral rather than unit-wise decomposition.
Second, this subspace is invariant under semantic-preserving 
transformations: paraphrases of one prompt elicit gradients that 
share a task-specific subspace while differing only in surface-form noise. 
singular value decomposition (SVD) on stacked paraphrase gradients 
exploits this directly—leading directions concentrate consistent 
task signal, trailing ones absorb paraphrase-specific 
variation—yielding a circuit description tied to the task rather 
than to any particular prompt.


Building on this, we propose \textbf{Spectral Circuit Analysis (SCA)}, a 
gradient-spectral pipeline that applies SVD to per-paraphrase gradients 
\emph{block-wise} within each $(\text{layer}, \text{type})$ block and 
redistributes spectral energy to attention heads and MLP units via a 
consistency-weighted share. The block-wise design follows a recurring pattern in transformer analysis: low-rank structure aligns with 
the architecture's per-block decomposition. Parameter-efficient adaptation 
exploits this directly-LoRA injects independent low-rank adapters per 
layer~\citep{hu2022lora}—and circuit discovery reflects the same geometry, 
with IOI units grouped into contiguous attention 
layers~\citep{wang2022interpretability}, factual recall localized to 
mid-layer MLPs~\citep{meng2022locating, geva2023dissecting}, and induction 
heads concentrated in early-layer attention~\citep{olsson2022context}. 
Units within a block share residual-stream geometry and low-rank 
communication channels~\citep{merullo2024talking, elhage2021mathematical}, 
making a shared spectral basis the natural object of analysis; cross-block 
structure then emerges from unit-level subspace overlap, whose induced 
geometric similarity \emph{also} provides the cross-task signal: classifying 
units as universal, family-shared, or 
task-specific~\citep{chughtai2023toy} within the same machinery. 
On twelve circuit-discovery tasks in GPT-2 
Small~\citep{radford2019language} and 
Pythia-160M~\citep{biderman2023pythia}, SCA performs comparably or exceeds EAP-family 
baselines~\citep{syed2024attribution, hanna2024have} under the standard 
recovery-score protocol, and recovers IOI canonical heads with $\mathrm{Precision}@K = 1.0$ in top-$10$ slices on GPT-2 Small~\citep{wang2022interpretability}  ;
Cross-task analysis exposes a sharp layer-wise partition: 
early-layer attention is broadly universal~\citep{gurnee2024universal, 
merullo2023circuit}, while late layers host narrow family-level 
circuits. 
Our contributions are as follow:
\begin{itemize}[leftmargin=1em]
    \item \textbf{A counterfactual-free reframing of circuit discovery.}
    We show that causally relevant computational directions are 
    recoverable from the spectral structure of forward gradients across 
    paraphrased prompts, complementing the patching paradigm with an input-consistent route.
    
    \item \textbf{The SCA algorithm.} A block-wise SVD pipeline that 
    extracts a task subspace per architectural block and redistributes 
    spectral energy to constituent units via a consistency-weighted share, 
    yielding per-unit rankings competitive with state-of-the-art 
    attribution patching while consuming strictly less supervision.
    
    \item \textbf{A unified cross-task analysis framework.} The same 
    per-unit subspaces that drive single-task discovery directly support 
    cross-task comparison, classifying units into universal, family-shared, 
    and task-specific roles and uncovering compositional structure.
\end{itemize}

\vspace{-5mm}
\section{Preliminaries}
\label{sec:prelim}

\noindent\textbf{Setup.} We study circuit discovery in pre-trained
transformer language models. Let $f_\theta$ denote a transformer with
parameter vector $\theta \in \mathbb{R}^D$, where $D$ is the total
parameter count. A \emph{task}
$\mathcal{T}$ is specified by an input distribution
$x \sim \mathcal{D}_\mathcal{T}$ that elicits a target behavior,
together with a scalar metric $\mathcal{M}$ that maps the model's
next-token logits to $\mathbb{R}$. For example, the logit difference
between a correct completion and a contrastive completion of matched
syntactic form, as in the IOI
task~\citep{wang2022interpretability}. We adopt standard decoder-only
transformer notation~\citep{vaswani2017attention}: a model has $L$
layers operating on residual streams of dimension $d_{\mathrm{model}}$,
each comprising $H$ attention heads of head-dimension
$d_{\mathrm{head}}$ and a position-wise MLP block of hidden dimension
$d_{\mathrm{mlp}}$.

\subsection{Problem Formulation}
\label{sec:prelim-problem}

\noindent\textbf{Units.}  Let $[D]:=\{1, \ldots, D\}$ index the $D$
trainable parameters of $\theta$. We decompose $\theta$ into a set of
architecturally meaningful \emph{units}
$\mathcal{U} = \{u_1, \ldots, u_{|\mathcal{U}|}\}$, 
where each unit $u$ is identified with the index set $\mathcal{I}_u \subseteq [D]$ of the parameter coordinates it owns.
The $\{\mathcal{I}_u\}$ are pairwise disjoint, and
$\bigcup_u \mathcal{I}_u \subsetneq [D]$: embedding, layer-norm, and
unembedding parameters are treated as fixed and excluded.
Each attention head $(\ell, h)$ contributes \emph{one} unit,
consisting of the value and output projections
$W_V^{(\ell,h)} \in \mathbb{R}^{d_{\mathrm{model}} \times d_{\mathrm{head}}}$
and
$W_O^{(\ell,h)} \in \mathbb{R}^{d_{\mathrm{head}} \times d_{\mathrm{model}}}$,
which we treat as a single read--write operator of rank
$d_{\mathrm{head}}$ following \citet{elhage2021mathematical}. Each MLP
block at layer $\ell$ contributes units $c_{\mathrm{mlp}}$ ($c_{\mathrm{mlp}} = 2$ for the models studied in
Section~\ref{sec:experiments}), namely
$W_{\mathrm{up}}^{(\ell)} \in \mathbb{R}^{d_{\mathrm{model}} \times d_{\mathrm{mlp}}}$
and
$W_{\mathrm{down}}^{(\ell)} \in \mathbb{R}^{d_{\mathrm{mlp}} \times d_{\mathrm{model}}}$,
treated as separate units because the two matrices admit
independent low-rank structure under our analysis. In total,
$|\mathcal{U}| = L H + L\, c_{\mathrm{mlp}}$.
For task $\mathcal{T}$ with loss $\mathcal{L}_\mathcal{T}$ and an input
$x$, the parameter gradient is
\begin{equation}
g(x) \;:=\; \nabla_\theta\, \mathcal{L}_\mathcal{T}(x; \theta) \in \mathbb{R}^{D}.
\end{equation}

\noindent\textbf{Circuit discovery and the ranking protocol.} A circuit
discovery method $\mathcal{A}$ takes $f_\theta$ and $\mathcal{T}$ as
input and outputs a description of the units most responsible for
$\mathcal{T}$. The form of this output varies: greedy search produces
a discrete subset~\citep{conmy2023towards}, while attribution patching
produces per-unit scalar
scores~\citep{syed2024attribution,hanna2024have}. We assume
$\mathcal{A}$ assigns each unit a real-valued importance score
$I : \mathcal{U} \to \mathbb{R}$, inducing a ranking:
$u_{(1)}, u_{(2)}, \ldots, u_{(|\mathcal{U}|)}$
\text{with} $I(u_{(1)}) \geq I(u_{(2)}) \geq \cdots,$
where parenthesized subscripts denote rank order. A circuit at any
budget $K$ is read off as the top-$K$ prefix
$\mathcal{C}_K = \{u_{(1)}, \ldots, u_{(K)}\}$. The quality of
$\mathcal{A}$ is measured by how rapidly the metric
$\mathcal{M}(f_\theta \mid \mathcal{C}_K)$---evaluated with units in
$\mathcal{C}_K$ replaced by their corrupted activations while the
remaining units retain their clean activations---departs from the
clean metric $\mathcal{M}(f_\theta)$ as $K$ grows.
We formalize this as the \emph{recovery score}
$r_\mathcal{A}(K)$ in Section~\ref{sec:experiments}.





\begin{figure*}[t]
    \centering
    \includegraphics[width=1\textwidth]{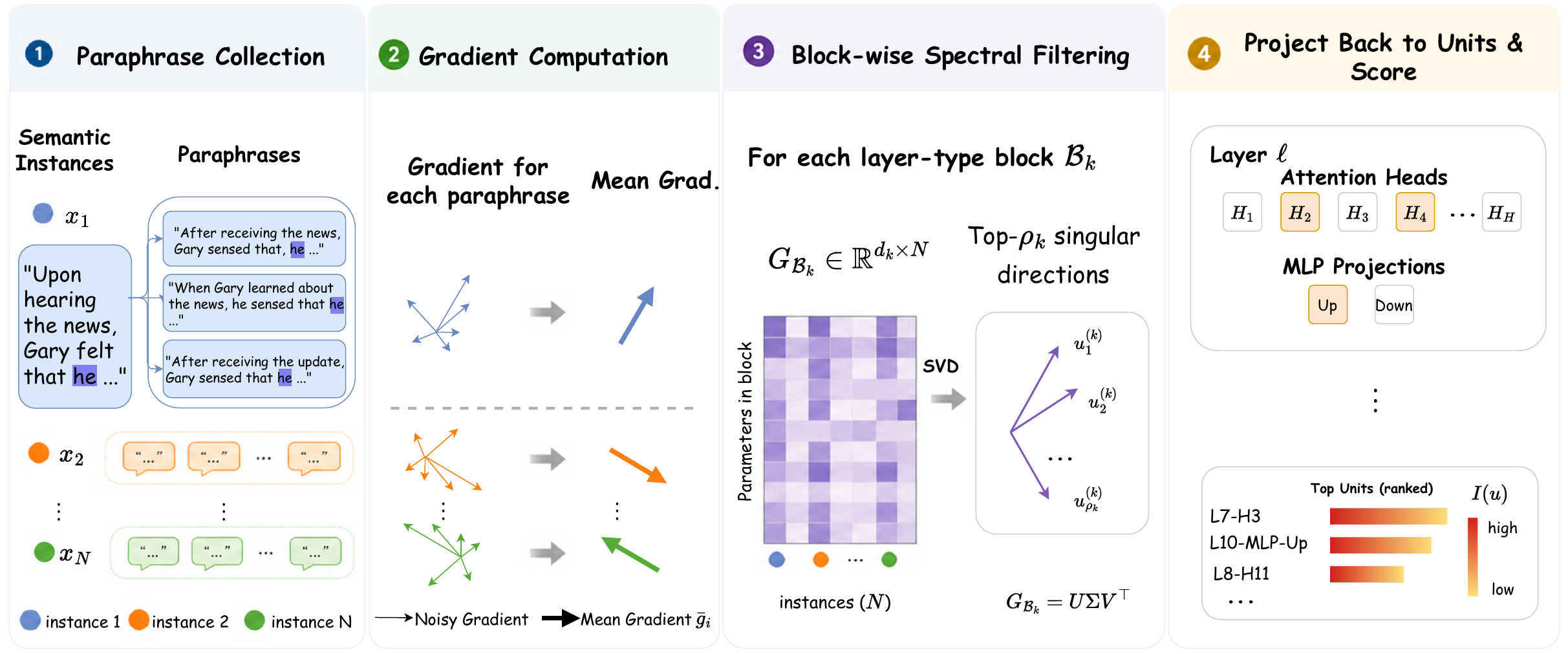}
    \vspace{-5mm}
    \caption{
    Overview of \textbf{Spectral Circuit Analysis} (SCA).
    SCA converts semantic-preserving paraphrases into instance-level mean
    gradients, extracts shared low-rank gradient subspaces within layer-type
    blocks via block-wise SVD, and projects the resulting spectral energy back
    to attention heads and MLP projections to produce a ranked list of circuit
    candidates.
    }
    \label{fig:sca_pipeline}
    \vspace{-5mm}
\end{figure*}




\section{SCA: Our Proposed Method}
\label{sec:method}

\subsection{Single-Task Circuit Discovery}
\label{sec:sca_method}

We propose \emph{Spectral Circuit Analysis} (SCA), a gradient-spectral
pipeline that ranks units by their participation in a
task's shared computation. 
Our key observation is that while single input gradients are
noisy, gradients of \emph{semantically equivalent} paraphrases share a common direction in
parameter space that reflects the task itself; across units that read from and write to the same residual stream,
this direction generalizes to a low-dimensional \emph{semantic
subspace} on which their gradients jointly lie. 
Building on this, we propose a consistent three-stage pipeline that recovers the
semantic subspace from paraphrase-averaged gradients via a per-block
SVD and redistributes its spectral content onto units to score them,
capturing collaborative computation that per-unit attribution discards.

\begin{wrapfigure}[26]{r}{0.5\textwidth}   
\begin{minipage}{0.5\textwidth}
\footnotesize
\hrule height 0.8pt
\vspace{3pt}
\refstepcounter{algorithm}
\noindent\textbf{Algorithm \thealgorithm:} SCA: Single-Task Per-Unit Importance Ranking%
\label{alg:sca}
\vspace{2pt}\hrule\vspace{2pt}
\begin{algorithmic}[1]
\REQUIRE model $f_\theta$; task $\mathcal{T}$ with $N$ semantic instances $\{x_i\}_{i=1}^N$, each with $M$ paraphrases $\{x_{i,j}\}_{j=1}^M$; unit set $\mathcal{U}$; blocks $\{\mathcal{B}_k\}_{k=1}^{B}$ from Eq.~\eqref{eq:block-rule}; energy threshold $\eta$=0.95
\ENSURE per-unit importance $I:\mathcal{U}\!\to\!\mathbb{R}$; block subspaces $\{U_{\mathcal{B}_k}\}_{k=1}^{B}$
\STATE \textit{// Stage 1: paraphrase-averaged gradients}
\FOR{$i = 1, \ldots, N$}
  \STATE $\bar g_i \leftarrow \tfrac{1}{M}\sum_{j=1}^{M}\nabla_\theta\mathcal{L}_\mathcal{T}(x_{i,j};\theta)$
\ENDFOR
\STATE $\mathbf{G}\leftarrow [\bar g_1,\ldots,\bar g_N]^{\!\top}$
\STATE \textit{// Stage 2: block-wise SVD}
\FOR{$k = 1, \ldots, B$}
  \STATE $G_{\mathcal{B}_k} \leftarrow \mathbf{G}[:,\mathcal{I}_{\mathcal{B}_k}]^{\!\top}$
  \STATE $\{(\sigma_j^{(k)}, u_j^{(k)})\} \leftarrow \textsc{SVD}(G_{\mathcal{B}_k})$ Gram Trick
  \STATE $\rho_k \leftarrow \min\{r : \tfrac{\sum_{j\leq r}(\sigma_j^{(k)})^2}{\sum_{j}(\sigma_j^{(k)})^2} \geq \eta\}$
  \STATE $U_{\mathcal{B}_k} \leftarrow [u_1^{(k)},\ldots,u_{\rho_k}^{(k)}]$
\ENDFOR
\STATE \textit{// Stage 3: two-factor importance}
\FOR{$k = 1, \ldots, B$}
  \STATE $\mathrm{cons}(\mathcal{B}_k) \leftarrow \|\tfrac{1}{N}\sum_i\mathrm{normalize}(\mathbf{G}[i,\mathcal{I}_{\mathcal{B}_k}])\|_2$
  \FOR{$u \in \mathcal{B}_k$}
    \STATE $\mathrm{unit\_share}(u) \leftarrow \tfrac{1}{d_u}\|U_{\mathcal{B}_k}[\mathrm{idx}_u^{\mathrm{loc}},:]\|_F^2$
    \STATE $I(u) \leftarrow \mathrm{cons}(\mathcal{B}_k)\cdot\mathrm{unit\_share}(u)$
  \ENDFOR
\ENDFOR
\RETURN $I,\ \{U_{\mathcal{B}_k}\}_{k=1}^{B}$
\end{algorithmic}
\vspace{1pt}\hrule height 0.8pt
\end{minipage}
\end{wrapfigure}

\paragraph{Setup.}
We instantiate $\mathcal{D}_\mathcal{T}$ from
Section~\ref{sec:prelim-problem} as $N$ \emph{semantic instances}
$\{x_i\}_{i=1}^{N}$, each with $M$ \emph{semantic-preserving
paraphrases} $\{x_{i,j}\}_{j=1}^{M}$. The instance gradient is the
paraphrase average
\begin{equation}
\bar g_i \;:=\; \frac{1}{M}\sum_{j=1}^{M} \nabla_\theta\mathcal{L}_\mathcal{T}(x_{i,j};\theta) \;\in\; \mathbb{R}^D.
\label{eq:method-grad}
\end{equation}
We group units into \emph{blocks} by layer and type. Letting
$\ell : \mathcal{U} \to [L]$ and
$\tau : \mathcal{U} \to \{\mathrm{attn}, \mathrm{mlp}\}$ denote the
layer and type of a unit,
\begin{equation}
\mathcal{B}_{\ell, \tau} \;:=\; \bigl\{ u \in \mathcal{U} \;:\; \ell(u) = \ell, \;\, \tau(u) = \tau \bigr\}
\label{eq:block-rule}
\end{equation}
yields $B := 2L$ blocks, with $k(u) \in [B]$ indexing the block of $u$. 
Blocks are
the natural locus for a shared task subspace: units within
one read from and write to the same residual stream, operating
collaboratively rather than independently~\citep{elhage2021mathematical, wang2022interpretability}.

\paragraph{The SCA pipeline.}
SCA (Algorithm~\ref{alg:sca})
follows three stages:
gradient collection, block-wise spectral decomposition, and two-factor
unit scoring. 
We first compute paraphrase-averaged instance gradients
$\bar g_i$ via Eq.~\eqref{eq:method-grad} and stack them into the
gradient matrix $\mathbf{G}$ (steps 2--5). 
Averaging preserves the task-invariant unit while suppressing input-specific noise at rate $O(M^{-1/2})$.
%
Step 7--12 extract each block's semantic subspace.
Since $N \ll d_k$, step 9 uses the
\emph{Gram trick}: eigendecompose
$G_{\mathcal{B}_k}^{\top} G_{\mathcal{B}_k}$ ($N \times N$) for
right singular vectors and squared singular values, then recover left
singular vectors by one matrix multiplication, reducing per-block cost
from $O(d_k N \min(d_k, N))$ to $O(d_k N^2)$. 
Step 10 truncates at the smallest rank $\rho_k$ satisfying
\begin{equation}
\frac{\sum_{j \leq \rho_k} (\sigma_j^{(k)})^2}{\sum_{j \leq N} (\sigma_j^{(k)})^2} \;\geq\; \eta, \qquad \eta = 0.95,
\label{eq:rho}
\end{equation}
retaining directions on which instances jointly vary.
Since the column span of
$G_{\mathcal{B}_k}$ is at most $N$-dimensional, the task signal
concentrates on these few directions; orthogonal ones contribute only
noise.
Step 14--19 score each unit by a product of two complementary factors.
$\mathrm{cons}(\mathcal{B}_k)$ (step 15) is the mean resultant
length~\citep{mardia2009directional} of the block's row-normalized
per-instance gradients on the unit sphere, answering: \emph{is the block
doing coherent task-relevant work?}
$\mathrm{unit\_share}(u)$ (step 17) is
the row energy of $u$'s parameter slice within $U_{\mathcal{B}_k}$,
normalized by parameter count, answering: \emph{is the unit carrying the
work within its block?} 
Since $U_{\mathcal{B}_k}$ has
orthonormal columns, the total energy
$\|U_{\mathcal{B}_k}\|_F^2 = \rho_k$ splits among units in
proportion to how strongly each participates
in the block's leading singular directions;
per-parameter normalization prevents larger units (e.g., $W_{\mathrm{up}}$) from inheriting score on size alone. 
The product
$I(u) = \mathrm{cons}(\mathcal{B}_{k(u)}) \cdot \mathrm{unit\_share}(u)$
(step 18) ranks a unit high only when both conditions hold: the block is active \emph{and} the unit carries its spectral mass. 
Sorting by $I(u)$ yields the ranking
$u_{(1)}, u_{(2)}, \ldots$ that defines the circuit
$\mathcal{C}_K = \{u_{(1)}, \ldots, u_{(K)}\}$ at any budget $K$.

\subsection{Cross-Task Generalization Analysis}
\label{sec:cross_task}

The pipeline of Section~\ref{sec:sca_method} produces, for one task, a
per-unit ranking $I(u)$ and block subspaces
$\{U_{\mathcal{B}_k}\}$. A natural question is whether the units
SCA identifies serve \emph{universal} computation, align with coherent \emph{task families}, or are \emph{task-specific}. 
Our key observation: the
a block's semantic subspace is a geometric signature of its computation; if two tasks recruit the same unit for the same operation, that unit's per-task subspaces should
\emph{geometrically agree} regardless of surface form. Building on this, we extend SCA jointly over $T$ tasks: 
project each block's semantic subspace to per-unit subspaces, measure
pairwise geometric agreement against a structural null, weight by task-side importance, and classify each unit
through priority-ordered gating tests. Quantities from
Section~\ref{sec:sca_method} acquire a task superscript when computed
under task $t$ (e.g.\ $U_{\mathcal{B}_k}^t$, $\rho_k^t$, $I^t(u)$).

\begin{wrapfigure}[44]{r}{0.5\textwidth}
\vspace{-1.5em}
\begin{minipage}{0.5\textwidth}
\footnotesize
\raggedright
\hrule height 0.8pt
\vspace{2pt}
\refstepcounter{algorithm}
\noindent\textbf{Algorithm \thealgorithm:} Cross-Task Family-Membership Classification%
\label{alg:sca-cross}
\vspace{1pt}\hrule\vspace{2pt}
\begin{algorithmic}[1]
\REQUIRE tasks $\{\mathcal{T}^t\}_{t=1}^{T}$ with single-task outputs $\{I^t,\ \{U_{\mathcal{B}_k}^t\}\}$ from Algorithm~\ref{alg:sca}; family layout $\mathcal{F}=\{F_1,\ldots,F_m\}$; thresholds $(\theta_{\mathrm{univ}}, \theta_F^{\mathrm{floor}}, \theta_{\mathrm{dom}}, \theta_{\mathrm{out}}, \delta)$; null sample size $R_{\mathrm{null}}$; energy threshold $\eta$
\ENSURE per-unit class $C:\mathcal{U}\to\{\textsc{Universal},\textsc{Family-}F,\textsc{Task-}t^*,\textsc{Mixed}\}$
\STATE \textit{// Stage 1: per-unit semantic subspaces}
\FOR{$u \in \mathcal{U}$, $t \in [T]$}
  \STATE $M_u^t \leftarrow U_{\mathcal{B}_{k(u)}}^t[\mathrm{idx}_u^{\mathrm{loc}},:] \in \mathbb{R}^{d_u\times \rho_{k(u)}^t}$
  \STATE $\{(\sigma_j, v_j)\} \leftarrow \textsc{CompactSVD}(M_u^t)$;\quad $\rho_u^t \leftarrow$ truncate at energy $\eta$
  \STATE $U_u^t \leftarrow [v_1,\ldots,v_{\rho_u^t}] \in \mathbb{R}^{d_u\times \rho_u^t}$
\ENDFOR
\STATE \textit{// Stage 2: importance-weighted, null-calibrated overlap}
\FOR{$u \in \mathcal{U}$, unordered pair $\{s,t\}\subseteq[T]$}
  \STATE $\mathrm{overlap}(u,s,t) \leftarrow \|(U_u^s)^{\!\top} U_u^t\|_F^2 / \min(\rho_u^s,\rho_u^t)$ \hfill$\triangleright$ Eq.~\eqref{eq:overlap}
  \STATE draw $R_{\mathrm{null}}$ random rank-matched orthonormal pairs in $\mathbb{R}^{d_u}$; flag significant if above $95^{\mathrm{th}}$ percentile
  \STATE $\widehat I^{\,r}(u) \leftarrow I^r(u)/\max_{u':\tau(u')=\tau(u)} I^r(u')$ for $r\in\{s,t\}$
  \STATE $w(u,s,t) \leftarrow \sqrt{\widehat I^s(u)\widehat I^t(u)}\cdot \mathrm{overlap}(u,s,t)$ \hfill$\triangleright$ Eq.~\eqref{eq:weighted}
\ENDFOR
\STATE \textit{// Stage 3: priority-ordered family-membership gating}
\FOR{$u \in \mathcal{U}$}
  \STATE $\mathrm{score}(u) \leftarrow \mathrm{agg}_{\{s,t\}\subseteq[T]} w$
  \FOR{$F \in \mathcal{F}$}
    \STATE $\mathrm{within}_F(u) \leftarrow \mathrm{agg}_{\{s,t\}\subseteq F} w$;\quad $\mathrm{cross}_F^{\max}(u) \leftarrow \max_{F'\neq F}\mathrm{within}_{F'}(u)$
  \ENDFOR
  \IF{$\mathrm{score}(u) \geq \theta_{\mathrm{univ}}$}
    \STATE $C(u) \leftarrow \textsc{Universal}$ \hfill\textit{// G1}
  \ELSIF{$\exists F:\ \mathrm{within}_F(u) \geq \theta_F^{\mathrm{floor}}\ \textbf{and}\ \mathrm{within}_F(u) - \mathrm{cross}_F^{\max}(u) \geq \delta$}
    \STATE $C(u) \leftarrow \textsc{Family-}F$ \hfill\textit{// G2}
  \ELSE
    \STATE $t^* \leftarrow \arg\max_t \widehat I^{\,t}(u)$
    \IF{$\widehat I^{\,t^*}(u)/\max_{t\neq t^*}\widehat I^{\,t}(u) \geq \theta_{\mathrm{dom}}\ \textbf{and}\ \max_{t\neq t^*} w(u,t^*,t) \leq \theta_{\mathrm{out}}$}
      \STATE $C(u) \leftarrow \textsc{Task-}t^*$ \hfill\textit{// G3}
    \ELSE
      \STATE $C(u) \leftarrow \textsc{Mixed}$
    \ENDIF
  \ENDIF
\ENDFOR
\RETURN $C$
\end{algorithmic}
\vspace{1pt}\hrule height 0.8pt
\end{minipage}
\end{wrapfigure}

\paragraph{The cross-task SCA pipeline.}
\label{sec:method-cross-task}
The cross-task analysis (Algorithm~\ref{alg:sca-cross}) has three stages: per-unit subspace extraction, importance-weighted overlap,
and family-membership gating. 
Step 2--6 project each block's semantic subspace to its constituent units. 
The row slice $M_u^t = U_{\mathcal{B}_{k(u)}}^t[\mathrm{idx}_u^{\mathrm{loc}},:]$
inherits the block's leading directions but is no longer
column-orthonormal; we re-orthonormalize via a secondary
compact SVD and truncate at the same threshold $\eta$ from
Eq.~\eqref{eq:rho}, yielding the \emph{per-unit semantic subspace}
$U_u^t \in \mathbb{R}^{d_u \times \rho_u^t}$. 
This allows each unit to retain only the block directions it actually participates in, so generally $\rho_u^t \leq \rho_{k(u)}^t$.
Step 8--13 measure cross-task agreement and weight it by importance. The geometric agreement (step 9) is the
normalized projection energy
\begin{equation}
\mathrm{overlap}(u, s, t) \;:=\;
\frac{\bigl\| (U_u^s)^{\!\top} U_u^t \bigr\|_F^2}{\min(\rho_u^s,\, \rho_u^t)}
\;\in\; [0, 1],
\label{eq:overlap}
\end{equation}
which equals $1$ iff one subspace contains the other. 
Since $\rho_u^s$ and $\rho_u^t$ vary across units and tasks, raw
overlap values are not directly comparable: a high overlap between
two rank-$1$ subspaces in $\mathbb{R}^{d_u}$ is geometrically
unsurprising, while the same value between two rank-$10$ subspaces is
not. Step 10 calibrates against this baseline using
$R_{\mathrm{null}}$ random orthonormal pairs of matched ranks
$(\rho_u^s, \rho_u^t)$ in $\mathbb{R}^{d_u}$, flagging
$\mathrm{overlap}(u,s,t)$ as significant when it exceeds the
$95^{\mathrm{th}}$ null percentile. Step 11 type-stratifies the
single-task scores so attention and MLP units are placed on a common scale, and
step 12 combines geometry and importance into the
\emph{importance-weighted overlap}
\begin{equation}
w(u, s, t) \;:=\; \sqrt{\widehat I^{\,s}(u)\,\widehat I^{\,t}(u)}\;\cdot\; \mathrm{overlap}(u, s, t).
\label{eq:weighted}
\end{equation}
The geometric mean ensures both tasks score the unit
non-trivially: a unit important on one task but irrelevant on
the other receives near-zero weight regardless of
subspace alignment, suppressing spurious overlaps that pure
geometric measures admit.
Step 14--26 classify each unit through three priority-ordered gates. 
We first aggregate $w(u,s,t)$ globally and
within each candidate family $F$ to obtain $\mathrm{score}(u)$ and
$\mathrm{within}_F(u)$ (steps 15--18); the family layout
$\mathcal{F}$ is supplied as input and encodes which task subsets are
hypothesized to share computation. 
The gates are then applied as follows. 
\textbf{G1} (step 19, \textsc{Universal}): a unit passes if
$\mathrm{score}(u)$ exceeds $\theta_{\mathrm{univ}}$,
indicating consistently strong agreement across all task pairs.
\textbf{G2} (step 21, \textsc{Family-}$F$): a unit passes if some
family $F$ exceeds the within-family floor $\theta_F^{\mathrm{floor}}$
\emph{and} beats every other family by a margin of at least $\delta$, preventing broadly active units from satisfying multiple families. 
\textbf{G3} (step 24, \textsc{Task-}$t^*$ where $t^* := \arg\max_t \widehat
I^{\,t}(u)$): a unit passes if its top task dominates the runner-up
by ratio $\theta_{\mathrm{dom}}$ \emph{and} its weighted overlap with
all other tasks is bounded by $\theta_{\mathrm{out}}$, ensuring the unit is task-selective without coincidental alignment.
Units passing none of G1--G3 are labeled \textsc{Mixed}. 
The five thresholds are 
fixed default whose robustness is verified in Appendix~\ref{app:threshold};
full
specification, family-layout robustness, and a specialized/broad split within \textsc{Family-}$F$ are deferred to
Appendix~\ref{app:family_details}.


\section{Experiments}
\label{sec:experiments}

\subsection{Experimental Setup}
\label{sec:faithfulness}

\noindent\textbf{Models and tasks.} We evaluate on two language models with
identical macro-architecture ($L=12$, $d_{\mathrm{model}}{=}768$,
$H=12$, $c_{\mathrm{mlp}}=2$, $|\mathcal{U}| = LH + L c_{\mathrm{mlp}} = 168$):
\textbf{GPT-2 Small} (124M;~\citealp{radford2019language}), the
canonical testbed for unit-level circuit discovery, and
\textbf{Pythia-160M} (162M;~\citealp{biderman2023pythia}). Pythia-160M
results are deferred to Appendix~\ref{app:pythia}.
We use twelve circuit-discovery tasks: Indirect Object Identification (IOI),
Greater-Than, Greater-Than-Price, Country-Capital, Person-Country,
Person-Profession, Colored Objects, Gender Bias, Gendered Pronoun,
Subject-Verb Agreement (SVA), SVA-Binary, and SVA-BLiMP. Per-task templates,
paraphrase generation ($M{=}5$ per, Appendix~\ref{app:m}), and the
logit-difference / probability-difference loss $\mathcal{L}_\mathcal{T}$
used for each task are specified in Appendix~\ref{app:tasks}.
We assemble $N{=}100$ semantic instances, each with $M=5$ paraphrases per task.



\noindent\textbf{Evaluation Metric.} We follow the standard clean/corrupted recovery-score protocol~\citep{wang2022interpretability,conmy2023towards}.
Each task $\mathcal{T}$ comes with paired distributions $\mathcal{D}_\mathcal{T}^{\mathrm{clean}}$  and
$\mathcal{D}_\mathcal{T}^{\mathrm{corr}}$ that contrast the target
behavior. For a unit set $\mathcal{C} \subseteq \mathcal{U}$ and
paired sample $(x, x')$, denoising patching restores units in $\mathcal{C}$
to their clean activations on $x$ during the corrupted forward pass on $x'$, and we write 
\begin{equation}
\mathcal{M}(f_\theta \mid \mathcal{C}) \;:=\; \mathbb{E}_{(x,x')}\bigl[\mathcal{M}\bigl(f_\theta\bigl(x' \mid \mathrm{patch}_\mathcal{C}(x' \leftarrow x)\bigr)\bigr)\bigr],
\end{equation}
so that $\mathcal{M}(f_\theta \mid \emptyset)$ is the corrupted
baseline and $\mathcal{M}(f_\theta \mid \mathcal{U})$ recovers the
clean metric $\mathcal{M}(f_\theta)$. Restricting attention to heads,
let
$\mathcal{U}_{\mathrm{attn}} := \{u \in \mathcal{U} : \tau(u) = \mathrm{attn}\}$
and let $\mathcal{C}_K^\mathcal{A} \subseteq \mathcal{U}_{\mathrm{attn}}$
be the top-$K$ prefix of method $\mathcal{A}$'s ranking. The \emph{recovery score} is 
\begin{equation}
r_\mathcal{A}(K) \;:=\;
\frac{\mathcal{M}(f_\theta \mid \mathcal{C}_K^\mathcal{A}) - \mathcal{M}_{(f_\theta \mid \mathcal{U}_{\mathrm{attn}})}}{\mathcal{M}_{\mathcal{M}(f_\theta \mid \emptyset)} - \mathcal{M}_{(f_\theta \mid \mathcal{U}_{\mathrm{attn}})}},
\label{eq:recovery}
\end{equation}
with $r_\mathcal{A}(0) = 1$ and
$r_\mathcal{A}(|\mathcal{U}_{\mathrm{attn}}|) = 0$ by construction.
A ranking that concentrates task-relevant heads at low $K$ drives
$r_\mathcal{A}(K)$ rapidly toward zero; uniform-random rankings
descend approximately linearly. \textbf{Lower $r_\mathcal{A}(K)$
indicates a better ranking.} On IOI we additionally report
$\mathrm{Recall}@K$ and Average Precision against the 26-head
canonical circuit of \citet{wang2022interpretability}. SCA recover $10/26$ canonical heads in their top-$10$ slices, with $\mathrm{Precision}@K = 1.0$ for the leading positions, details deferred to Appendix~\ref{app:ioi-case-study}).


\subsection{Baselines}
\label{sec:baselines}

We compare SCA against six per-unit ranking baselines under a unified
cumulative-patching protocol. The \textbf{Random} baseline samples a uniform permutation of $\mathcal{U}$ as the ranking, with results
averaged over five random seeds. (i) Two raw-gradient baselines operate on the same per-paraphrase
gradients $g_{i,j}$ that SCA consumes and reduce them to a
per-component scalar via either gradient magnitude
$\|\bar g_i[\mathcal{I}_u]\|_2^2$ averaged over instances
(\textbf{Magnitude}) or directional coherence
$\|\sum_i \bar g_i[\mathcal{I}_u] / \|\bar g_i[\mathcal{I}_u]\|\|_2 / N$
(\textbf{Direction}); they isolate the
two factors SCA's spectral filtering jointly accounts for. (ii) The EAP-family baselines: \textbf{EAP}~\citep{syed2024attribution}, the original first-order formulation; \textbf{EAP-IG}~\citep{hanna2024have}, which integrates
the gradient over a clean$\to$corrupted input interpolation to
mitigate gradient saturation; and \textbf{EAP-IG-KL}, an EAP-IG variant
that replaces the task metric $\mathcal{M}$ with the KL divergence
between clean and patched output distributions. More explanation of EAP family is in Appendix~\ref{app:eap} All seven methods (SCA + six
baselines) are evaluated under the protocol of
Eq.~\eqref{eq:recovery}.

\subsection{Main Results}
\label{sec:results}

\subsubsection{Single-Task Circuit Discovery}
\label{sec:sca_results}

\begin{figure}[t]
    \centering
    \begin{subfigure}[t]{0.32\textwidth}
        \centering
        \includegraphics[width=\linewidth]{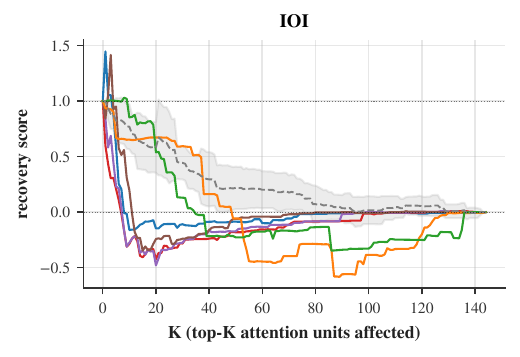}
        \caption{}
        \label{fig:faith-gpt2-ioi}
    \end{subfigure}
    \begin{subfigure}[t]{0.32\textwidth}
        \centering
        \includegraphics[width=\linewidth]{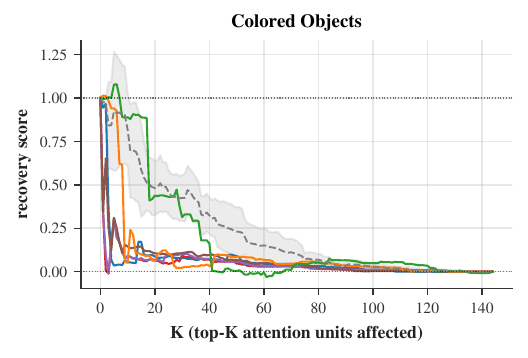}
        \caption{}
        \label{fig:faith-gpt2-co}
    \end{subfigure}
    \begin{subfigure}[t]{0.32\textwidth}
        \centering
        \includegraphics[width=\linewidth]{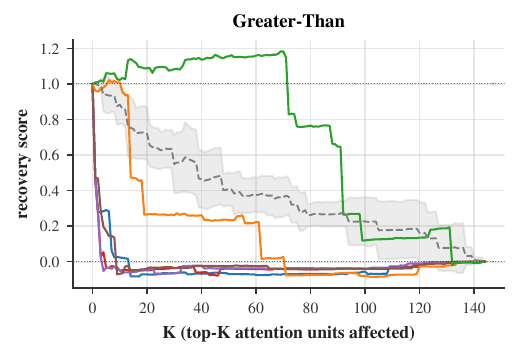}
        \caption{}
        \label{fig:faith-gpt2-gt}
    \end{subfigure}
    \begin{subfigure}[t]{0.32\textwidth}
        \centering
        \includegraphics[width=\linewidth]{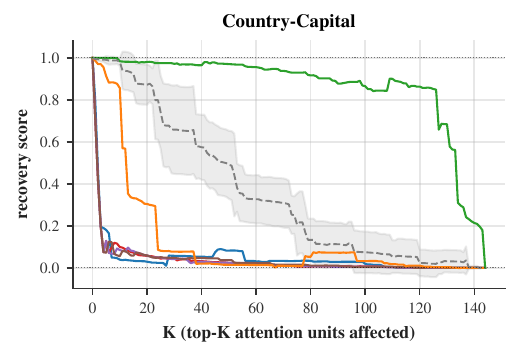}
        \caption{}
        \label{fig:faith-gpt2-cc}
    \end{subfigure}
    \begin{subfigure}[t]{0.32\textwidth}
        \centering
        \includegraphics[width=\linewidth]{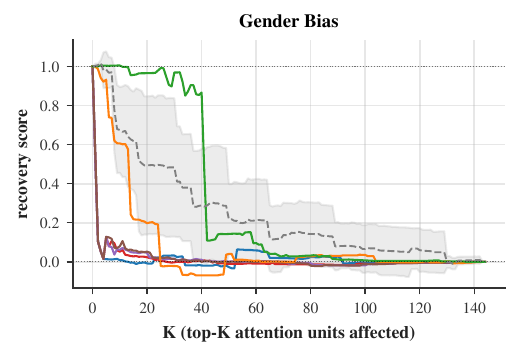}
        \caption{}
        \label{fig:faith-gpt2-gb}
    \end{subfigure}
    \begin{subfigure}[t]{0.32\textwidth}
        \centering
        \includegraphics[width=\linewidth]{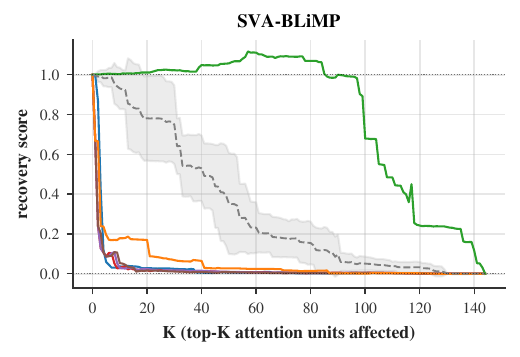}
        \caption{}
        \label{fig:faith-gpt2-sva-blimp}
    \end{subfigure}
    \begin{subfigure}[t]{0.95\textwidth}
        \centering
        \includegraphics[width=\linewidth]{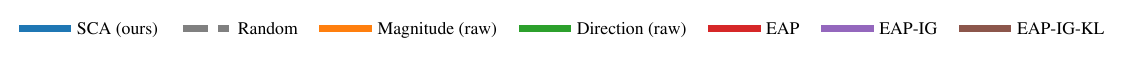}
    \end{subfigure}
    \vspace{-2mm}
    \caption{\textbf{Recovery score $r_\mathcal{A}(K)$ on six 
GPT-2 Small circuit-discovery tasks.} For each method $\mathcal{A}$, 
attention heads are ranked by importance and the top-$K$ are patched 
with corrupted activations; $r_\mathcal{A}(K)$ measures the residual 
clean--floor gap (Eq.~\ref{eq:recovery}). Lower is better. Gray 
shading: Random baseline mean $\pm 1\sigma$ over 5 seeds.}
    \label{fig:faithfulness-gpt2-6}
\end{figure}

Figure~\ref{fig:faithfulness-gpt2-6} reports the recovery score
$r_\mathcal{A}(K)$ for SCA and six baselines on six tasks: IOI,
Colored Objects, Greater-Than, Country-Capital, SVA-BLiMP, and
Gender Bias. Lower curves indicate rankings that concentrate
causally relevant heads at low $K$. Results on the remaining tasks
are deferred to Appendix~\ref{app:faithful-gpt2-6}.

\noindent\textbf{SCA matches the EAP family without counterfactual prompts.}
On all six tasks, SCA's curve tracks the EAP variants closely or
lies below them across $K \in [1, 144]$. The match is most
pronounced at small budgets, where circuit-level interpretation is
most informative: on Colored Objects, Country-Capital, Gender Bias,
and SVA-BLiMP (Figures~\ref{fig:faith-gpt2-co},
\ref{fig:faith-gpt2-cc}, \ref{fig:faith-gpt2-gb},
\ref{fig:faith-gpt2-sva-blimp}), SCA falls below all three EAP
variants within the top-$10$ budget. SCA achieves this from
forward-task gradients alone, whereas EAP variants require
matched clean--corrupted prompt
pairs~\citep{syed2024attribution,hanna2024have}. 
On the 26-head canonical IOI circuit of
\citet{wang2022interpretability}, SCA matches EAP-IG at the top of
the ranking---both recover 10/26 heads in their top-10 slices with
$\mathrm{Precision}@K \approx 1.0$ at leading positions. A per-category
breakdown and full retrieval analysis are in
Appendix~\ref{app:ioi-case-study}.

\noindent\textbf{Subspace geometry, not gradient magnitude or direction,
drives SCA.}
The Magnitude and Direction baselines consume the same per-paraphrase
gradients as SCA but reduce them to first-moment statistics. Both
underperform SCA and the EAP family across all six tasks: Direction
sits at or above the Random band over most of the budget range on
Country-Capital, Greater-Than, Gender Bias, and SVA-BLiMP, indicating
that per-instance directional coherence is uninformative or
anti-correlated with causal relevance; Magnitude tracks Random
closely but never matches SCA on any task. Replacing this
first-moment reduction with block-wise SVD on the same gradient input
recovers a ranking comparable to EAP-family attribution---isolating
\emph{cross-instance subspace geometry} as the operative signal
underlying SCA.

\subsection{Cross-task }
\label{sec:cross-task}

\subsection{Cross-Task Analysis: Family Structure and Unit-Level Specialization}
\label{sec:cross-task}

\begin{figure*}[t]
\label{fig:gpt-fam}
    \centering
    \includegraphics[width=0.97\textwidth]{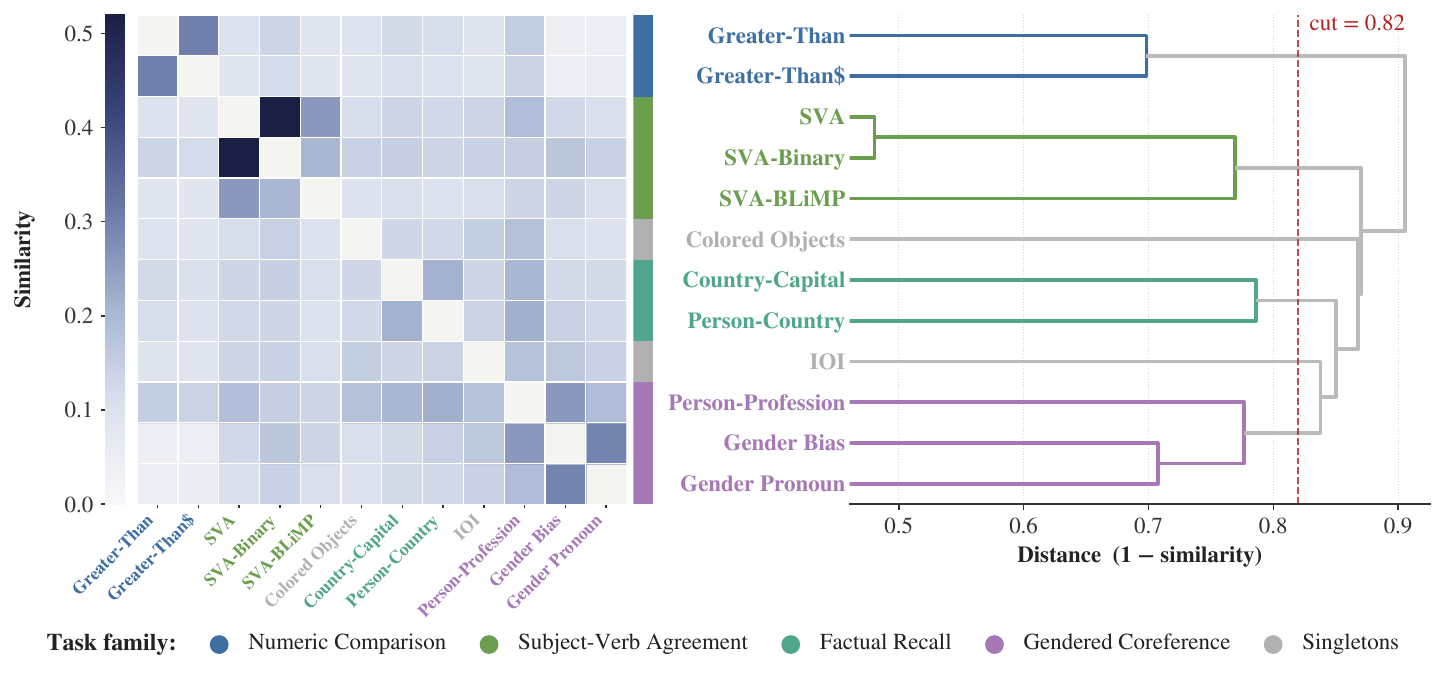}
    \vspace{-3mm}
    \caption{\textbf{Cross-task similarity on 12 GPT-2 tasks.} Heatmap (Left): pairwise similarity, diagonal masked; vertical band marks family. Dendrogram (Right): average-linkage on $d = 1 - \text{similarity}$, branches colored by family when all leaves agree.}\label{fig:gpt-fam}
\end{figure*}

\noindent\textbf{From task similarity to a family-level prior.}
We first establish a task-level family structure that will serve as the
backbone for unit-level analysis. The spectral similarity matrix over the
12 tasks (in Figure~\ref{fig:gpt-fam}) admits a clean hierarchical decomposition, yielding four families and two singletons:
\textbf{Numeric Comparison} (GT, GT\}),
\textbf{Subject--Verb Agreement} (SVA, SVA-Bi, SVA-BLi),
\textbf{Factual Recall} (CC, PC), and
\textbf{Gendered Coreference} (PP, GB, GP),
with CO and IOI remaining as singletons.
The two surface forms of Greater-Than collapse onto a single near-diagonal
block, confirming that they share the underlying computation; the SVA
variants likewise form a tight block, with SVA/SVA-Bi
the most internally similar pair. 

\begin{table*}[t]
\centering
\caption{Cross-task subspace lift across candidate component sets.
Each row is a candidate set of GPT-2 Small units (attention heads + MLPs);
each column is one of 12 evaluation tasks. Values are mean lift in target-token
logit difference when the residual stream is restricted to the candidate's
weighted unit-SVD subspace. $|S|$ is the number of components in the set.
\textbf{Bold} indicates positive effect with lift $\geq 0.25$.
Significance vs.\ size-matched Monte-Carlo null:
\textsuperscript{*}\,$|z|\!\geq\!2$,
\textsuperscript{$\dagger$}\,$|z|\!\geq\!3$,
\textsuperscript{$\ddagger$}\,$|z|\!\geq\!5$.
Task abbreviations: GT = Greater-Than, GT\$ = Greater-Than-Price,
SVA = Subject--Verb Agreement, SVA-Bi = SVA-Binary, SVA-BLi = SVA-BLiMP,
CO = Color--Object, CC = Country--Capital, PC = Person--Country,
IOI = Indirect Object Identification, PP = Person--Pronoun,
GB = Gender Bias, GP = Gendered Pronoun.}
\label{tab:cross-task-lift}
\footnotesize
\setlength{\tabcolsep}{4.2pt}
\renewcommand{\arraystretch}{1.05}
\begin{tabular}{@{}l c S S S S S S S S S S S S@{}}
\toprule
\multicolumn{1}{c}{\textbf{Set}} & {$|S|$}
 & {\textbf{GT}} & {\textbf{GT\$}}
 & {\textbf{SVA}} & {\textbf{SVA-Bi}} & {\textbf{SVA-BLi}}
 & {\textbf{CO}} & {\textbf{CC}} & {\textbf{PC}}
 & {\textbf{IOI}} & {\textbf{PP}}
 & {\textbf{GB}} & {\textbf{GP}} \\
\midrule
Universal & 9
 & \bfseries 0.99\textsuperscript{$\ddagger$}
 & \bfseries 0.86\textsuperscript{$\ddagger$}
 & \bfseries 1.79\textsuperscript{$\ddagger$}
 & \bfseries 1.74\textsuperscript{$\ddagger$}
 & \bfseries 1.79\textsuperscript{$\ddagger$}
 & \bfseries 1.34\textsuperscript{*}
 & \bfseries 1.47\textsuperscript{$\ddagger$}
 & \bfseries 1.25\textsuperscript{$\dagger$}
 & \bfseries 0.86\textsuperscript{$\dagger$}
 & \bfseries 1.11\textsuperscript{$\ddagger$}
 & \bfseries 4.51\textsuperscript{$\ddagger$}
 & \bfseries 1.82\textsuperscript{$\ddagger$} \\
\midrule
{GT\textsubscript{spec}\textsuperscript{a}} & 2
 & \bfseries 0.32\textsuperscript{$\ddagger$}
 & \bfseries 0.40\textsuperscript{$\ddagger$}
 & -0.02 & -0.03 & -0.06
 & -0.30\textsuperscript{*}
 & -0.13\textsuperscript{$\ddagger$}
 & -0.05\textsuperscript{$\dagger$}
 & -0.50\textsuperscript{$\ddagger$}
 & -0.08
 & -0.06 & -0.02 \\
{GT\textsubscript{broad-4}\textsuperscript{b}} & 4
 & \bfseries 0.50 & \bfseries 0.45
 & 0.09 & 0.19 & 0.15
 & -0.22 & -0.06 & -0.03
 & -0.07
 & -0.05
 & -0.70 & -0.18 \\
{GT\textsubscript{all}} & 8
 & \bfseries 0.51 & \bfseries 0.51
 & -0.24 & -0.20 & -0.11
 & -0.70 & -0.67 & -0.45
 & -0.69
 & -0.34
 & -1.77 & -0.41 \\
\addlinespace[2pt]
{SVA\textsubscript{spec}} & 2
 & 0.00 & 0.01
 & \bfseries 0.53\textsuperscript{$\ddagger$}
 & \bfseries 0.62\textsuperscript{$\ddagger$}
 & \bfseries 0.80\textsuperscript{$\dagger$}
 & 0.07 & 0.00 & 0.00
 & -0.01
 & -0.01
 & \bfseries 2.20\textsuperscript{$\ddagger$}
 & \bfseries 0.95\textsuperscript{$\ddagger$} \\
{SVA\textsubscript{broad}} & 3
 & -0.01 & -0.01
 & \bfseries 0.27\textsuperscript{*}
 & \bfseries 0.33\textsuperscript{*}
 & \bfseries 0.70\textsuperscript{*}
 & -0.11 & -0.06 & -0.04
 & -0.05
 & -0.04
 & -0.20 & -0.06 \\
{SVA\textsubscript{all}} & 5
 & -0.09 & -0.10
 & \bfseries 1.05\textsuperscript{$\ddagger$}
 & \bfseries 1.06\textsuperscript{$\ddagger$}
 & \bfseries 1.73\textsuperscript{$\ddagger$}
 & -0.17 & -0.01 & 0.00
 & -0.10
 & 0.03
 & \bfseries 1.92\textsuperscript{*}
 & \bfseries 0.82\textsuperscript{*} \\
\addlinespace[2pt]
{Country} & 1
 & 0.00 & 0.00
 & 0.00 & 0.00 & 0.00
 & -0.02
 & \bfseries 0.59\textsuperscript{$\ddagger$}
 & \bfseries 0.82\textsuperscript{$\ddagger$}
 & 0.00
 & -0.11\textsuperscript{$\ddagger$}
 & 0.02 & -0.01 \\
{TS\textsubscript{IOI}} & 6
 & -0.12 & -0.21
 & -0.04 & -0.08 & -0.08
 & -0.53\textsuperscript{*}
 & -0.12 & -0.16
 & \bfseries 0.67\textsuperscript{$\ddagger$}
 & -0.14
 & -0.46 & -0.16 \\
\addlinespace[2pt]
{TS\textsubscript{CC}} & 1
 & 0.09\textsuperscript{$\ddagger$}
 & 0.08\textsuperscript{$\dagger$}
 & 0.00 & 0.00 & 0.00
 & 0.11
 & \bfseries 0.40\textsuperscript{$\ddagger$}
 & \bfseries 0.27\textsuperscript{$\ddagger$}
 & 0.01
 & 0.06\textsuperscript{$\ddagger$}
 & \bfseries 0.42\textsuperscript{$\ddagger$}
 & 0.08\textsuperscript{*} \\
{TS\textsubscript{PP}} & 1
 & 0.00 & 0.00
 & 0.00 & 0.00 & 0.00
 & 0.08 & -0.01 & 0.00
 & 0.01
 & 0.05\textsuperscript{$\ddagger$}
 & 0.01 & -0.01 \\
\midrule
{Random-22} & 22
 & -0.30 & -0.29
 & -0.38 & -0.43\textsuperscript{*} & -0.69
 & 0.12 & -0.78 & -0.56
 & -0.56
 & -0.60
 & -1.63 & -0.61 \\
\bottomrule
\addlinespace[2pt]
\multicolumn{14}{l}{\footnotesize\textsuperscript{a}\,GT\textsubscript{spec} collapses GT\_specialized\_top2.} \\[-1pt]
\multicolumn{14}{l}{\footnotesize\textsuperscript{b}\,GT\textsubscript{broad-4} collapses GT\_broad\_top4.} \\[-1pt]
\end{tabular}
\end{table*}

\begin{figure}[t]
    \centering
    \begin{minipage}{0.45\textwidth}
    \includegraphics[width=\linewidth]{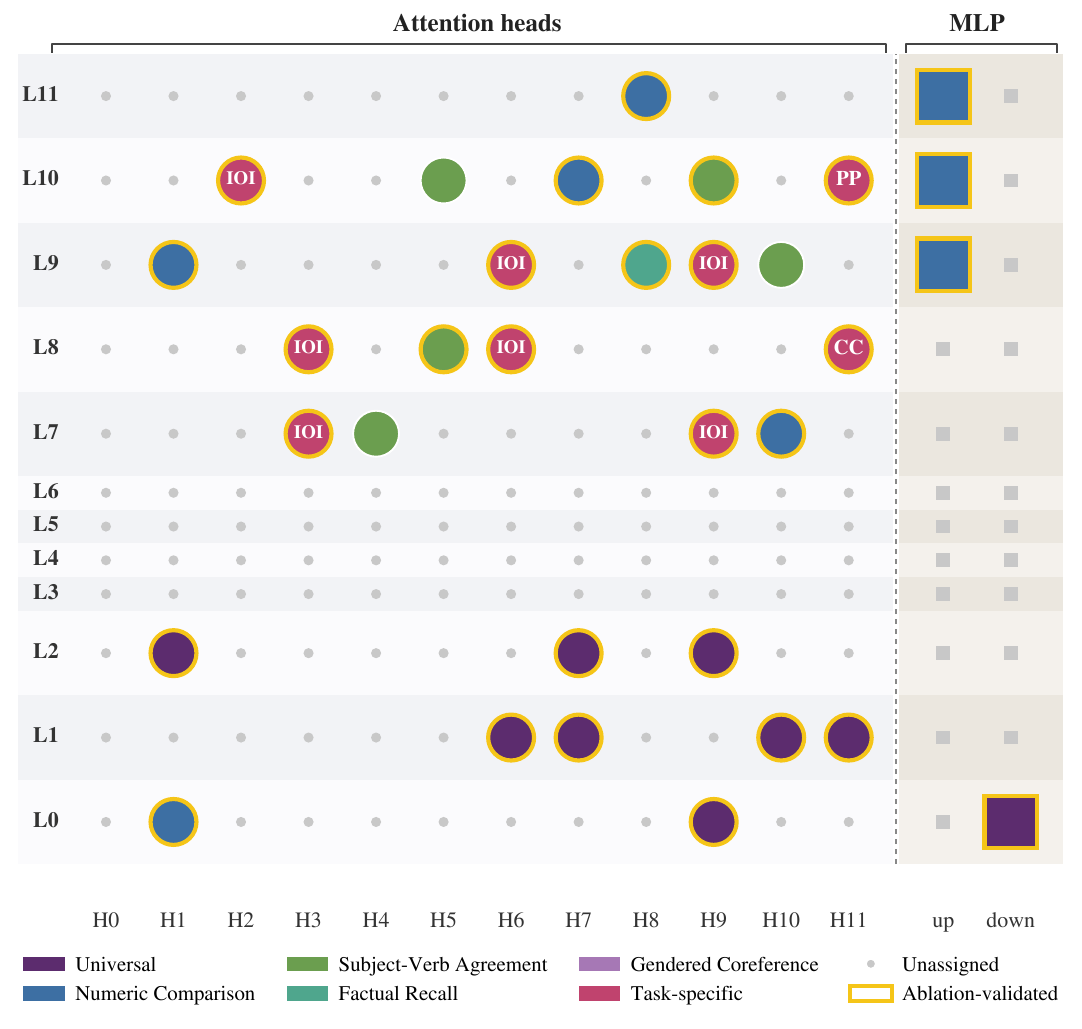}
    \end{minipage}\hspace{1em}%
    \begin{minipage}{0.45\textwidth}
    \caption{\textbf{Unit-level cross-task circuit in GPT-2 small.} A unit is colored Universal when it ranks among the top-importance units in a majority of tasks; Family-shared when its importance concentrates within one family of related tasks, and task-specific when it is uniquely required by a single task; Task-specific markers carry that task's abbreviation. A yellow outline marks units that pass causal validation  $|z|\!\geq\!3$. Grey dots denote units not assigned to any role at our discovery threshold.}
    \label{fig:gpt-fam-map}
    \end{minipage}
\end{figure}

\noindent\textbf{Universal infrastructure dominates early layers.}
The unit-level circuit in Figure~\ref{fig:gpt-fam-map} reveals a sharp
layer-wise functional partition. Layers 0--2 are almost exclusively
populated by Universal units---\texttt{L0H1}, \texttt{L0H9},
\texttt{L1H6/H7/H10/H11}, \texttt{L2H1/H7/H9}---all of which pass
ablation validation (yellow ring), recovering the
attention heads typically associated with induction and early
positional routing in the literature \citep{olsson2022context}.
Causally, the Universal set lifts every one of the 12 tasks
(lift $\in [+0.86, +4.51]$, all $|z|\!\geq\!2.8$, see Table~\ref{tab:cross-task-lift}),
confirming that these units are not merely correlated with task gradients
but are causally relied upon across the entire task suite.
Mid-layer attention (L3--L6) is conspicuously empty of family or
task labels, suggesting that GPT-2 Small's
specialization machinery is concentrated downstream of L7.

\noindent\textbf{Late-layer attention hosts family- and task-specific circuits.}
Specialization concentrates in L7--L11 attention. We highlight four
findings, each cross-validated against ablation lift in Table~\ref{tab:cross-task-lift}.

\textit{(i) The Numeric Comparison family is hosted by an exceptionally narrow
circuit.} Just two heads (\textsc{GT}\textsubscript{spec}, principally \texttt{L9H1}, \texttt{L10H7}) drive the numeric
comparison computation. Restoring the projection onto these two units
yields a modest in-family lift of $+0.32 / +0.40$, but with extreme
size-matched z-scores ($z_{\textsc{GT}}\!=\!+79.5$,
$z_{\textsc{GT\$}}\!=\!+46.2$); all other ten tasks have $|z|\!<\!1$.
The in-family/out-of-family ratio of $|2.9|$ marks GT as the
most computationally compressed family in our suite.

\textit{(ii) The Factual Recall family is even more isolated.}
The 2-head broad set lifts \textsc{PC} ($z\!=\!+119$) and
\textsc{CC} ($z\!=\!+67.6$) and almost nothing else, giving an
in/out ratio of $|72.6|$---the highest of any family and consistent
with entity$\to$attribute retrieval being a highly localized operation.

\textit{(iii) The 7-unit IOI-specific cohort recovers the canonical
IOI circuit} \citep{wang2022interpretability}.
Six of the seven SCA-discovered IOI task-specific units---
\texttt{L7H3}, \texttt{L7H9} (S-Inhibition);
\texttt{L8H6} (S-Inhibition);
\texttt{L9H6}, \texttt{L9H9} (Name Mover);
\texttt{L10H2} (Backup Name Mover)---fall within Wang et al.'s
26-head canonical IOI circuit, and all three layers of the canonical
hierarchy (S-Inhibition, Name Mover, Backup Name Mover) are recovered.
The remaining unit (\texttt{L8H3}) is an immediate layer-mate of an
S-Inhibition head not in Wang's enumeration. Causally, this cohort
lifts only \textsc{IOI} ($+0.67$, $z\!=\!+3.6$); $|z|\!<\!1$ on every
other task. The fact that an importance-and-overlap procedure
recovers a community-validated circuit \emph{without using IOI as a
target} is, to our knowledge, the strongest external validation
available for unsupervised circuit-discovery methods of this type.

\textit{(iv) Single-head readers exhibit graded specificity.}
\texttt{L8H11} carries the highest \textsc{CC} z-score ($z\!=\!+46$)
but exhibits real spillover into \textsc{PC} ($z\!=\!+40$),
\textsc{GB} ($z\!=\!+18$), and \textsc{GT} ($z\!=\!+18$),
identifying it as an entity$\to$attribute reader most strongly recruited
by \textsc{CC} rather than as a strictly single-task head.
\texttt{L10H11}, by contrast, is genuinely \textsc{PP}-specific
($z_{\textsc{PP}}\!=\!+5.4$, all other $|z|\!<\!1$).




\section{Conclusion}

In this work, we presented a gradient-spectral perspective on \emph{consistent} 
circuit discovery, showing that the information identifying causally 
relevant components is already encoded in the spectrum of forward task 
gradients across paraphrased prompts. We demonstrated that prior 
gradient-based attribution methods, by reducing per-instance gradients to 
component-wise scalars, are blind to the rotated low-rank subspace 
structure that distinguishes task-relevant directions from task-irrelevant 
ones. To address this, we proposed SCA, which applies block-wise SVD to 
paraphrase-derived gradients and redistributes spectral energy to attention 
heads and MLP units via a consistency-weighted share, yielding a circuit 
description that is stable across input formulations of the same task. 
SCA performs comparably or exceeds EAP-family baselines on twelve circuit-discovery 
tasks in GPT-2 Small and Pythia-160M, and the same per-unit subspaces 
that drive single-task discovery directly support cross-task comparison, 
exposing universal infrastructure, narrow family-level circuits, and 
compositional reuse patterns invisible to per-task analysis. 
We discuss limitations and future work 
directions in Appendix~\ref{app:limitaion}.


\newpage
{
\small
\bibliographystyle{abbrvnat}
\bibliography{ref}
}

\newpage
\appendix
\section*{\Large{Appendix}}
\newcommand{\appendixtoc}{
    \begingroup
    \let\oldcontentsline\contentsline
    \renewcommand{\contentsname}{\large\textbf{Appendix}}
    \startcontents[appendices]
    \printcontents[appendices]{}{1}{\setcounter{tocdepth}{2}}
    \endgroup
}
\appendixtoc
\newpage

\section{Limitation and Future Work}
\label{app:limitaion}

SCA's spectral subspaces are interpretive lenses rather than guaranteed 
mediators; we validate them via projection-based ablation (Table~\ref{tab:cross-task-lift}) 
but acknowledge that subspace recovery, like activation 
patching~\citep{zhang2023towards}, has design choices—paraphrase-set 
composition (whose underlying sign-consistency distribution is itself 
model-dependent; see Appendix~\ref{app:pythia}), energy threshold 
$\eta$, layer-type aggregation—whose sensitivity warrants further study.
Our experiments are confined to GPT-2 
Small and Pythia-160M; scaling SCA to billion-parameter 
models will require Gram-trick variants and is an immediate next step. More 
broadly, we view SCA as complementary to the matched-counterfactual paradigm 
rather than a replacement: counterfactual-driven methods provide direct 
causal interventions on token-level contrasts, while paraphrase-driven 
spectral methods provide a task-level geometric view. Combining the two—e.g., 
using SCA-discovered subspaces to seed targeted counterfactual 
patching—appears to be a productive direction.

\paragraph{Outlook.}
Several recent threads converge on the view that transformer computation is 
fundamentally low-rank and task-conditioned~\citep{park2023linear, 
todd2023function, merullo2024talking, hu2025understanding}. SCA offers one 
operationalization of this view: it shows that the spectrum of the forward 
gradient, taken across paraphrases, is rich enough to recover circuit-level 
structure, identify cross-task computational primitives, and surface 
compositional patterns that single-task discovery cannot reveal. We hope this 
encourages further work on attribution methods that leverage the geometric 
structure of model parameters and gradients beyond per-edge causal 
intervention.


\section{Extended Discussion on Related Work}
\label{app:related-work}
\textbf{Circuit Discovery and Attribution.}
Circuit-discovery methods differ primarily in how they quantify component importance. Intervention-based approaches estimate importance via activation replacement, including causal mediation analysis~\cite{vig2020investigating}, ACDC~\cite{conmy2023towards}, activation patching~\cite{meng2022locating,meng2022mass}, and its first-order approximations~\cite{nanda2023attribution,syed2024attribution,hanna2024have,kramar2024atp}.
Counterfactual-free~\cite{nainani2024adaptive} alternatives instead rely on different attribution objects:
token-level propagation rules~\cite{ferrando2024information}, causal graphs over dictionary-learned features~\cite{marks2024sparse}, or analytical layer-wise decompositions~\cite{hsu2024efficient}.
SCA likewise operates without counterfactual prompts, but takes the cross-instance spectrum of task gradients as its attribution object.

\textbf{Geometric Structure as a Discovery Procedure.}
There's a line of research leverages low-rank structure for mechanistic discovery. DAS~\cite{geiger2024finding} aligns interpretable variables with activation subspaces; \cite{meng2022locating,meng2022mass} write rank-one updates to install factual associations; function vectors~\cite{todd2023function} are read out and patched to trigger task execution; SVD-of-weights analyses~\cite{millidge2022singular} and inter-layer communication studies~\cite{merullo2024talking} use singular directions as objects of interpretation.
Closest to our spectral framing, \citet{ahmad2026beyond} apply SVD to a transformer's pre-trained component weights and use an optimization-based mask to identify functionally important singular directions within components. However, in SCA, the SVD is taken over forward task gradients across paraphrases, not over frozen weights.

\section{Notations and Symbols}
\label{notations}
We summarize the mathematical notations used throughout the paper in
Table~\ref{tab:notions}.

\begin{tabularx}{\linewidth}{@{}cX@{}}
\caption{Summary of notations used in this paper.}
\label{tab:notions}\\
\toprule
\textbf{Symbol} & \textbf{Meaning} \\
\midrule
\endfirsthead

\caption[]{Summary of notations used in this paper. (continued)}\\
\toprule
\textbf{Symbol} & \textbf{Meaning} \\
\midrule
\endhead

\midrule
\multicolumn{2}{r}{\emph{Continued on next page}}\\
\endfoot

\bottomrule
\endlastfoot

$f_\theta$ & Pre-trained decoder-only language model with parameters $\theta$. \\
$\theta$ & Model parameter vector. \\
$D$ & Total number of model parameters. \\
$\mathcal{T}$ & A circuit-discovery task specified by an input distribution and metric. \\
$\mathcal{D}_{\mathcal{T}}$ & Input distribution for task $\mathcal{T}$. \\
$x$ & A task instance sampled from $\mathcal{D}_{\mathcal{T}}$. \\
$\mathcal{L}_{\mathcal{T}}$ & Task loss used to compute gradients. \\
$\mathcal{M}$ & Task metric computed from next-token logits, e.g., logit difference. \\
$L$ & Number of transformer layers. \\
$H$ & Number of attention heads per layer. \\
$d_{\mathrm{model}}$ & Residual-stream dimension. \\
$d_{\mathrm{head}}$ & Attention-head dimension. \\
$d_{\mathrm{mlp}}$ & MLP hidden dimension. \\
$c_{\mathrm{mlp}}$ & Number of MLP parameter units per layer; $c_{\mathrm{mlp}}=2$ in our experiments. \\

\midrule
$\mathcal{U}$ & Set of architecturally meaningful units, including attention heads and MLP projections. \\
$u$ & A single attention-head or MLP-projection unit. \\
$\mathcal{I}_u$ & Parameter coordinates associated with unit $u$. \\
$d_u$ & Number of parameters in unit $u$, i.e., $d_u = |I_u|$. \\
$\ell(u)$ & Layer index of unit $u$. \\
$\tau(u)$ & Type of unit $u$, either $\mathrm{attn}$ or $\mathrm{mlp}$. \\
$\mathcal{U}_{\mathrm{attn}}$ & Set of attention-head units. \\
$W_V^{(\ell,h)}, W_O^{(\ell,h)}$ & Value and output projections of attention head $(\ell,h)$. \\
$W_{\mathrm{up}}^{(\ell)}, W_{\mathrm{down}}^{(\ell)}$ & Up- and down-projection matrices of the MLP at layer $\ell$. \\

\midrule
$x_i$ & The $i$-th semantic instance. \\
$x_{i,j}$ & The $j$-th semantic-preserving paraphrase of instance $x_i$. \\
$N$ & Number of semantic instances. \\
$M$ & Number of paraphrases per semantic instance. \\
$g(x)$ & Parameter gradient $\nabla_\theta L_{\mathcal{T}}(x;\theta)$. \\
$g_{i,j}$ & Gradient for paraphrase $x_{i,j}$. \\
$\bar{g}_i$ & Instance-level mean gradient, averaged over paraphrases. \\
$G$ & Gradient matrix whose $i$-th row is $\bar{g}_i^\top$. \\

\midrule
$\mathcal{B}_{\ell,\tau}$ & Layer-type block containing units in layer $\ell$ with type $\tau$. \\
$\mathcal{B}_k$ & The $k$-th indexed layer-type block. \\
$B$ & Number of layer-type blocks, $B=2L$. \\
$k(u)$ & Index of the unique block containing unit $u$. \\
$\mathcal{I}_{\mathcal{B}_k}$ & Union of parameter coordinates in block $\mathcal{B}_k$. \\
$d_k$ & Number of parameters in block $\mathcal{B}_k$, i.e., $d_k=|I_{\mathcal{B}_k}|$. \\
$G_{\mathcal{B}_k}$ & Block-restricted gradient matrix. \\
$\sigma_j^{(k)}$ & $j$-th singular value of $G_{\mathcal{B}_k}$. \\
$\eta$ & Energy threshold for effective-rank truncation; we use $\eta=0.95$. \\
$\rho_k$ & Effective rank of block $\mathcal{B}_k$. \\
$U_{\mathcal{B}_k}$ & Block-level semantic subspace from the top left singular vectors. \\
$\mathrm{idx}^{\mathrm{loc}}_u$ & Local row indices of unit $u$ within its block. \\

\midrule
$\tilde{g}^{(k)}_i$ & Row-normalized gradient of instance $i$ restricted to block $\mathcal{B}_k$. \\
$\mu_{\mathcal{B}_k}$ & Mean normalized gradient direction in block $\mathcal{B}_k$. \\
$\mathrm{cons}(\mathcal{B}_k)$ & Directional consistency of gradients within block $\mathcal{B}_k$. \\
$\mathrm{unit\_share}(u)$ & Parameter-normalized row energy of unit $u$ in its block subspace. \\
$I(u)$ & SCA importance score for unit $u$. \\
$u_{(r)}$ & Unit ranked $r$-th by importance. \\
$\mathcal{C}_K$ & Top-$K$ prefix of a unit ranking. \\

\midrule
$t,s$ & Task indices in cross-task analysis. \\
$U_{\mathcal{B}_k}^{t}$ & Block semantic subspace for task $t$. \\
$I^{t}(u)$ & Importance score of unit $u$ for task $t$. \\
$U_u^t$ & Per-unit semantic subspace for unit $u$ and task $t$. \\
$\rho_u^t$ & Effective rank of the per-unit subspace $U_u^t$. \\
$\mathrm{overlap}(u,s,t)$ & Normalized projection energy between unit-level subspaces for tasks $s$ and $t$. \\
$\widehat{I}^{t}(u)$ & Type-normalized importance of unit $u$ for task $t$. \\
$w(u,s,t)$ & Importance-weighted cross-task subspace overlap. \\
$\mathcal{F}$ & Task-family layout used for cross-task classification. \\
$\mathrm{within}_{F}(u)$ & Aggregated overlap of unit $u$ within task family $F$. \\
$\mathrm{crossmax}_{F}(u)$ & Maximum within-family score of unit $u$ outside family $F$. \\
$\mathrm{score}(u)$ & Aggregated cross-task overlap score for universal-unit gating. \\
$\theta_{\mathrm{univ}}, \theta_F^{\mathrm{floor}}, \theta_{\mathrm{dom}}, \theta_{\mathrm{out}}, \delta$ & Auto-calibrated gating thresholds for universal, family-shared, and task-specific classification. \\

\midrule
$\mathcal{D}_{\mathcal{T}}^{\mathrm{clean}}$ & Clean prompt distribution for patching-based evaluation. \\
$\mathcal{D}_{\mathcal{T}}^{\mathrm{corr}}$ & Corrupted prompt distribution for patching-based evaluation. \\
$\mathrm{patch}_{C}(x' \leftarrow x)$ & Forward pass on corrupted prompt $x'$ with units in $C$ patched from clean prompt $x$. \\
$\mathcal{M}(f_\theta \mid C)$ & Expected metric under patching with unit set $C$. \\
$r_A(K)$ & Recovery score of method $A$ at top-$K$ budget; lower is better. \\

\end{tabularx}

\section{Definition}

\subsection{Edge Attribution Patching}
\label{app:eap}

Activation patching~\citep{wang2022interpretability} measures the causal
contribution of an internal component by replacing its activation with
a counterfactual value drawn from a corrupted run and observing the
resulting change in a task metric $\mathcal{M}$. Let $E$ denote the set of edges in the model's computation graph between attention heads, MLPs, and the residual stream. Applied
exhaustively at the edge level, this procedure requires one forward
pass per edge, which scales poorly: even GPT-2 Small contains on the
order of $10^4$ edges between attention heads, MLPs, and the residual
stream. Edge Attribution Patching (EAP) and its variants approximate
this effect with gradients, reducing the cost from $\mathcal{O}(|E|)$
forward passes to a small constant number of forward--backward passes.

\paragraph{EAP.}
Edge Attribution Patching~\citep{syed2024attribution} produces a
ranking of the form by approximating activation
patching via a first-order Taylor expansion. Given a clean prompt
$x^{\mathrm{clean}}$ that elicits the target behavior and an aligned
corrupted counterpart $x^{\mathrm{corr}}$ that does not, each edge
$u \to v$ carries an activation $a_u$ produced by the source unit $u$
and consumed by the destination unit $v$. EAP scores this edge as
\begin{equation}
s_{\mathrm{EAP}}(u \to v) \;:=\; \bigl(a_u^{\mathrm{corr}} - a_u^{\mathrm{clean}}\bigr) \cdot \nabla_{a_v} \mathcal{M}\bigm|_{x^{\mathrm{clean}}},
\label{eq:eap}
\end{equation}
where $a_u^{\mathrm{clean}}$ and $a_u^{\mathrm{corr}}$ denote the source
activation under the clean and corrupted runs respectively, and the
gradient is taken with respect to $v$'s input from $u$, evaluated at
the clean forward pass. Equation~\eqref{eq:eap} is precisely the
linearization of $\mathcal{M}$ around $a_u^{\mathrm{clean}}$:
$\Delta \mathcal{M} \approx \nabla_{a} \mathcal{M} \cdot (a_u^{\mathrm{corr}} - a_u^{\mathrm{clean}})$.
Per-unit scores are then obtained by summing absolute edge scores over
incident edges. In typical mechanistic interpretability tasks,
$\mathcal{M}$ is taken to be the logit difference between a designated
correct token and an incorrect token (e.g., the indirect vs.\ direct
object in IOI). The total cost is two forward passes (clean and
corrupted) and a single backward pass, regardless of $|E|$.

\paragraph{EAP-IG.}
The integrated-gradient variant, EAP-IG~\citep{hanna2024have},
addresses a known failure mode of EAP: when the loss surface is locally
flat at $x^{\mathrm{clean}}$, the single-point gradient
$\nabla_{a_v}\mathcal{M}|_{x^{\mathrm{clean}}}$ underestimates the true
effect of the intervention $a_u^{\mathrm{clean}} \!\to\! a_u^{\mathrm{corr}}$,
even when the corresponding edge is causally important. EAP-IG
mitigates this saturation by replacing the clean-point gradient with
an average along the linear interpolation
$x^{\mathrm{corr}} \!\to\! x^{\mathrm{clean}}$ in input embedding space,
following the integrated-gradients construction
of~\citet{sundararajan2017axiomatic}:
\begin{equation}
s_{\mathrm{EAP\text{-}IG}}(u \to v) \;:=\; \bigl(a_u^{\mathrm{corr}} - a_u^{\mathrm{clean}}\bigr) \cdot \frac{1}{m}\sum_{k=1}^{m} \nabla_{a_v} \mathcal{M}\bigm|_{x^{(k)}},
\label{eq:eap-ig}
\end{equation}
where $x^{(k)} = x^{\mathrm{corr}} + \tfrac{k}{m}\bigl(x^{\mathrm{clean}} - x^{\mathrm{corr}}\bigr)$
and $m$ is the number of interpolation steps. The cost grows to $m$ forward--backward
passes, which remains independent of $|E|$.

\paragraph{EAP-IG-KL.}
The KL-based variant, EAP-IG-KL~\citep{hanna2024have}, further removes
the dependence on a task-specific logit-difference metric by taking
$\mathcal{M}$ to be the KL divergence between the clean and intervened
next-token distributions,
\begin{equation}
\mathcal{M}_{\mathrm{KL}}(x) \;:=\; \mathrm{KL}\!\bigl(p(\,\cdot \mid x^{\mathrm{clean}}) \,\big\|\, p(\,\cdot \mid x)\bigr).
\label{eq:eap-ig-kl}
\end{equation}
This metric is well-defined for arbitrary tasks without requiring a
contrastive pair of output tokens, and is therefore applicable in
multi-answer or open-ended settings where logit difference is
ill-posed (e.g., factual recall tasks with multiple admissible
completions). It is also conceptually aligned with the
faithfulness evaluation protocol itself, which typically measures
post-ablation behavior in terms of KL divergence from the clean
distribution; using KL as the attribution metric thus optimizes
directly for the quantity by which circuits are subsequently
judged. The trade-off is the loss of directional signal: KL treats
all probability redistribution as equally costly, and cannot
distinguish edges that push probability toward the correct answer
from edges that push it away from incorrect ones, which can in
some cases lead EAP-IG-KL to omit components that contribute
negatively but importantly to task behavior~\citep{hanna2024have}.

\paragraph{Common requirement.}
All three formulations--EAP, EAP-IG, and EAP-IG-KL--require an
aligned corrupted counterpart $x^{\mathrm{corr}}$ as input. The
quality of the resulting ranking depends critically on this
alignment: $x^{\mathrm{corr}}$ must differ from $x^{\mathrm{clean}}$
in exactly the task-relevant feature while preserving surface form,
length, and tokenization, since otherwise the activation difference
$a_u^{\mathrm{corr}} - a_u^{\mathrm{clean}}$ conflates the target
manipulation with incidental lexical noise. Constructing such
counterfactual pairs is feasible for templated tasks like IOI or
Greater-Than but becomes substantially harder for naturalistic or
multi-step behaviors, motivating attribution methods that operate
without an explicit corrupted reference.

\subsection{mAP}
\label{app:map}
We rank all $144$ attention heads by each method's importance 
score and evaluate the ranking against the $|\mathrm{GT}|{=}26$ 
canonical IOI heads from~\citet{wang2022interpretability}. Ranking quality is 
summarized by mean Average Precision,
\begin{equation}
\mathrm{mAP} \;=\; \frac{1}{|\mathrm{GT}|}\sum_{k=1}^{144} \mathrm{Precision}@k \cdot \mathbb{1}\!\bigl[u_{(k)} \in \mathrm{GT}\bigr],
\label{eq:map}
\end{equation}
the area under the precision--recall curve, with chance level 
$|\mathrm{Ground Truth}|/144 \approx 0.18$.

\subsection{Monte Carlo Significance Baseline}
\paragraph{Significance baseline.} Even when $U_u^{(a)}$ and
$U_u^{(b)}$ are sampled independently from the uniform (Haar)
distribution on the Stiefel manifold, the expected overlap is positive
and grows with the rank-to-ambient ratio: in particular,
$\mathbb{E}[\mathrm{ov}_u(a, b)] = \rho_u^{(a)} \rho_u^{(b)} / (d_u \cdot \min(\rho_u^{(a)}, \rho_u^{(b)}))$
under equal ranks. To control for this dimension-induced floor, we
construct a non-parametric significance threshold per shape. For each
unique triple $(d_u, \rho_u^{(a)}, \rho_u^{(b)})$ encountered across
$\mathcal{U} \times [T]^2$, draw $S = 200$ pairs of Haar-random
orthonormal frames and record their 99th-percentile overlap,
\begin{equation}
\beta\!\bigl(d_u, \rho_u^{(a)}, \rho_u^{(b)}\bigr) \;:=\;
\mathrm{quantile}_{0.99}\!\Bigl\{ \mathrm{ov}\!\bigl(\widetilde U^{(s)}_a,\, \widetilde U^{(s)}_b\bigr) \Bigr\}_{s=1}^{S}.
\label{eq:baseline}
\end{equation}
The pair $(u, a, b)$ is declared \emph{significant} iff
$\mathrm{ov}_u(a, b) > \beta(d_u, \rho_u^{(a)}, \rho_u^{(b)})$.
Baselines are computed once per unique triple and amortized across all
$|\mathcal{U}| \cdot \binom{T}{2}$ pair evaluations.

\section{Extend Method}
\subsection{Paraphrase Augmentation Diagnostics}
\label{app:para_gradient}

\subsubsection{Diagnostic: Paraphrase-Augmented Signal}
\label{sec:diag_nsr}

To verify that paraphrase averaging in Eq.~\eqref{eq:method-grad}
isolates a task-semantic signal rather than lexical noise, we compute
the \emph{noise-to-signal ratio} (NSR) between intra-instance and
inter-instance gradient distances. For each instance $i$,
\begin{equation}
d^{(i)}_{\mathrm{intra}} \;:=\; \frac{1}{M(M-1)}\sum_{j \neq k}\|g_{i,j} - g_{i,k}\|_2,
\qquad
d^{(i,i')}_{\mathrm{inter}} \;:=\; \|\bar g_i - \bar g_{i'}\|_2,
\end{equation}
and aggregate over the $N$ instances:
\begin{equation}
\mathrm{NSR} \;:=\; \frac{\frac{1}{N}\sum_{i} d^{(i)}_{\mathrm{intra}}}{\frac{1}{N(N-1)}\sum_{i \neq i'} d^{(i,i')}_{\mathrm{inter}}}.
\end{equation}

\subsubsection{Diagnostic: Sign Consistency at the Parameter Level}
\label{sec:diag_sc}

We complement the instance-level NSR with a parameter-coordinate-level
diagnostic. Writing $g_{i,j}^{(d)} \in \mathbb{R}$ for the $d$-th
coordinate of $g_{i,j}$ ($d \in [D]$), define the per-instance and
mean sign-consistency
\begin{equation}
\mathrm{SC}^{(d,i)} \;:=\; \biggl|\frac{1}{M}\sum_{j=1}^{M}\mathrm{sign}\bigl(g_{i,j}^{(d)}\bigr)\biggr| \in [0, 1],
\qquad
\mathrm{SC}^{(d)} \;:=\; \frac{1}{N}\sum_{i=1}^{N} \mathrm{SC}^{(d,i)}.
\end{equation}
Under the null of i.i.d.\ uniform random signs across the $M=5$
paraphrases of an instance, $\mathbb{E}[\mathrm{SC}^{(d)}] = 0.375$;
coordinates with $threshold \geq 0.7$ thus exhibit
sign-coherence well above chance.
Results of Augmented Signal Analysis and Sign Consistency in GPT-2 Small are shown in Table~\ref{tab:cross_task_metrics}.

\begin{table}[htbp]
\centering
\caption{Cross-task subspace overlap metrics on GPT-2 Small.}
\label{tab:cross_task_metrics}
\small
\setlength{\tabcolsep}{6pt}
\begin{tabular}{lcccc}
\toprule
Task & NSR & $d_{\text{intra}}^{\text{mean}}$ & $d_{\text{inter}}^{\text{mean}}$ & $SC_{\text{mean}}$ \\
\midrule
Indirect Object Identification & 0.5667 & 0.4336 & 0.7651 & 0.6157 \\
Greater-Than                   & 0.3402 & 0.1905 & 0.5601 & 0.7279 \\
Greater-Than-Price             & 0.5273 & 0.2858 & 0.5420 & 0.6762 \\
Country-Capital                & 0.2089 & 0.1845 & 0.8834 & 0.7246 \\
Person-Country                 & 0.2949 & 0.2406 & 0.8157 & 0.7142 \\
Person-Profession              & 0.2929 & 0.2608 & 0.8905 & 0.7072 \\
Colored Objects                & 0.8095 & 0.6262 & 0.7736 & 0.5435 \\
Gender Bias                    & 0.2899 & 0.2357 & 0.8129 & 0.6848 \\
Gender Pronoun                 & 0.2652 & 0.2257 & 0.8509 & 0.7033 \\
Subject-Verb Agreement         & 0.1916 & 0.1735 & 0.9057 & 0.7353 \\
SVA-Binary                     & 0.1318 & 0.1185 & 0.8985 & 0.7649 \\
SVA-BLiMP                      & 0.4572 & 0.4006 & 0.8762 & 0.6138 \\
\bottomrule
\end{tabular}
\end{table}


\subsection{Cross-Task Analysis: Implementation Details}
\label{app:family_details}

This appendix provides the implementation details deferred from
Section~\ref{sec:cross_task}: the auto-calibration recipe for the
gating thresholds (Appendix~\ref{app:thresholds}), the
specialized/broad refinement of the \textsc{Family-}$F$ label
(Appendix~\ref{app:specialized_broad}), and the family-layout
robustness protocol (Appendix~\ref{app:layout}). Notation follows
Sections~\ref{sec:prelim}--\ref{sec:cross_task}.

\subsubsection{Auto-Calibrated Gating Thresholds}
\label{app:thresholds}

The gates G1--G3 of Section~\ref{sec:cross_task} depend on five
scalar thresholds and one null-distribution sample size. To avoid
hand-tuning, all are derived from empirical quantiles over the unit
population $\mathcal{U}$, computed under the importance-weighted
overlap $w$ of Eq.~\eqref{eq:weighted}.

\paragraph{Quantile-based thresholds.} For a real-valued statistic
$\phi : \mathcal{U} \to \mathbb{R}$ and $\alpha \in (0, 1)$, let
$q_\alpha(\phi) := \inf\{c : |\{u : \phi(u) \leq c\}| / |\mathcal{U}| \geq \alpha\}$
denote the $\alpha$-quantile of $\phi$ over $\mathcal{U}$. The five
thresholds are

\begin{align}
\theta_{\mathrm{univ}} = q_{0.90}(\mathrm{score}),
\quad
\theta_F^{\mathrm{floor}} = q_{0.95}(\mathrm{cross}_F^{\max}), \notag \\
\theta_{\mathrm{dom}} = q_{0.90}(\widehat I^{\mathrm{ratio}}),
\quad
\theta_{\mathrm{out}} = q_{0.50}(w_{\max}^{\neq t^*}),
\quad
\delta = 0.05,
\label{eq:thresholds}
\end{align}

where $\widehat I^{\mathrm{ratio}}(u) := \widehat I^{\,t^*\!}(u) / \max_{t \neq t^*} \widehat I^{\,t}(u)$
and $w_{\max}^{\neq t^*}(u) := \max_{t \neq t^*} w(u, t^*, t)$ are
evaluated at $t^* = \arg\max_t \widehat I^{\,t}(u)$. The
$\theta_F^{\mathrm{floor}}$ rule guarantees by construction that any
qualifying within-family score exceeds at least 95\% of cross-family
scores; the additive margin $\delta$ removes near-ties on the
G2 boundary.

\paragraph{Null-distribution sample size.} The significance test for
$\mathrm{overlap}(u, s, t)$ in Eq.~\eqref{eq:overlap} compares against
a Monte-Carlo null formed from $R_{\mathrm{null}} = 1000$ pairs of
random orthonormal subspaces of matched ranks $(\rho_u^s, \rho_u^t)$
in $\mathbb{R}^{d_u}$, drawn by orthonormalizing standard Gaussian
matrices. The $95^\mathrm{th}$ percentile is computed empirically
without distributional assumptions.

\paragraph{Family-specific floor overrides.} A small number of
families exhibit naturally low cross-overlap with the rest of $[T]$,
which depresses $\mathrm{cross}_F^{\max}$ and lets G2 admit weak
within-family signals. For such families we override
$\theta_F^{\mathrm{floor}}$ with the larger of
$q_{0.95}(\mathrm{cross}_F^{\max})$ and a global floor
$q_{0.50}(\mathrm{within}_F)$ aggregated across all families; the
override is reported alongside any G2 assignment.

\subsubsection{Specialized vs.\ Broad Family Members}
\label{app:specialized_broad}

Units classified \textsc{Family-}$F$ by gate G2 are further refined
by intra-family magnitude. A unit $u$ assigned to family $F$ is
\textsc{specialized} if
\begin{equation}
\mathrm{within}_F(u) \;\geq\; \theta_F^{\mathrm{strict}},
\qquad
\theta_F^{\mathrm{strict}} := q_{0.99}(\mathrm{within}_F),
\end{equation}
and \textsc{broad} otherwise. The \textsc{specialized} tag isolates
the strongest in-family contributors---typically a handful of units
per family---and is used in the experiments only to inspect the head
of the within-family distribution; the headline G1--G3 labels do not
depend on this refinement.

\subsubsection{Family-Layout Robustness}
\label{app:layout}

The family-membership classifier of Section~\ref{sec:cross_task} is
parametrized by the family layout $\mathcal{F}$. Singleton-as-family
layouts are degenerate: a single-member family has no within-family
pair, leaving $\mathrm{within}_F$ undefined. We therefore evaluate two
non-degenerate layouts in parallel:
\begin{itemize}
\item \textbf{v1} (strict): $\mathcal{F}$ contains only the
multi-task families specified \emph{a priori} from the task taxonomy;
tasks not belonging to any such family are family-ineligible and can
only be assigned by gates G1, G3, or G4.
\item \textbf{v2} (catch-all): $\mathcal{F}$ additionally pools every
otherwise-orphan task into a single catch-all family, allowing all
tasks to participate in G2 evaluation.
\end{itemize}
For each gate's two aggregation choices ($\min$, $\mathrm{mean}$) we
re-run the classifier under both layouts, yielding four label
assignments per unit. Units receiving the same headline label
(\textsc{Universal}, \textsc{Family-}$F$ for the same $F$,
\textsc{Task-}$t^*$ for the same $t^*$, or \textsc{Mixed}) under all
four configurations are tagged \emph{layout-stable}; only
layout-stable assignments enter the headline counts of
Section~\ref{sec:experiments}, while layout-sensitive units are
reported separately.

\section{Tasks and Datasets}
\label{app:tasks}

\subsection{Per-Task Descriptions}

We construct paraphrases for eleven mechanistic-interpretability benchmarks,
each available in matched GPT-2 and Pythia variants. For every task, each
source row receives \textbf{five} paraphrase of $p_\text{clean}$ that preserve the original task semantics (the model's
required next-token prediction is unchanged) while varying lexical and
syntactic surface form. The single exception is IOI, for which we provide
15 paraphrases per row because we use it as a template task to study sensitivity analysis about the number of paraphrases $M$. Below, for each task, we first give the task
definition, then its source and citation, then its \texttt{clean} /
\texttt{corrupted} format with one representative example.
 
\subsubsection{IOI --- Indirect Object Identification}
 
\paragraph{Task.}
Given a two-name introduction followed by a transfer event, predict the
\emph{indirect object} --- the name that does not repeat as the agent of the
transfer.
 
\paragraph{Source.}
\emph{Interpretability in the Wild: a Circuit for
Indirect Object Identification in GPT-2 Small}, \citet{wang2022interpretability}. The dataset is
released with the \texttt{interpretability-wild} repository accompanying
that paper. Seven base templates underlie the dataset (\texttt{fun},
\texttt{argument}, \texttt{went\_then}, \texttt{thinking}, \texttt{working},
\texttt{after\_went}, \texttt{got}).
 
\paragraph{Format and example.}
\begin{verbatim}
clean        :  "Then, Lauren and Laura were working at the office.
                 Lauren decided to give a kiss to"
correct_token:  " Laura"
corrupted    :  "Then, Lauren and Laura were working at the office.
                 Kristen decided to give a kiss to"
corrupted_token: " Lauren"
\end{verbatim}
The released GPT-2 / Pythia version of the dataset truncates the prompt
before the indirect object; the answer is stored as \texttt{correct\_token}.
The corruption is implicit: the discovered IOI circuit is evaluated against
an ABC-shuffled name set rather than a single corrupted prompt, following
\citet{wang2022interpretability}.
 
\subsubsection{GB --- Gendered Bias}
 
\paragraph{Task.}
Predict the gendered pronoun (`` he'' / `` she'') referring to the sentence
subject. The clean answer reflects the occupation's stereotypical gender; the
corrupted answer reflects an explicit gendered noun (\texttt{man} /
\texttt{woman}).
 
\paragraph{Source.}
Sentence templates of the form ``\emph{The \{occupation\} \{verb-phrase\}
\{complementiser\}}'', derived from gender-bias evaluation work in the
spirit of the WinoBias coreference benchmark \citep{zhao2018gender} and
the StereoSet stereotype benchmark \citep{nadeem2021stereoset}. The
clean--corrupted contrast flips the subject from a stereotypically gendered
occupation to a generic gendered noun, isolating the contribution of the
occupation itself to the predicted pronoun.
 
\paragraph{Format and example.}
\begin{verbatim}
clean            : "The photographer wished that" 
                    <- clean (female-stereotyped)
corrupted        : "The man wished that"          
                    <- corrupted (explicit male noun)
clean_answer     : " she"
corrupted_answer : " he"
\end{verbatim}
 
\subsubsection{GP --- Gendered Pronoun}
 
\paragraph{Task.}
Predict the pronoun (`` he'' / `` she'') co-referent with the named subject
of the sentence.
 
\paragraph{Source.}
Sentence templates of the form ``\emph{\textless adverbial\textgreater,
\{name\} \{verb-phrase\} \{complementiser\}}'' over 18 base templates,
using real person names with a conventional gender association. The
clean--corrupted contrast swaps to a name of the opposite gender, isolating
gender attribution from sentence structure. Related to the pronoun-resolution
probing work of \citet{vig2020investigating} and the gender-bias-in-coreference
setup of \citet{zhao2018gender}.
 
\paragraph{Format and example.}
\begin{verbatim}
clean_query         : "After the long meeting, Ronald said that"  
                        <- clean (male)
corrupted_query     : "After the long meeting, Amy said that"     
                        <- corrupted (female)
clean_completion    : " he"
corrupted_completion: " she"
\end{verbatim}

\subsubsection{SVA --- Subject-Verb Agreement}
 
\paragraph{Task.}
At the prediction position, choose between a singular-agreeing verb form
and a plural-agreeing verb form. The evaluation aggregates probability mass
over a curated pool of $\sim$1,100 singular/plural verb pairs (e.g.\ \emph{abandons / abandon},
\emph{achieves / achieve}, \dots, \emph{is / are}, \emph{was / were}), so the
target is a \emph{verb pool} rather than a single token pair.
 
\paragraph{Source.}
\emph{Targeted Syntactic Evaluation of Language
Models}, \citet{marvin2018targeted}. Templates cover 14 syntactic groups (object-relative
clauses, subject-relative clauses, sentential complements, prepositional
phrases, etc.). The clean--corrupted contrast flips the head noun's number
while leaving every other token identical. The accompanying singular/plural
verb pool follows the agreement-form list released with that work.
 
\paragraph{Format and example.}
\begin{verbatim}
clean            : "The song the ministers admire"  
                    <- clean (singular subject "song")
corrupted        : "The songs the ministers admire" 
                    <- corrupted (plural)
clean_answer     : (canonical singular form; one entry in the verb pool)
corrupted_answer : " are" (canonical plural form; one entry in the verb pool)
group            : :" is" ML-obj_rel_no_comp_across_inanim_all"
plural_meaning   : "singular_subject"
\end{verbatim}
The released JSON stores the canonical pair `` is'' / `` are'' in the
\texttt{clean\_answer} / \texttt{corrupted\_answer} fields for convenience;
evaluation aggregates over the full singular/plural verb pool.

\subsubsection{SVA-Binary --- Subject-Verb Agreement (Binary)}

\paragraph{Task.}
A simplified, distractor-free variant of SVA. Predict the verb form
(`` is'' / `` are'') that agrees in number with the head noun, where
the head noun is the only noun that could control agreement at the
prediction position. The clean--corrupted contrast flips the head
noun's number; the evaluation metric is the logit difference between
the two target tokens, $\mathrm{logit}[\text{`` is''}] - \mathrm{logit}[\text{`` are''}]$.

\paragraph{Source.}
A binary-target subset of \citet{marvin2018targeted}, restricted to the
syntactic groups \texttt{simple\_agrmt} and \texttt{sent\_comp\_all},
in which no relative-clause or prepositional-phrase distractor noun
intervenes between the head noun and the agreement target. The intent
is to isolate the simplest is/are decision and make it directly
comparable in format to the binary-target tasks elsewhere in this
benchmark (GB, GP, CO, etc.).

\paragraph{Format and example.}
\begin{verbatim}
clean            : "The author"
                    <- clean (singular subject "author")
corrupted        : "The authors"
                    <- corrupted (plural)
clean_answer     : " is"
corrupted_answer : " are"
group            : "ML-simple_agrmt"
plural_meaning   : "singular_subject"
\end{verbatim}
 
\subsubsection{SVA-BLiMP --- Subject-Verb Agreement (BLiMP)}
 
\paragraph{Task.}
As in SVA, but using BLiMP-style minimal pairs in which the subject's number
is flipped together with its determiner.
 
\paragraph{Source.}
\emph{BLiMP: The Benchmark of Linguistic Minimal
Pairs for English}, \citet{warstadt2020blimp}. Specifically the
\texttt{distractor\_agreement\_relative\_clause} and related agreement groups
in which a relative clause separates the subject from the agreement target.
 
\paragraph{Format and example.}
\begin{verbatim}
clean            : "These students who loved the Borgias"   
                    <- clean (plural)
corrupted        : "This student who loved the Borgias"     
                    <- corrupted (singular)
group            : "blimp-distractor_agreement_relative_clause"
plural_meaning   : "plural_subject"
\end{verbatim}
 
\subsubsection{GT --- Greater-Than}
 
\paragraph{Task.}
Given a sentence describing an event spanning years \texttt{YYZZ} to
\texttt{YY}, predict any 2-digit year strictly greater than \texttt{ZZ}.
 
\paragraph{Source.}
\emph{How does GPT-2 compute greater-than?}, \citet{hanna2023does}. The corrupted prompt fixes \texttt{ZZ = 01}, removing the
lower-bound constraint (any 2-digit year $>01$ is admissible) and thereby
serving as the corruption baseline against which the greater-than circuit
is identified.
 
\paragraph{Format and example.}
\begin{verbatim}
clean       : "The plan lasted from the year 1760 to the year 17"  
                <- clean (ZZ=02)
corrupted   : "The plan lasted from the year 1701 to the year 17"  
                <- corrupted (ZZ=01)
correct_idx : 60
\end{verbatim}
 
\subsubsection{GT-Price --- Greater-Than Price}
 
\paragraph{Task.}
Given ``\emph{The price of that \{noun\} ranges from \$ XXZZ to \$ XX}'',
predict any 2-digit suffix strictly greater than \texttt{ZZ}.
 
\paragraph{Source.}
A price-domain analogue of Greater-Than (the Year version), sharing the underlying numeric
reasoning structure of \citet{hanna2023does} but substituting a shopping
context for a temporal one. The substitution preserves the greater-than
constraint while exposing whether the discovered circuit transfers across
domains.
 
\paragraph{Format and example.}
\begin{verbatim}
clean        : "The price of that table ranges from $ 1165 to $ 11"  
                <- clean (ZZ=65)
corrupted    : "The price of that table ranges from $ 1101 to $ 11"  
                <- corrupted (ZZ=01)
correct_idx  : 65
\end{verbatim}
 
\subsubsection{CC --- Country--Capital}
 
\paragraph{Task.}
Given a capital city, predict the country it is the capital of.
 
\paragraph{Source.}
\emph{Locating and Editing Factual Associations in
GPT}, Capital-name prompts following the factual-association probing methodology
of~\citet{meng2022locating}. The clean--corrupted contrast replaces the capital with a different real capital whose answer differs.
 
\paragraph{Format and example.}
\begin{verbatim}
clean              : " Ashgabat, the capital of"   -> " Turkmenistan"  
                        <- clean
corrupted          : " Thimphu, the capital of"    -> " Bhutan"        
                        <- corrupted
capital            : "Ashgabat"
country            : "Turkmenistan"
corrupted_capital  : "Thimphu"
corrupted_country  : "Bhutan"
\end{verbatim}

\subsubsection{CO --- Colored Objects}
 
\paragraph{Task.}
Given a small visual scene description, identify either the colour of a
queried object (inventory style) or the object having a queried colour (QA
style).
 
\paragraph{Source.}
Object--colour binding stimuli in the BIG-Bench \emph{colored\_objects} task
family, \emph{Beyond the Imitation Game:
Quantifying and Extrapolating the Capabilities of Language Models} \citep{srivastava2023beyond}.
Two formats are released:
\begin{itemize}
\item \textbf{CO (inventory-style)} --- list-then-query.
\item \textbf{CO-QA (question-answering style)} --- descriptive sentences
with a wh-question.
\end{itemize}
In the inventory format, corruption swaps the \emph{queried object} (so the
predicted colour flips). In the QA format, corruption swaps the
\emph{colour--object binding} in the scene (so the answer to the same
question is a different object).
 
\paragraph{Format and example.}
\begin{verbatim}
clean (inventory): " Inventory: a white dress, a grey shelf, an orange pencil.
                    Item pencil is"   -> " orange"     <- clean
corrupted        : " Inventory: a white dress, a grey shelf, an orange pencil.
                    Item dress is"    -> " white"      <- corrupted
 
clean (QA)       : " The ruler is brown. The folder is purple.
                    What is brown? It is the"   -> " ruler"   <- clean
corrupted        : " The ruler is purple. The folder is brown.
                    What is brown? It is the"   -> " folder"  <- corrupted
\end{verbatim}
 
\subsubsection{PC --- Person--Country}
 
\paragraph{Task.}
Given a famous person, predict their country of citizenship.
 
\paragraph{Source.}
Sentences of the form ``\emph{\textless prefix\textgreater\ \{name\}
\textless verb-phrase\textgreater\ of/in}'' over 8 base templates, where
\texttt{\{name\}} is a real person and the prediction target is a country
of citizenship. The construction follows the factual-association probing
methodology of \citet{meng2022locating} and \citet{hernandez2023linearity},
\emph{Linearity of Relation Decoding in Transformer Language Models}, in
which a known relation triple (\emph{subject}, \emph{relation},
\emph{object}) is realised as a free-form completion prompt. The
clean--corrupted contrast replaces the named entity with another whose
citizenship differs.
 
\paragraph{Format and example.}
\begin{verbatim}
clean            : " Netanyahu was a national of"  
                    <- clean (Israeli politician)
corrupted        : " Orwell was a national of"     
                    <- corrupted (English author)
clean_answer     : " Israel"
corrupted_answer : " England"
template         : "{name} was a national of"
\end{verbatim}
 
\subsubsection{PP --- Person--Profession}
 
\paragraph{Task.}
Given a famous person, predict their profession.
 
\paragraph{Source.}
Sentences of the form ``\emph{\textless prefix\textgreater\ \{name\}
\textless verb-phrase\textgreater\ a}'' over 24 base templates, where
\texttt{\{name\}} is a real person and the prediction target is a
single-token profession noun. As with PC, the construction follows the
factual-association probing methodology of \citet{meng2022locating} and
\citet{hernandez2023linearity}. The clean--corrupted contrast replaces the
named entity with another whose profession differs.
 
\paragraph{Format and example.}
\begin{verbatim}
clean            : " From an early age, Hegel aspired to become a"
                    <- clean (German philosopher)
corrupted        : " From an early age, Columbus aspired to become a"
                    <- corrupted (Italian explorer)
clean_answer     : " philosopher"
corrupted_answer : " sailor"
template         : "From an early age, {name} aspired to become a"
\end{verbatim}

\subsection{Paraphrase Methodology}
\label{app:paraphrase}
 
\subsubsection{Approach}\label{app:para_approach}

\subsection{Approach}
\label{sec:appendix:approach}

All paraphrase templates were \textbf{generated by a large language model
and then verified by the authors}. Specifically, Anthropic's Claude
(Sonnet/Opus, accessed through the Claude.ai web interface) was prompted
to produce candidate paraphrase templates for each task, given the original
\texttt{clean} / \texttt{corrupted} stimulus and the task-specific hard
invariants (last-token preservation, entity-count preservation, no
answer-leaking vocabulary in the template frame, etc.). The model also
generated the verification scripts that statically and dynamically check
these invariants. The authors then manually inspected each template for
fluency and semantic equivalence, edited or rejected templates that failed
inspection, and ran the verification scripts on every generated paraphrase
pair. Templates that did not pass all hard invariants were discarded; only
templates that passed both human inspection and automated verification are
included in the released datasets.

\textbf{The released datasets themselves are produced by deterministic
template-substitution scripts (Python 3, no model dependencies)}; no
paraphrase text is generated by a language model at runtime. Random
sampling steps (e.g.\ selecting one of twelve discourse-setting frames in
SVA / Country-Capital) use \texttt{random.Random(42)}, so the released
datasets are fully reproducible.

Why a template-based release rather than runtime neural paraphrasing? Three
reasons:

\begin{enumerate}
\item \textbf{Exact preservation of the corruption pattern.}
Mechanistic-interpretability evaluation requires that $p_\text{clean}$ and
$p_\text{corrupted}$ differ \emph{only} at the entity position(s) of the
original task, with the same flipped values as the original. A neural
paraphraser used at runtime cannot guarantee this byte-level invariant;
even strong models substitute pronouns, drop modifiers, or reorder
constituents in ways that violate the corruption identity. Authoring
templates once and substituting slot fillers deterministically lets us
guarantee the invariant exactly.

\item \textbf{Hard-constraint verifiability.}
We require every paraphrase pair to satisfy a fixed set of \emph{hard
invariants} (last-token preservation, entity-count preservation,
byte-identical prefix and suffix between $p_\text{clean}$ and
$p_\text{corrupted}$, no answer-leaking vocabulary in the template frame).
These invariants are statically and runtime-verifiable for fixed templates,
and infeasible to enforce on free-form neural paraphraser outputs.

\item \textbf{Forbidden-vocabulary enforcement.}
Several tasks require that no answer-leaking word appear in the template
frame: profession words for PP, country/nationality words for PC and
CC, gendered pronouns for GB and GP, colour and object words
outside the original scene for CO. Static forbidden-list checks are
straightforward for fixed templates and difficult for free-form
neural-paraphraser outputs.
\end{enumerate}

Every generated paraphrase pair $(p_\text{clean}, p_\text{corrupted})$ is
verified against the hard invariants in Table~\ref{tab:invariants};
generation halts with an assertion error if any invariant fails.

\begin{table}[ht]
\centering
\small
\begin{tabular}{ll}
\toprule
\textbf{ID} & \textbf{Invariant} \\
\midrule
H1 & Last token of $p_\text{clean}$ equals last token of original \texttt{clean} (preserves prediction position). \\
H2 & The critical entity (name, capital, start-year, head-noun, etc.) appears the same number of times \\
   & in $p_\text{clean}$ as in original \texttt{clean}. \\
H3 & $p_\text{clean}$ and $p_\text{corrupted}$ differ ONLY at the corruption positions, with the same \\
   & flipped values as the original. \\
H4 & The template frame contains no answer-leaking vocabulary (gendered words, country words, \\
   & profession words, etc., as appropriate per task). \\
H5 & The template frame contains no third-party document-source word (\emph{narrator}, \emph{journalist}, \\
   & \emph{biographer}, \emph{archive}, etc.) that introduces a new entity not present in the original. \\
\bottomrule
\end{tabular}
\vspace{0.5em}
\caption{Hard invariants verified at generation time for every paraphrase
pair.}
\label{tab:invariants}
\end{table}
 
H3 is enforced via byte-level partition around the entity slot, so each
paraphrase decomposes as $\texttt{prefix} + \texttt{slot\_filler} +
\texttt{suffix}$ and the prefix and suffix bytes are required to match
exactly between $p_\text{clean}$ and $p_\text{corrupted}$. This is stricter
than mere lexical identity; it eliminates any prefix/suffix drift that could
contaminate the clean-vs-corrupted logit difference.
 
\subsubsection{Per-Task Generation Strategy}
\label{app:paraphrase-template}
 
\paragraph{IOI.}
15 paraphrases per row, each ending with the literal token \texttt{to}.
Within each of the seven base templates, the operations applied are:
\begin{itemize}
\item \textbf{Subordinate-clause fronting} ---
  ``While A and B were ..., S handed a Y to''
\item \textbf{It-cleft on the event clause} ---
  ``It was at the X that S gave a Y to''
\item \textbf{Participial / possessive NP} ---
  ``Following A and B's trip to the X, ...''
\item \textbf{Sentence splitting} ---
  ``A and B visited the X. S then handed a Y to''
\item \textbf{Connective and verb variation} ---
  Then $\to$ Afterwards / Later / Next; gave $\to$ handed / passed / offered / presented
\item \textbf{Light reversal} ---
  location fronting (``At the X, A and B \dots'')
\end{itemize}
 
\paragraph{GB.}
5 paraphrases per row, each preserving the original complementiser
(\texttt{that} or \texttt{because}) at the final position:
\begin{itemize}
\item \textbf{Adverbial fronting} ---
  ``Earlier that day, the \{occupation\} \{verb\} \{comp\}''
\item \textbf{Evaluative adverb} ---
  ``Apparently, the \{occupation\} \{verb\} \{comp\}''
\item \textbf{It-cleft on the subject} ---
  ``It was the \{occupation\} who \{verb\} \{comp\}''
\item \textbf{Parenthetical insertion} ---
  ``The \{occupation\}, by all accounts, \{verb\} \{comp\}''
\item \textbf{Post-nominal modifier} ---
  ``The \{occupation\} at the time \{verb\} \{comp\}''
\end{itemize}
Gendered tokens (\texttt{he}, \texttt{she}, \texttt{man}, \texttt{woman},
etc.) are forbidden in the template frame.
 
\paragraph{GP.}
5 paraphrases per \emph{base template} (18 base templates $\times$ 5 = 90
unique paraphrase templates), authored by direct restructuring of the
original sentence --- no third-party narrators, journalists, archives, or
other external sources are introduced. Operations:
\begin{itemize}
\item \textbf{Adverbial-clause synonym} ---
  ``After the long meeting, \dots'' $\to$ ``Following the long meeting, \dots''
\item \textbf{Subordinate-time recasting} ---
  ``After the long meeting, \dots'' $\to$ ``Once the long meeting was over, \dots''
\item \textbf{Prepositional-phrase recasting} ---
  ``After the long meeting, \dots'' $\to$ ``At the end of the long meeting, \dots''
\item \textbf{Sentence splitting} ---
  ``After the long meeting, X said that'' $\to$ ``The long meeting was finally over. X said that''
\item \textbf{Mid-position adverbial insertion} ---
  ``X, after the long meeting, said that''
\end{itemize}
A subset of templates uses participial or conditional restructurings.
 
\paragraph{SVA/SVA-Binary.}
5 paraphrases per row through \emph{discourse embedding}. The original
stimulus is wrapped in a matrix clause from one of twelve discourse
settings (novel paragraph, news report, dialogue line, journal entry,
screenplay direction, biographer's account, witness statement, archival
note, voice-over, podcast intro, narrator account, letter). For each row
we deterministically sample one setting and produce five syntactically
distinct realisations of that setting:
\begin{itemize}
\item \textbf{Declarative + that-complement} ---
  ``The novel records that \{stim\}''
\item \textbf{Passive + that-complement} ---
  ``It is recorded in the novel that \{stim\}''
\item \textbf{What-cleft} ---
  ``What the novel records is that \{stim\}''
\item \textbf{Adverbial fronting (no that)} ---
  ``In the novel, \{stim\}''
\item \textbf{Biclausal conditional} ---
  ``If you turn the next page of the novel, you will read that \{stim\}''
\end{itemize}
The original SVA contrast (head-noun number flip) is preserved exactly
inside the embedded clause.

\paragraph{SVA-BLiMP.}
5 surface-form perturbations per row, applied within the embedded relative
clause and on peripheral modifiers:
\begin{itemize}
\item \textbf{Relative-pronoun substitution} ---
  ``students who loved'' $\to$ ``students that loved''
\item \textbf{Participial recasting} ---
  ``students who loved'' $\to$ ``students having loved''
\item \textbf{Reduced-relative recasting} ---
  ``students who loved X'' $\to$ ``students fond of X''
\item \textbf{Mid-position adverbial insertion} ---
  ``These students who loved X'' $\to$ ``These students, in their youth, who loved X''
\item \textbf{Determiner / quantifier alternation} ---
  ``These students'' $\to$ ``All of these students''
\end{itemize}
Pronouns and other number-marked elements outside the original head NP are
forbidden in the template frame.
 
\paragraph{GT.}
5 paraphrases per row, all retaining the literal sentence-final substring
``the year YY'':
\begin{itemize}
\item \textbf{Declarative active SVO} ---
  ``The \{noun\} ran from the year YYZZ until the year YY''
\item \textbf{Passive PP} ---
  ``The \{noun\} was conducted from the year YYZZ to the year YY''
\item \textbf{Fronted temporal adverbial} ---
  ``Beginning in the year YYZZ, the \{noun\} continued until the year YY''
\item \textbf{Nominal inversion} ---
  ``The duration of the \{noun\} spanned from the year YYZZ to the year YY''
\item \textbf{It-cleft} ---
  ``It was the \{noun\} that ran from the year YYZZ to the year YY''
\end{itemize}
Template frames contain no digits, so no spurious 4-digit year or stray
2-digit YY value can leak into the prompt.
 
\paragraph{GT-Price.}
5 paraphrases per row using shopping-domain verbs and frames, ending with
the literal substring ``\$ XX'':
\begin{itemize}
\item \textbf{Declarative active SVO} ---
  ``That \{noun\} sells from \$ XXZZ up to \$ XX''
\item \textbf{Passive PP} ---
  ``That \{noun\} is priced from \$ XXZZ to \$ XX''
\item \textbf{Fronted temporal adverbial} ---
  ``Starting at \$ XXZZ, the price of that \{noun\} can reach \$ XX''
\item \textbf{Nominal inversion} ---
  ``The cost of that \{noun\} spans from \$ XXZZ to \$ XX''
\item \textbf{It-cleft} ---
  ``It is that \{noun\} that costs anywhere from \$ XXZZ to \$ XX''
\end{itemize}
As in GT, the template frame contains no digits.
 
\paragraph{CC.}
We sample one of twelve discourse-setting extensions per row (encyclopedia,
advertisement, auction-catalogue, journalist column, retail description,
customer review, valuation report, market summary, e-commerce listing,
appraiser's note, gift-guide blurb, designer-brand press caption --- drawn
from a pool disjoint from SVA's twelve settings), and produce 5
paraphrases that preserve the chosen setting's semantic theme while
realising five distinct sentence types. Using the encyclopedia setting as
illustration:
\begin{itemize}
\item \textbf{Active declarative} ---
  ``Encyclopedia sources note that \{capital\}, the capital of''
\item \textbf{Passive} ---
  ``In encyclopedia sources, \{capital\} is recorded as the capital of''
\item \textbf{What-cleft} ---
  ``What encyclopedia sources record is that \{capital\}, the capital of''
\item \textbf{Adverbial fronting} ---
  ``According to encyclopedia sources, \{capital\}, the capital of''
\item \textbf{Biclausal conditional} ---
  ``If you consult any standard reference, you will see that \{capital\}, the capital of''
\end{itemize}
All variants end with the literal substring ``the capital of''. Frames
contain no specific country, nationality, language, or currency.
 
\paragraph{CO (inventory and QA).}
5 paraphrases per row, preserving the matrix-sentence final-position token
and the scene-internal binding semantics:
\begin{itemize}
\item \textbf{Connective alternation between scene clauses} ---
  ``.'' $\to$ ``;'' $\to$ ``and''
\item \textbf{Question-form alternation} ---
  ``What is brown?'' $\to$ ``Which item is brown?''
\item \textbf{Determiner pattern alternation} ---
  ``a white dress'' $\to$ ``one white dress''
\item \textbf{Order alternation among scene clauses} ---
  reorder the clauses listing the objects (without changing the queried item)
\item \textbf{Connective + question-form combined} ---
  combinations of the above
\end{itemize}
Frames contain no extra colour, object, or location words beyond the
original scene.
 
\paragraph{PC.}
5 paraphrases per \emph{base template} (8 $\times$ 5 = 40 unique paraphrase
templates), authored by direct restructuring of the original sentence:
\begin{itemize}
\item \textbf{Adverbial fronting} ---
  ``On legal documents, \{name\} was a citizen of''
\item \textbf{Mid-position adverbial} ---
  ``\{name\}, throughout life, was a citizen of''
\item \textbf{Prepositional-phrase recasting} ---
  ``By legal status, \{name\} was a citizen of''
\item \textbf{Adverbial-synonym replacement} ---
  ``By all accounts, \{name\} \dots'' $\to$ ``As widely acknowledged, \{name\} \dots''
\item \textbf{Matrix-it / relative-clause / appositive-fronting / participial}
  --- e.g.\ ``It is widely held that \{name\} was a citizen of'';
  ``\{name\}, who was a celebrated figure, was a citizen of'';
  ``Being a celebrated figure, \{name\} was a citizen of''
\end{itemize}
Template frames contain no specific country, nationality, language, or
continent name (a forbidden list of 50+ tokens), no third-party sources,
and no gendered pronouns.
 
\paragraph{PP.}
5 paraphrases per \emph{base template} (24 $\times$ 5 = 120 unique
paraphrase templates):
\begin{itemize}
\item \textbf{Adverbial fronting} ---
  ``Throughout life, \{name\} pursued a career as a''
\item \textbf{Mid-position adverbial} ---
  ``\{name\}, by trade, worked as a''
\item \textbf{Lexical synonym} ---
  ``\{name\} dedicated their life to being a'' $\to$
  ``\{name\} devoted their entire life to being a''
\item \textbf{Prepositional-phrase recasting} ---
  ``By profession, \{name\} was a'' $\to$
  ``As a matter of profession, \{name\} was a''
\item \textbf{Restructuring (active $\leftrightarrow$ passive, conjunction,
  participial, conditional)} --- e.g.\ ``By most people, \{name\} is known
  as a''; ``\{name\} was a celebrated figure and was a''; ``If asked, anyone
  will tell you \{name\} was a''
\end{itemize}
Template frames contain no profession-related vocabulary (a forbidden list
of all 22 dataset answer tokens plus 60+ near-synonyms covering academic
disciplines, royal titles, military ranks, and generic vocational terms),
no gendered pronouns, and no third-party sources.

\subsection{Evaluation Metrics}
\label{app:loss}
 
We evaluate circuit recovery across paraphrases using two complementary
signal-magnitude metrics together with the three standard circuit-quality
metrics from the literature.
 
\subsubsection{Logit difference}
 
For a paraphrase pair $(p_\text{clean}, p_\text{corrupted})$ with target
tokens $t_\text{clean}$ and $t_\text{corrupted}$, the logit difference at
the prediction position is
\begin{align*}
\mathrm{LD}(p_\text{clean})     &= \mathrm{logit}_{p_\text{clean}}[t_\text{clean}]
                                 - \mathrm{logit}_{p_\text{clean}}[t_\text{corrupted}], \\
\mathrm{LD}(p_\text{corrupted}) &= \mathrm{logit}_{p_\text{corrupted}}[t_\text{clean}]
                                 - \mathrm{logit}_{p_\text{corrupted}}[t_\text{corrupted}].
\end{align*}
A correctly functioning circuit produces a \emph{positive}
$\mathrm{LD}(p_\text{clean})$ and a \emph{negative}
$\mathrm{LD}(p_\text{corrupted})$. The clean-vs-corrupted spread,
\begin{equation*}
\Delta\mathrm{LD} = \mathrm{LD}(p_\text{clean}) - \mathrm{LD}(p_\text{corrupted}),
\end{equation*}
is the canonical signal-magnitude statistic used in IOI-style circuit
analysis \citep{wang2022interpretability}. For tasks whose target spans multiple tokens
(e.g. multi-token country names in CC or PC), we use the first
target token throughout, consistent with the original task releases.
 
\subsubsection{Probability difference}
 
For tasks where unnormalised logits can be misleading (e.g. when adjacent-
vocabulary tokens absorb probability mass disproportionately, or when
prompts have very different total log-probability mass), we additionally
report the probability difference at the prediction position. Let
$p(\cdot) = \mathrm{softmax}(\mathrm{logit}(\cdot))$. Then
\begin{align*}
\mathrm{PD}(p_\text{clean})     &= p_{p_\text{clean}}[t_\text{clean}]
                                 - p_{p_\text{clean}}[t_\text{corrupted}], \\
\mathrm{PD}(p_\text{corrupted}) &= p_{p_\text{corrupted}}[t_\text{clean}]
                                 - p_{p_\text{corrupted}}[t_\text{corrupted}],
\end{align*}
with the spread
\begin{equation*}
\Delta\mathrm{PD} = \mathrm{PD}(p_\text{clean}) - \mathrm{PD}(p_\text{corrupted}).
\end{equation*}
Probability difference is bounded in $[-1, 1]$ and is more directly
comparable across prompts of different total log-probability mass than
unnormalised logit difference.
 
For the greater-than tasks (GT-Year, GT-Price), and SVA tasks (SVA, SVA-BLiMP) where the target is a
\emph{set} of admissible 2-digit completions rather than a single token,
we use the set-aggregated probability-mass form following
\citet{hanna2024have}:
\begin{equation*}
\mathrm{PD}_{\mathrm{GT}}(p) = \sum_{y \,>\, ZZ} p[\,y\,] \;-\; \sum_{y \,\le\, ZZ} p[\,y\,],
\end{equation*}
summed over all 2-digit-year (or 2-digit-suffix) tokens $y$ admissible at
the prediction position.
 
\subsubsection{Per-Task Metric Applicability}
\label{app:metricapply}
 
Table~\ref{tab:metric-applicability} summarises which metrics apply to
each task. Tasks with a single-token contrast between $t_\text{clean}$
and $t_\text{corrupted}$ (\emph{token-pair} tasks) admit both
$\Delta\mathrm{LD}$ and $\Delta\mathrm{PD}$ in their basic form. Tasks
whose answer is a \emph{set} of admissible completions (the GT family)
use the set-aggregated $\Delta\mathrm{PD}_{\mathrm{GT}}$ formulation
above; for these tasks $\Delta\mathrm{LD}$ is reported only as a
single-token approximation against the immediate boundary token
$ZZ\!+\!1$ vs.\ $01$. The IOI task is reported under the canonical
$\Delta\mathrm{LD}$ between the indirect-object name and the subject name
(both present in the prompt), as in \citet{wang2022interpretability}.

\begin{table}[ht]
\centering
\resizebox{\textwidth}{!}{%
\small
\begin{tabular}{lllll}
\toprule
\textbf{Task} & \textbf{Target structure} & \textbf{$\Delta\mathrm{LD}$} & \textbf{$\Delta\mathrm{PD}$} & \textbf{Faith./Comp./Min.} \\
\midrule
IOI         & token pair (IO vs.\ S)            & \checkmark           & \checkmark         & \checkmark \\
GB          & token pair (\emph{he} vs.\ \emph{she})  & \checkmark      & \checkmark         & \checkmark \\
GP          & token pair (\emph{he} vs.\ \emph{she})  & \checkmark      & \checkmark         & \checkmark \\
SVA         & verb pool ($V_\text{plur}$ vs.\ $V_\text{sing}$, $\sim$1.1k pairs) & approx.\ only & \checkmark (pool-agg.) & \checkmark \\
SVA-Binary         & token pair (\emph{is} vs.\ \emph{are})  & \checkmark      & \checkmark         & \checkmark \\
SVA-BLiMP         & verb pool ($V_\text{plur}$ vs.\ $V_\text{sing}$, $\sim$1.1k pairs) & approx.\ only & \checkmark (pool-agg.) & \checkmark \\
GT     & set $\{y \!>\! ZZ\}$ vs.\ $\{y \!\le\! ZZ\}$  & approx.\ only & \checkmark (set-agg.) & \checkmark \\
GT-Price    & set $\{y \!>\! ZZ\}$ vs.\ $\{y \!\le\! ZZ\}$  & approx.\ only & \checkmark (set-agg.) & \checkmark \\
CC & token pair (capital$_1$'s country vs.\ capital$_2$'s country) & \checkmark$^\dagger$ & \checkmark$^\dagger$ & \checkmark \\
CO          & token pair (colour$_1$ vs.\ colour$_2$)         & \checkmark   & \checkmark         & \checkmark \\
CO-QA       & token pair (object$_1$ vs.\ object$_2$)         & \checkmark   & \checkmark         & \checkmark \\
PC          & token pair (country$_1$ vs.\ country$_2$)       & \checkmark$^\dagger$ & \checkmark$^\dagger$ & \checkmark \\
PP          & token pair (profession$_1$ vs.\ profession$_2$) & \checkmark   & \checkmark         & \checkmark \\
\bottomrule
\end{tabular}}
\vspace{0.5em}
\caption{Metric applicability per task. \checkmark\ = applies in its basic
form; \emph{approx.\ only} = the single-token version is a boundary-token
approximation; \emph{(set-agg.)} = computed in the set-aggregated form of
\citet{hanna2023does}. $\dagger$ = first-target-token convention used when
the answer is multi-token (e.g.\ \emph{Turkmenistan}).}
\label{tab:metric-applicability}
\end{table}
 
 
 
 
 

\section{More Experiment Results}\label{app:more_exp}
\subsection{SCA Evaluation on GPT-2 Small with More Tasks}
\label{app:faithful-gpt2-6}

\begin{figure}[t]
    \centering
    \begin{subfigure}[t]{0.32\textwidth}
        \centering
        \includegraphics[width=\linewidth]{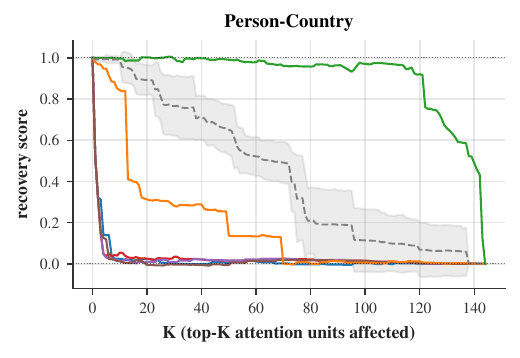}
        \caption{}
        \label{fig:faith-gpt2-pc}
    \end{subfigure}
    \begin{subfigure}[t]{0.32\textwidth}
        \centering
        \includegraphics[width=\linewidth]{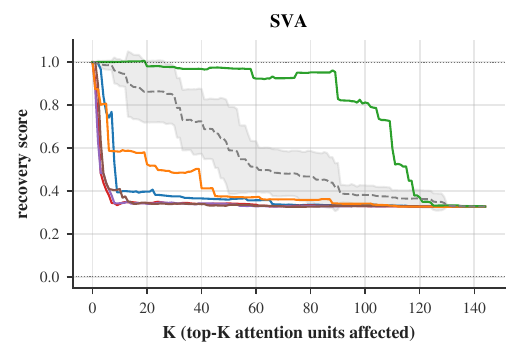}
        \caption{}
        \label{fig:faith-gpt2-sva}
    \end{subfigure}
    \begin{subfigure}[t]{0.32\textwidth}
        \centering
        \includegraphics[width=\linewidth]{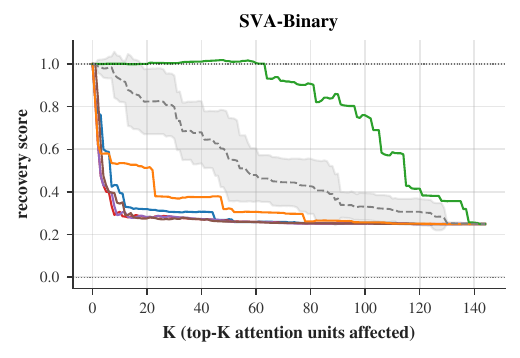}
        \caption{}
        \label{fig:faith-gpt2-sva-isare}
    \end{subfigure}
    \begin{subfigure}[t]{0.32\textwidth}
        \centering
        \includegraphics[width=\linewidth]{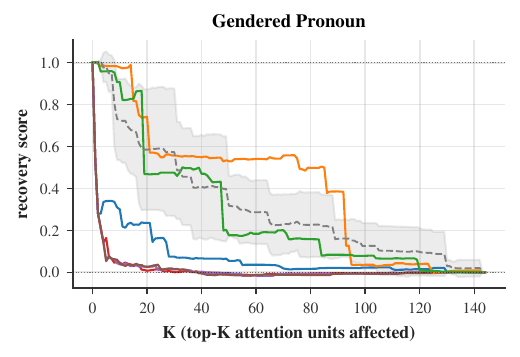}
        \caption{}
        \label{fig:faith-gpt2-gp}
    \end{subfigure}
    \begin{subfigure}[t]{0.32\textwidth}
        \centering
        \includegraphics[width=\linewidth]{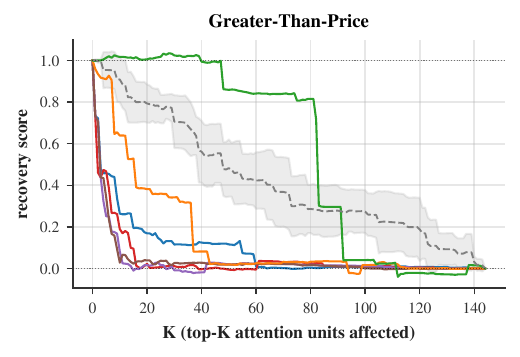}
        \caption{}
        \label{fig:faith-gpt2-gt-price}
    \end{subfigure}
    \begin{subfigure}[t]{0.32\textwidth}
        \centering
        \includegraphics[width=\linewidth]{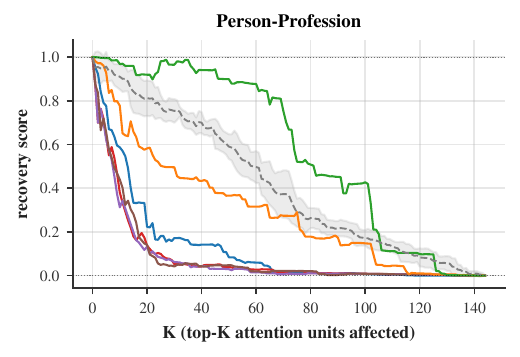}
        \caption{}
        \label{fig:faith-gpt2-pp}
    \end{subfigure}
    \begin{subfigure}[t]{0.95\textwidth}
        \centering
        \includegraphics[width=\linewidth]{pics/gpt2_per_task/legend.pdf}
    \end{subfigure}
    \caption{\textbf{Recovery score $r_\mathcal{A}(K)$ on six additional 
GPT-2 Small circuit-discovery tasks.} For each method $\mathcal{A}$, 
attention heads are ranked by importance and the top-$K$ are patched 
with corrupted activations; $r_\mathcal{A}(K)$ measures the residual 
clean--floor gap (Eq.~\ref{eq:recovery}). Lower is better. Gray 
shading: Random baseline mean $\pm 1\sigma$ over 5 seeds.}
    \label{fig:faithfulness-gpt2-6-app}
\end{figure}

Figure~\ref{fig:faithfulness-gpt2-6-app} reports recovery scores
$r_\mathcal{A}(K)$ for SCA and the six baselines on the remaining
six tasks: Person--Country, SVA, SVA-Binary, Gendered Pronoun,
Greater-Than-Price, and Person--Profession.

\noindent\textbf{SCA matches the EAP family on Person--Country, SVA,
and SVA-Binary.} On Person--Country
(Figure~\ref{fig:faith-gpt2-pc}), SCA descends jointly with the EAP
variants and reaches near-zero recovery within $K \in [10, 20]$. On
SVA and SVA-Binary
(Figures~\ref{fig:faith-gpt2-sva},~\ref{fig:faith-gpt2-sva-blimp}), all
causally grounded attribution methods---SCA, the three EAP variants,
and Magnitude---plateau at $r_\mathcal{A}(K) \approx 0.25$--$0.35$,
while Direction sits at or above Random throughout. The shared floor
across methods that disagree on ranking criterion indicates a
property of the task and the attention-head-only evaluation budget
rather than an attribution failure: a non-trivial fraction of the
clean--floor logit gap on these subject--verb agreement tasks is
not recovered by patching attention heads alone, consistent with
prior evidence that MLP sublayers carry substantial agreement
computation~\citep{geva2021transformer}. Within this ceiling, SCA
reaches the floor at the same $K$ as the EAP-IG variants.

\noindent\textbf{A modest gap appears on Gendered Pronoun,
Greater-Than-Price, and Person--Profession.} On these three tasks
(Figures~\ref{fig:faith-gpt2-gp},~\ref{fig:faith-gpt2-gt-price},
\ref{fig:faith-gpt2-pp}), SCA reaches the floor but with a small
$K$-offset relative to EAP-IG and EAP-IG-KL: SCA passes through
$r_\mathcal{A} \approx 0.2$ at $K \approx 20$--$40$, while the
EAP-IG variants pass the same value at $K \approx 5$--$15$. The gap
is most visible on Gendered Pronoun, where SCA exhibits a brief
plateau around $K \in [10, 40]$ before resuming descent. We do not
read this as a refutation of the method---all three curves still
fall well below the Random band and reach near-zero by $K \approx 60$,
and SCA achieves this without requiring counterfactual prompt
pairs---but it does mark an honest boundary: edge-level activation
attribution with integrated gradients enjoys an advantage on tasks
where the relevant computation is concentrated in a small number of
specific edges, whereas SCA's subspace estimator distributes
sensitivity across cross-instance geometry and pays a small cost
in ranking sharpness on such tasks. We return to this trade-off in
Appendix~\ref{app:limitaion}.

\noindent\textbf{The Magnitude/Direction ablation pattern is robust
across all twelve tasks.} On every panel of
Figure~\ref{fig:faithfulness-gpt2-6-app}, Magnitude and Direction
underperform both SCA and the EAP family, with Direction at or
above the Random band on five of the six panels. Combined with the
main text, this gives twelve out of twelve tasks on which the
first-moment reductions of the same per-paraphrase gradient input
fail to recover EAP-comparable rankings, while block-wise SVD on
that same input does---supporting the claim that cross-instance
subspace geometry, not gradient magnitude or direction, is the
operative signal in SCA.

\noindent\textbf{Two findings worth emphasizing.}
The cross-task evidence supports two claims that the family-level
similarity matrix alone could not establish.

\emph{Pronoun resolution has no dedicated family-level circuit.}
The Gendered Coreference cluster would suggest
a coreference family with its own machinery. However,
across any of our multi-classifier configurations,
the top-10 importance units for the
three pronoun tasks contain no member of the putative family.
Instead, the units that drive these tasks are drawn from
\textsc{SVA}-specialized heads, \textsc{GT}-specialized heads,
\textsc{CC}-readers, and Universal infrastructure---suggesting that
pronoun resolution in GPT-2 Small is \emph{compositional} rather
than \emph{dedicated}, assembled from binding, retrieval, and routing
machinery built for other purposes.

\emph{The SVA-specialized circuit causally supports pronoun tasks.}
Direct evidence for the compositional account comes from the SVA
specialized 4-head set: in addition to lifting its three target tasks
($z\!\in\![4.4, 8.9]$), it produces a substantially larger lift on
\textsc{GP} ($z\!=\!+60.4$) and \textsc{GB} ($z\!=\!+25.1$).
Ablating the SVA-binding mechanism damages pronoun tasks more than
it damages SVA itself. We interpret this as evidence that the
SVA circuit is repurposed for pronoun--antecedent
binding, with the same heads being read out under different downstream
projections---a concrete instance of the sign-agnostic, projection-dependent
nature of gradient-based importance discussed in Section~\ref{sec:method}.
This pattern---a single mechanical primitive recruited by multiple
syntactic-semantic tasks---would be invisible to per-task analyses
and is a direct consequence of the cross-task framing.

\subsection{Case Study: IOI on GPT-2 Small}
\label{app:ioi-case-study}

\subsubsection{Gradient Analysis}

\begin{figure}[t]
    \centering
    \begin{subfigure}[t]{0.28\textwidth}
        \centering
        \includegraphics[width=\linewidth]{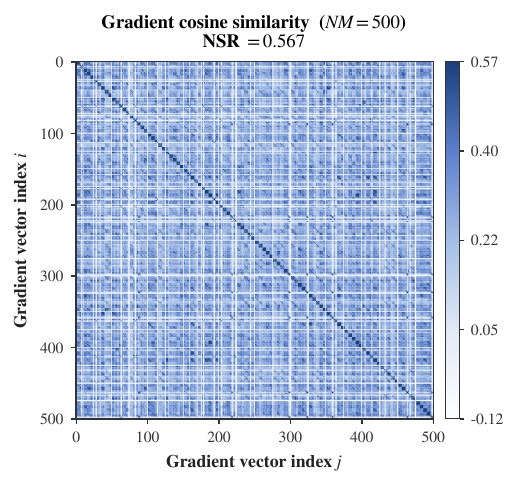}
        \caption{}
        \label{fig:ioi-grad}
    \end{subfigure}
    \hfill
    \begin{subfigure}[t]{0.35\textwidth}
        \centering
        \includegraphics[width=\linewidth]{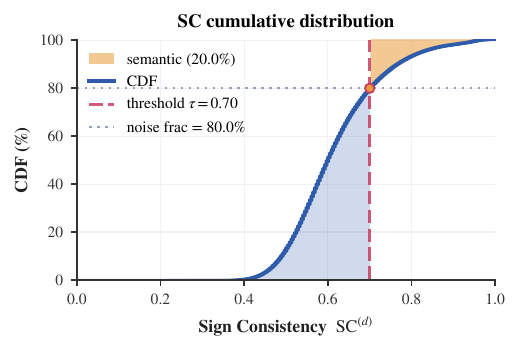}
        \caption{}
        \label{fig:ioi-sc}
    \end{subfigure}
    \hfill
    \begin{subfigure}[t]{0.35\textwidth}
    \centering
    \includegraphics[width=\linewidth]{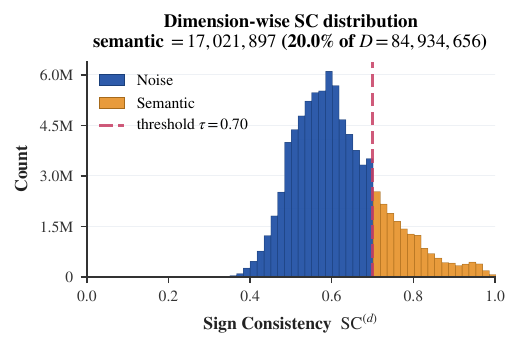}
    \caption{}
    \label{fig:ioi-his}
    \end{subfigure}
    \caption{\textbf{Gradient cosine similarity matrix on IOI in GPT-2 Small} ($NM{=}500$ vectors,
    $N{=}100$ instances $\times M{=}10$ paraphrases). Within-instance
    blocks along the diagonal are systematically darker than the
    background, but the off-diagonal mass remains elevated, yielding
    $\mathrm{NSR}=0.567$. This motivates the dimension-wise sign
    consistency filter, which removes the common gradient direction.}
    \label{fig:ioi-grad-all}
\end{figure}

Before running block discovery and SVD, we validate two upstream
assumptions of SCA on the IOI testbed: (i) that paraphrase-conditioned
gradients carry a recoverable instance-level signal above the noise
floor, and (ii) that sign-consistency (SC) filtering yields a
well-separated semantic dimension subset.

\paragraph{Noise-to-signal ratio (NSR).}
We compute the $NM \times NM$ cosine similarity matrix
$C_{ij} = \langle \tilde{g}_i, \tilde{g}_j \rangle / (\|\tilde{g}_i\|\|\tilde{g}_j\|)$
over $N{=}100$ IOI instances and $M{=}5$ paraphrases each, where
$\tilde{g}$ denotes the per-(instance, paraphrase) parameter gradient
flattened across all attention and MLP weights. As shown in
Figure~\ref{fig:ioi-grad}, IOI yields $\mathrm{NSR}=0.567$.
Within-instance blocks (along the diagonal) are visibly darker than
the across-instance background, confirming a real paraphrase-invariant
signal. However, the elevated background
indicates a non-trivial \emph{common gradient direction} dominated by
early-layer backpropagation magnitude and shared task structure;
this motivates the SC filter, which removes coordinates that fluctuate
in sign across instances.

\paragraph{Sign consistency separates semantic from noise dimensions.}
Sign consistency measures per dimension, the average within-instance agreement
of gradient signs across paraphrases. Coordinates whose signs are
genuinely paraphrase-invariant approach $\mathrm{SC}^{(d)} \to 1$;
purely noisy coordinates approach the chance level $\mathrm{SC}^{(d)} \to 0.5$
as $M \to \infty$.

Figure~\ref{fig:ioi-sc} shows the empirical SC
distribution on IOI. We observe a clear bimodal structure: a
near-Gaussian \emph{noise} mode centered at $\mathrm{SC} \approx 0.58$
(mildly elevated above the random-sign baseline of $0.5$, consistent
with the residual common gradient direction visible in the Gram
heatmap), and a long right tail of \emph{semantic} dimensions
extending to $\mathrm{SC} = 1$. We retain dimensions with
$\mathrm{SC}^{(d)} \geq \tau = 0.70$, which selects $17{,}021{,}897$
coordinates ($20.0\%$ of $D$); the threshold sits at the valley between
the two modes, and the corresponding CDF (Fig.~\ref{fig:ioi-his})
confirms that the inflection lies in the $[0.65, 0.75]$ window.

\subsubsection{SCA Discoverd Circuit}
\label{sec:case-study}

\begin{figure}[t]
    \centering
    \begin{subfigure}[t]{0.40\textwidth}
        \centering
        \includegraphics[width=\linewidth]{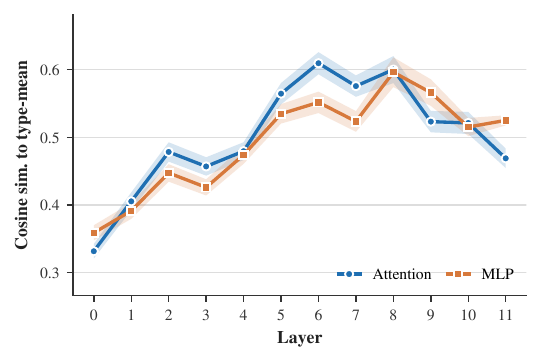}
        \caption{}
        \label{fig:ioi-cons}
    \end{subfigure}
    \hfill
    \begin{subfigure}[t]{0.58\textwidth}
        \centering
        \includegraphics[width=\linewidth]{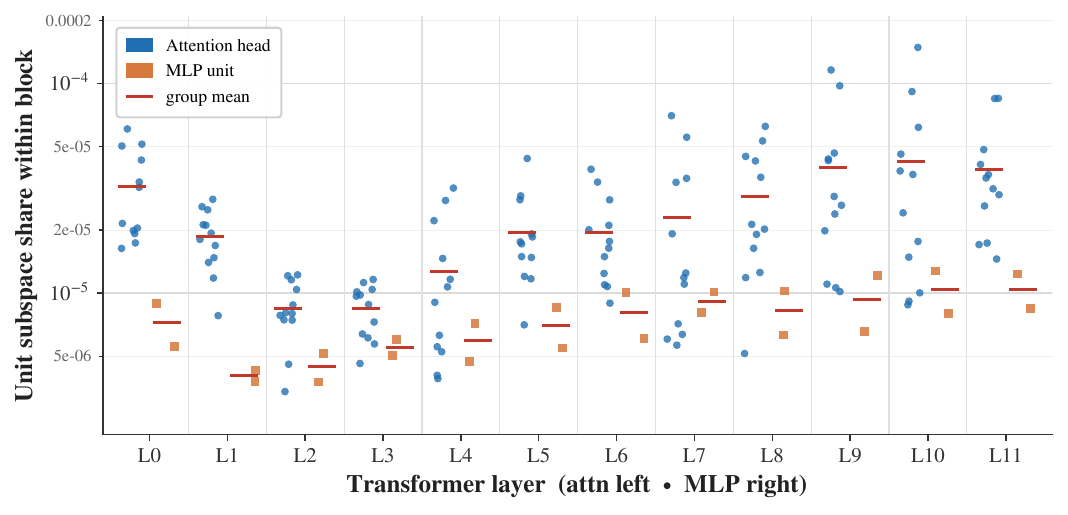}
        \caption{}
        \label{fig:ioi-share}
    \end{subfigure}
    \caption{\textbf{Decomposition of the consistency-weighted importance 
    $I(u)$ on the IOI task in GPT-2 small}. 
    \textbf{(a)} Per-block layer-type consistency 
    $\mathrm{cons}(\mathcal{B}_{\ell,\tau})$. 
    \textbf{(b)} Per-unit subspace share $\mathrm{unit\_share}(u)$ 
    within each block $\mathcal{B}_{\ell,\tau}$.}
    \label{fig:ioi-decomp}
\end{figure}

\paragraph{Geometric and directional structure of the IOI circuit.} 
Figure~\ref{fig:ioi-decomp} decomposes $I(u)$ into its two factors. 
Panel~(b) shows that $\mathrm{unit\_share}(u)$, as a purely geometric separation,
is dominated by attention 
units at every depth: late-layer heads ($\ell \in \{9, 10, 11\}$) 
attain the largest within-block shares (top heads $\approx 1.4 \times 
10^{-4}$, group means $3$--$4 \times 10^{-5}$). Panel~(a) 
provides the complementary layer-level signal: 
$\mathrm{cons}(\mathcal{B}_{\ell,\tau})$ peaks near $0.60$ at 
$\ell \in \{6, 7, 8\}$ for both component types, indicating maximal 
cross-instance directional coherence in mid-network blocks. The 
product $I(u) = \mathrm{cons} \cdot \mathrm{unit\_share}$ therefore 
concentrates importance in mid- to late-layer attention 
units: consistent with the name-mover and S-inhibition heads 
identified for the IOI circuit by~\citet{wang2022interpretability}.

\begin{figure*}[t]
    \centering
    \includegraphics[width=0.97\textwidth]{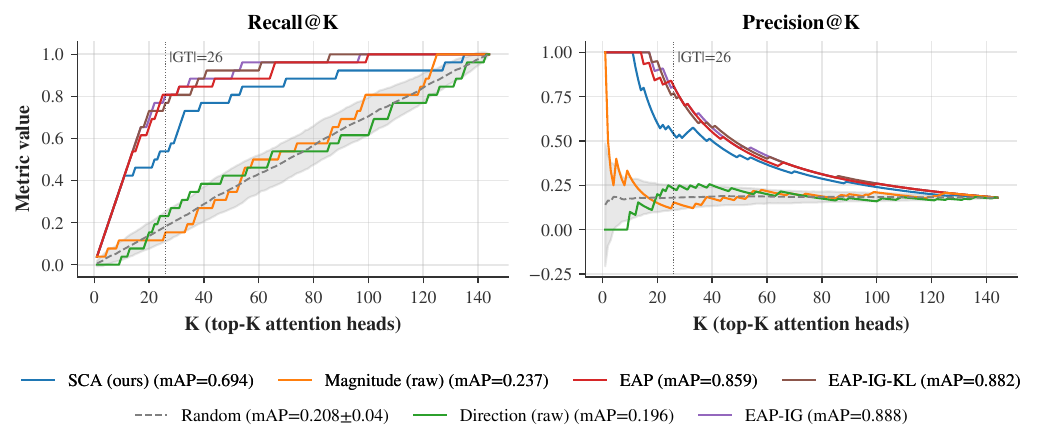}
    \vspace{-3mm}
    \caption{\textbf{Top-$K$ retrieval of canonical IOI attention heads in GPT-2 small}. All $144$ attention heads are ranked by each method's importance score and evaluated against the "Ground Truth" ($|\mathrm{GT}|{=}26$) canonical IOI heads from~\citet{wang2022interpretability}. 
    Panels show Recall@$K$ (left) and Precision@$K$ (right) as $K$
    sweeps over $[1, 144]$. SCA outperforms gradient-based baselines without using counterfactual prompt pairs; the counterfactual-based EAP family achieves the highest mAP (definition in Appendix~\ref{app:map}.)}
    \label{fig:ioi-retrieval}
\end{figure*}

\paragraph{Retrieval evaluation against canonical IOI heads.} 
 Figure~\ref{fig:ioi-retrieval} 
yields three observations. \textbf{(i)} \textbf{SCA matches the 
strongest baseline at the top of the ranking}: SCA and EAP-IG 
both recover $10/26$ canonical heads in their top-$10$ slices, 
with $\mathrm{Precision}@K \approx 1.0$ for the leading positions 
in both methods, indicating that pure paraphrase-driven spectral attribution
identifies a comparably faithful set of heads to clean--corrupted
attribution patching for IOI.
\textbf{(ii)} \textbf{SCA is preferentially sensitive to 
late-stage decision heads.} Among canonical IOI categories, SCA 
ranks Name Mover heads at mean position $2.3$ (versus $6.3$ for 
EAP-IG) and S-Inhibition heads at $5.8$ (versus $6.8$),
reflecting that the block-wise SVD basis emphasizes mid- to 
late-layer attention units that carry the largest geometric 
share of the task gradient. \textbf{(iii)} \textbf{The remaining mAP gap} 
($0.694$ vs $0.86$--$0.89$) is concentrated in the mid-range 
$K \in [15, 50]$, where EAP recovers Induction and Duplicate-Token 
heads that SCA's purely gradient-geometric criterion does not 
isolate. We do not claim like-for-like superiority over the 
counterfactual-supervised EAP family on single-task retrieval; 
SCA's distinct contribution is the cross-task subspace structure 
analyzed in Section~\ref{sec:cross-task}, which the EAP family 
does not provide.


\subsection{Transferable Experiment: Pythia-160M}
\label{app:pythia}

\subsubsection{SCA Recovery Score Evaluation}
\label{app:faith-pythia}

\begin{figure}[t]
    \centering
    \begin{subfigure}[t]{0.32\textwidth}
        \centering
        \includegraphics[width=\linewidth]{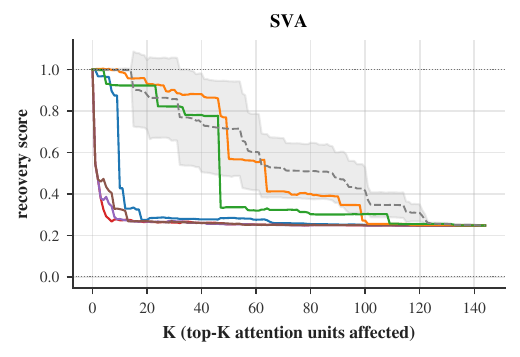}
        \caption{}
        \label{fig:faith-pythia-sva}
    \end{subfigure}
    \begin{subfigure}[t]{0.32\textwidth}
        \centering
        \includegraphics[width=\linewidth]{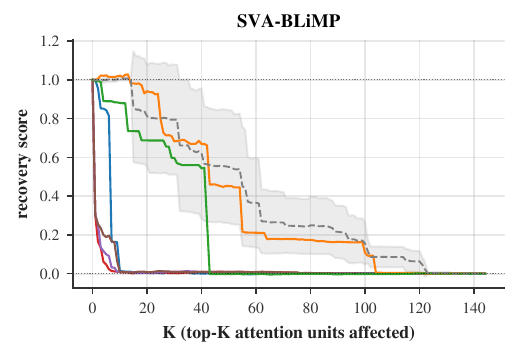}
        \caption{}
        \label{fig:faith-pythia-sva-blimp}
    \end{subfigure}
    \begin{subfigure}[t]{0.32\textwidth}
        \centering
        \includegraphics[width=\linewidth]{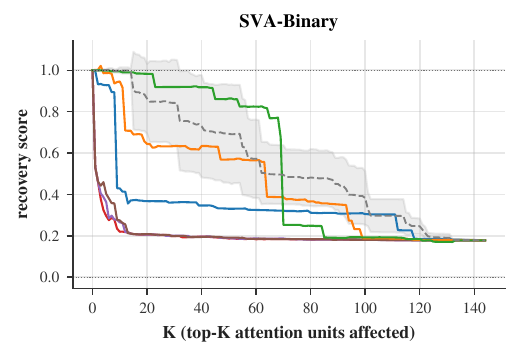}
        \caption{}
        \label{fig:faith-pythia-sva-isare}
    \end{subfigure}
    \begin{subfigure}[t]{0.32\textwidth}
        \centering
        \includegraphics[width=\linewidth]{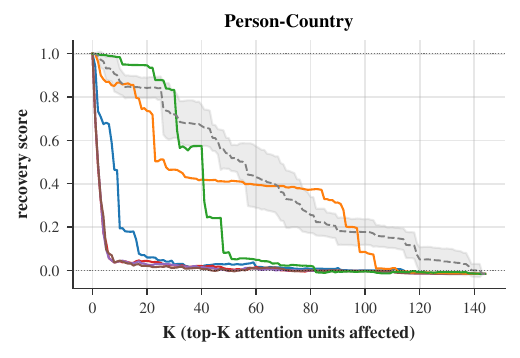}
        \caption{}
        \label{fig:faith-pythia-pc}
    \end{subfigure}
        \begin{subfigure}[t]{0.32\textwidth}
        \centering
        \includegraphics[width=\linewidth]{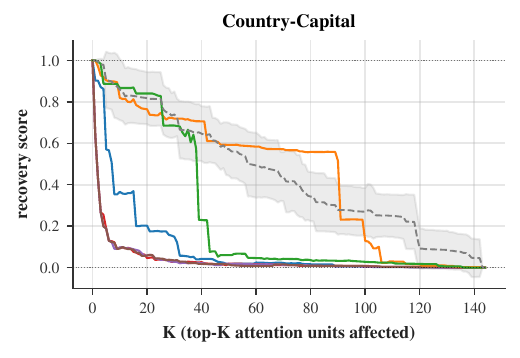}
        \caption{}
        \label{fig:faith-pythia-cc}
    \end{subfigure}
    \begin{subfigure}[t]{0.32\textwidth}
        \centering
        \includegraphics[width=\linewidth]{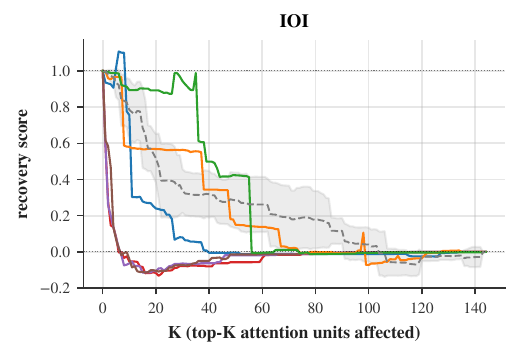}
        \caption{}
        \label{fig:faith-pythia-ioi}
    \end{subfigure}
    \begin{subfigure}[t]{0.32\textwidth}
        \centering
        \includegraphics[width=\linewidth]{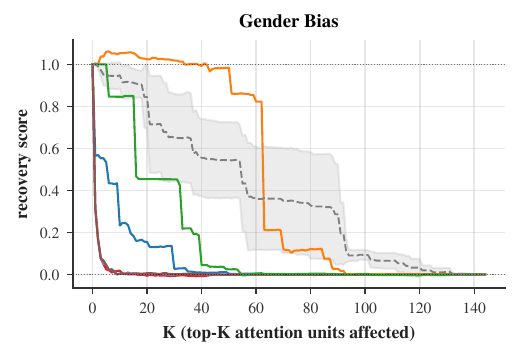}
        \caption{}
        \label{fig:faith-pythia-gb}
    \end{subfigure}
    \begin{subfigure}[t]{0.32\textwidth}
        \centering
        \includegraphics[width=\linewidth]{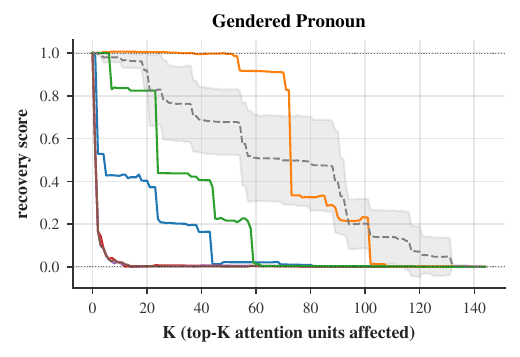}
        \caption{}
        \label{fig:faith-pythia-gp}
    \end{subfigure}
    \begin{subfigure}[t]{0.32\textwidth}
        \centering
        \includegraphics[width=\linewidth]{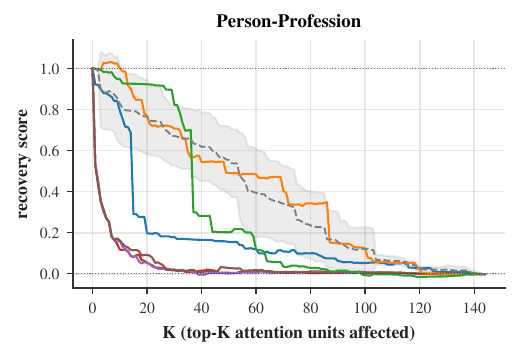}
        \caption{}
        \label{fig:faith-pythia-pp}
    \end{subfigure}
        \begin{subfigure}[t]{0.32\textwidth}
        \centering
        \includegraphics[width=\linewidth]{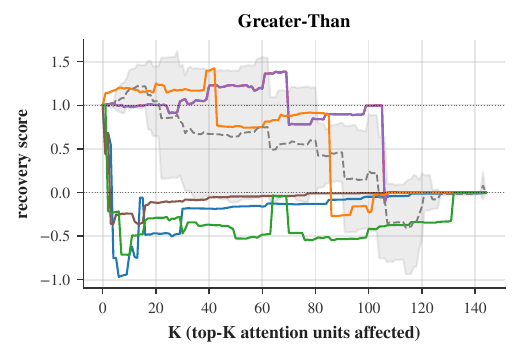}
        \caption{}
        \label{fig:faith-pythia-gt}
    \end{subfigure}
    \begin{subfigure}[t]{0.32\textwidth}
        \centering
        \includegraphics[width=\linewidth]{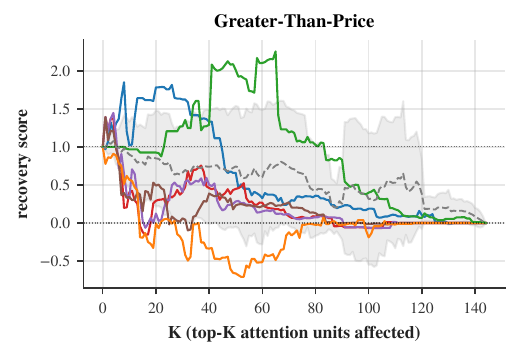}
        \caption{}
        \label{fig:faith-pythia-gt-price}
    \end{subfigure}
    \begin{subfigure}[t]{0.32\textwidth}
        \centering
        \includegraphics[width=\linewidth]{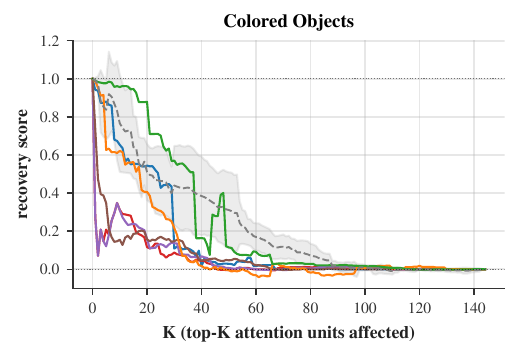}
        \caption{}
        \label{fig:faith-pythia-co}
    \end{subfigure}
    \begin{subfigure}[t]{0.95\textwidth}
        \centering
        \includegraphics[width=\linewidth]{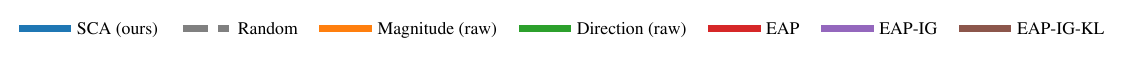}
    \end{subfigure}
    \caption{\textbf{Recovery score $r_\mathcal{A}(K)$ on twelve  
Pythia-160M circuit-discovery tasks.} For each method $\mathcal{A}$, 
attention heads are ranked by importance and the top-$K$ are patched 
with corrupted activations; $r_\mathcal{A}(K)$ measures the residual 
clean--floor gap (Eq.~\ref{eq:recovery}). Lower is better. Gray 
shading: Random baseline mean $\pm 1\sigma$ over 5 seeds.}
    \label{fig:faithfulness-pythia-12-app}
\end{figure}

Figure~\ref{fig:faithfulness-pythia-12-app} extends the recovery-score
analysis to Pythia-160M on twelve tasks. Panels (a--f)
show tasks on which SCA tracks the EAP family closely; panels
(g--l) show three distinct failure modes that we discuss in turn.

\paragraph{SCA performs comparably EAP family on six tasks (a--f).}
On Person--Country (d), Country--Capital (e), and IOI (f), SCA
descends jointly with the EAP variants and reaches near-zero
recovery within $K \in [10, 30]$, reproducing the GPT-2 Small
result. On SVA (a) and SVA-Binary (c), all causally grounded
methods---SCA, the three EAP variants, and Magnitude---share a
floor of $r_\mathcal{A} \approx 0.2$--$0.3$, mirroring the
attention-only evaluation ceiling observed on GPT-2 Small
(Figure~\ref{fig:faithfulness-gpt2-6} and Figure~\ref{fig:faithfulness-gpt2-6-app}); within this ceiling
SCA reaches the floor at the same $K$ as EAP-IG. SVA-BLiMP (b)
does not exhibit the same ceiling on this model---all methods
descend to zero---and SCA reaches the floor by $K \approx 40$,
slightly behind the EAP family ($K \approx 10$). Across the six
panels, Magnitude and Direction underperform both SCA and the
EAP family, consistent with the GPT-2 Small ablation.

\paragraph{Stair-step descent on gender- and profession-related
tasks (g--i).}
On Gender Bias (g), Gendered Pronoun (h), and Person--Profession
(i), SCA exhibits a stair-step trajectory: it passes
$r_\mathcal{A} \approx 0.4$--$0.5$ within $K \approx 5$--$10$ but
plateaus before resuming descent around $K \approx 30$--$50$,
while EAP-IG and EAP-IG-KL pass through the same value
monotonically and reach the floor by $K \approx 20$. We attribute
this gap to the gradient-noise regime on Pythia-160M.
Per-dimension sign-consistency analysis on this model yields a
heavy-tailed SC distribution: a primary mode near
$\mathrm{SC} \approx 0.38$ accounts for roughly $80\%$ of
parameter dimensions, with a substantial right tail of
high-consistency dimensions extending to $\mathrm{SC} \geq 0.9$.
The semantic signal is therefore present but more diluted than on
GPT-2 Small, and the unfiltered SVD spectrum is correspondingly
more susceptible to noise directions becoming top components on
tasks whose circuit-level gradient signal is weaker in absolute
magnitude. Tasks (g--i) appear to fall in this regime, while
tasks (a--f) carry sufficient signal to dominate the SVD spectrum
even under the heavier noise floor. All three SCA curves still
reach near-zero by $K \approx 60$ and remain below the Random band
throughout. Appendix~\ref{app:limitaion} discusses applying a
more aggressive SC threshold on high-NSR models as a concrete
extension to address this regime.

\paragraph{Greater-Than family: evaluation-metric instability,
not an attribution failure (j, k).}
On Greater-Than (j) and Greater-Than-Price (k), \emph{every}
attribution method---SCA, the three EAP variants, Magnitude, and
Direction---produces curves with $r_\mathcal{A}$ values outside
the canonical $[0, 1]$ range, with excursions to
$r_\mathcal{A} > 1.5$ or $r_\mathcal{A} < -0.5$ at intermediate
$K$. The Random baseline shows the same instability (gray shading
spans a comparable range), ruling out an attribution-specific
failure. We attribute this to the recovery-score normalization
(Eq.~\ref{eq:recovery}): on these task families the
clean--floor logit gap on Pythia-160M is small, which
makes the ratio $r_\mathcal{A}$ numerically ill-conditioned and
amplifies seed-level variance into large excursions. We observe
stable Greater-Than curves on GPT-2 Small
(Figure~\ref{fig:faith-gpt2-gt}), supporting the
small-denominator diagnosis. We report these panels for
completeness rather than as a comparative claim; disambiguating
attribution quality from metric instability on the Greater-Than
family would require either a model with a larger clean--floor
gap or an alternative faithfulness metric robust to small
denominators.

\paragraph{Colored Objects: an isolated noisy descent (l).}
On Colored Objects, SCA reaches the floor by $K \approx 40$ but
with a noticeably noisier trajectory at small $K$, including a
brief dip and recovery before $K = 10$, while the EAP family
descends cleanly. Colored Objects descends cleanly on GPT-2 Small
(Figure~\ref{fig:faith-gpt2-co}) and we have no analogous
behavior on the other eleven tasks here, so we treat this as a
single noise-level observation rather than a systematic
limitation; we do not have a principled diagnosis at this point.

\subsubsection{Cross-task Analysis}


\begin{figure*}[t]
\label{fig:pythia-fam}
    \centering
    \includegraphics[width=0.97\textwidth]{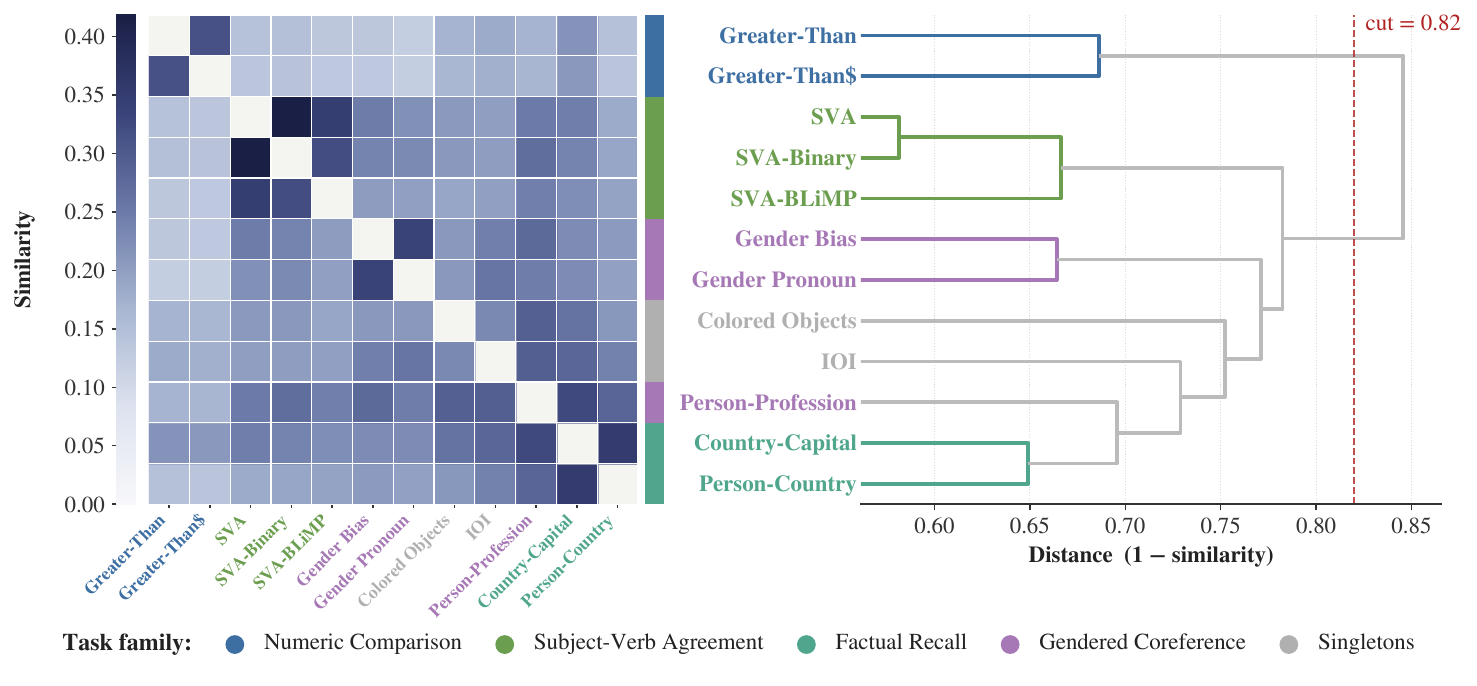}
    \vspace{-3mm}
    \caption{\textbf{Cross-task similarity on 12 Pythia-160M tasks.} Heatmap (Left): pairwise similarity, diagonal masked; vertical band marks family. Dendrogram (Right): average-linkage on $d = 1 - \text{similarity}$, branches colored by family when all leaves agree.}\label{fig:pythia-fam}
\end{figure*}

\begin{table*}[t]
\centering
\caption{Cross-task subspace lift across candidate component sets (alternative configuration).
Each row is a candidate set of Pythia-160M components (attention heads + MLPs);
each column is one of 12 evaluation tasks. Values are mean lift in target-token
logit difference when the residual stream is projected onto the candidate's
weighted unit-SVD subspace. $|S|$ is the number of components in the set.
\textbf{Bold} indicates positive effect with lift $\geq 0.25$.
Significance vs.\ size-matched Monte-Carlo null:
\textsuperscript{*}\,$|z|\!\geq\!2$,
\textsuperscript{$\dagger$}\,$|z|\!\geq\!3$,
\textsuperscript{$\ddagger$}\,$|z|\!\geq\!5$.
Task abbreviations: GT = Greater-Than, GT\$ = Greater-Than-Price,
SVA = Subject--Verb Agreement, SVA-Bi = SVA-Binary, SVA-BLi = SVA-BLiMP,
CO = Color--Object, CC = Country--Capital, PC = Person--Country,
IOI = Indirect Object Identification, PP = Person--Pronoun,
GB = Gender Bias, GP = Gendered Pronoun.}
\label{tab:cross-task-lift-alt}
\footnotesize
\setlength{\tabcolsep}{4pt}
\renewcommand{\arraystretch}{1.05}
\begin{tabular}{@{}l c S S S S S S S S S S S S@{}}
\toprule
\multicolumn{1}{c}{\textbf{Set}} & {$|S|$}
 & {\textbf{GT}} & {\textbf{GT\$}}
 & {\textbf{SVA}} & {\textbf{SVA-Bi}} & {\textbf{SVA-BLi}}
 & {\textbf{CO}} & {\textbf{CC}} & {\textbf{PC}}
 & {\textbf{IOI}} & {\textbf{PP}}
 & {\textbf{GB}} & {\textbf{GP}} \\
\midrule
Universal & 9
 & -0.04 & \bfseries 0.34
 & -0.04 & -0.01 & -0.10
 & -0.05 & 0.16 & 0.18\textsuperscript{*}
 & -0.15 & -0.05
 & \bfseries 0.96 & \bfseries 0.57 \\
\midrule
{GT\textsubscript{spec}\textsuperscript{a}} & 1
 & \bfseries 0.63 & 0.08
 & -0.28 & -0.24 & -0.41
 & -0.05 & -0.03 & -0.21
 & 0.12 & -0.03
 & \bfseries 0.39 & 0.10 \\
{GT\textsubscript{broad-4}\textsuperscript{b}} & 4
 & \bfseries 0.58 & \bfseries 0.35
 & \bfseries 0.41 & \bfseries 0.35 & \bfseries 0.26
 & -0.12 & -0.21 & -0.20
 & -0.15 & -0.04
 & -1.27 & -0.61 \\
{GT\textsubscript{broad-mlp}\textsuperscript{c}} & 1
 & \bfseries 0.55 & \bfseries 0.35
 & \bfseries 0.57 & \bfseries 0.52 & \bfseries 0.50
 & 0.06 & -0.03 & -0.06
 & 0.05 & 0.10
 & -0.87 & -0.38 \\
{GT\textsubscript{all}} & 5
 & \bfseries 0.85 & \bfseries 0.71\textsuperscript{*}
 & \bfseries 0.40 & \bfseries 0.32 & 0.17
 & 0.12 & -0.10 & -0.22
 & 0.18 & 0.14
 & -0.97 & -0.50 \\
\addlinespace[2pt]
{SVA\textsubscript{broad}\textsuperscript{d}} & 3
 & \bfseries 0.67 & \bfseries 0.39
 & \bfseries 0.71 & \bfseries 0.60 & \bfseries 0.68
 & -0.05 & -0.23 & -0.29
 & -0.05 & -0.06
 & -1.64 & -0.72 \\
\addlinespace[2pt]
{Country} & 1
 & 0.03 & -0.02
 & -0.03 & -0.04 & -0.05
 & -0.01 & 0.02 & 0.00
 & -0.02 & -0.03
 & -0.27 & -0.10 \\
{TS\textsubscript{GT}} & 5
 & \bfseries 1.01\textsuperscript{$\dagger$} & 0.19
 & -0.23 & -0.24 & -0.36
 & -0.20 & -0.18 & -0.15
 & \bfseries 0.37\textsuperscript{*} & -0.08
 & -0.51 & -0.19 \\
{TS\textsubscript{PC}} & 1
 & 0.03 & -0.02
 & -0.03 & -0.04 & -0.05
 & 0.01
 & \bfseries 0.30\textsuperscript{$\ddagger$}
 & \bfseries 0.39\textsuperscript{$\ddagger$}
 & 0.00 & 0.04
 & -0.27 & -0.10 \\
{TS\textsubscript{IOI}} & 2
 & 0.00 & 0.03
 & -0.01 & -0.01 & 0.00
 & -0.02 & -0.06 & -0.07
 & -0.05 & -0.03
 & -0.08 & -0.04 \\
{TS\textsubscript{CO}} & 1
 & 0.03 & -0.02
 & -0.03 & -0.04 & -0.05
 & -0.01 & -0.02 & -0.01
 & -0.02 & -0.03
 & -0.27 & -0.10 \\
\midrule
{Random-22} & 22
 & \bfseries 0.46 & -0.06
 & 0.12 & 0.13 & \bfseries 0.31
 & \bfseries 0.57 & -0.26 & -0.30
 & 0.00 & 0.12
 & \bfseries 2.12 & \bfseries 0.73 \\
\bottomrule
\addlinespace[2pt]
\multicolumn{14}{l}{\footnotesize\textsuperscript{a}\,GT\textsubscript{spec} collapses GT\_specialized\_1.} \\[-1pt]
\multicolumn{14}{l}{\footnotesize\textsuperscript{b}\,GT\textsubscript{broad-4} collapses GT\_broad\_top4.} \\[-1pt]
\multicolumn{14}{l}{\footnotesize\textsuperscript{c}\,GT\textsubscript{broad-mlp} collapses GT\_broad\_mlps\_top4.} \\[-1pt]
\multicolumn{14}{l}{\footnotesize\textsuperscript{d}\,SVA\textsubscript{broad} collapses SVA\_all.} \\
\end{tabular}
\end{table*}

\noindent\textbf{Family structure replicates with one re-assignment.}
Despite the noisier gradient regime documented in
Appendix~\ref{app:faith-pythia}, the spectral similarity matrix
on Pythia-160M (Figure~\ref{fig:pythia-fam}) recovers a family
decomposition nearly identical to GPT-2 Small at the same cut
threshold of $0.82$:
\textbf{Numeric Comparison} (GT, GT\$),
\textbf{Subject--Verb Agreement} (SVA, SVA-Bi, SVA-BLi),
\textbf{Factual Recall} (CC, PC), and
\textbf{Gendered Coreference} (GB, GP), with CO, IOI, and
Person--Profession remaining as singletons. The one re-assignment
relative to GPT-2 Small is Person--Profession, which grouped with
the Gendered Coreference pair on GPT-2 Small but separates here---
indicating that Pythia-160M encodes profession-attribute retrieval
as a distinct computation from gender coreference. The within-family
blocks are visibly weaker than on GPT-2 Small (peak similarity
$\approx 0.4$ vs.\ $\approx 0.6$), consistent with the elevated
noise floor on this model, but family boundaries remain sharp
enough to survive the cut.

\begin{figure}[t]
    \centering
    \begin{minipage}{0.45\textwidth}
    \includegraphics[width=\linewidth]{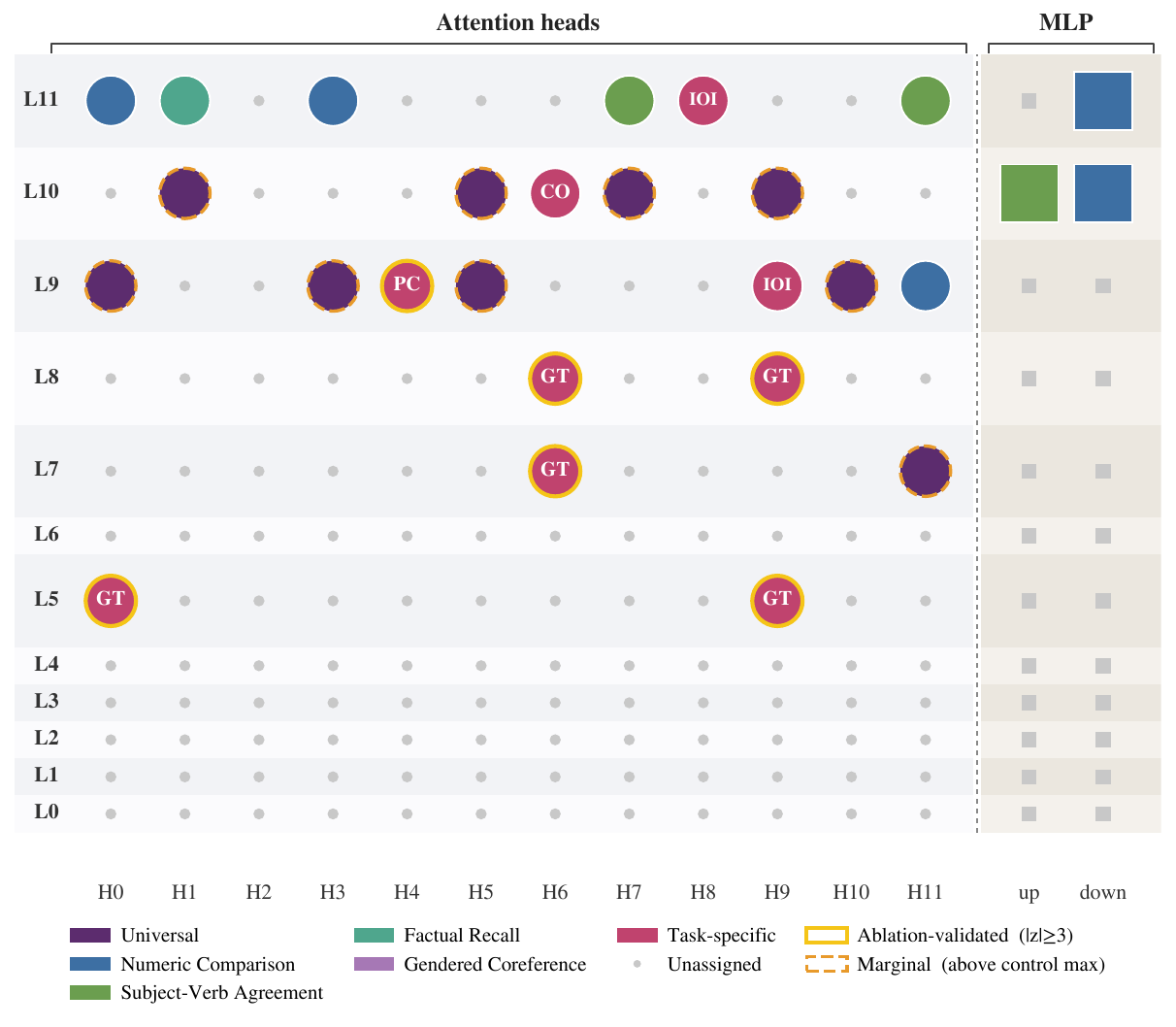}
    \end{minipage}\hspace{1em}%
    \begin{minipage}{0.45\textwidth}
    \caption{\textbf{Unit-level cross-task circuit in Pythia-160M.} A unit is colored Universal when it ranks among the top-importance units in a majority of tasks; Family-shared when its importance concentrates within one family of related tasks, and task-specific when it is uniquely required by a single task; Task-specific markers carry that task's abbreviation. An orange outline marks unit that pass caucal validation $|z|\!\geq\!2.5$ and yellow means $|z|\!\geq\!3$. Grey dots denote units not assigned to any role at our discovery threshold.}
    \label{fig:pythia-fam-map}
    \end{minipage}
\end{figure}

\noindent\textbf{Universal infrastructure migrates from early to mid-late layers.}
The most striking architectural contrast with GPT-2 Small is the
location and causal status of Universal heads
(Figure~\ref{fig:pythia-fam-map}). On GPT-2 Small, layers L0--L2
are densely populated by causally validated Universal heads
associated with induction and early positional routing
\citep{olsson2022context}. On Pythia-160M, layers L0--L4 are
\emph{completely empty} of any task or family assignment; the
Universal set instead concentrates at L9--L10 with marginal
validation (\texttt{L9H0}, \texttt{L9H3}, \texttt{L9H5},
\texttt{L9H9} ablation-validated; \texttt{L10H1}, \texttt{L10H5},
\texttt{L10H7}, \texttt{L10H9}, \texttt{L7H11} marginal). Causally,
Table~\ref{tab:cross-task-lift-alt} shows that this 9-head set
produces appreciable positive lift only on \textsc{GB} ($+0.96$)
and \textsc{GP} ($+0.57$) and yields no significance markers on
any task; the size-controlled Random baseline matches or exceeds
these values on the same gender tasks. We interpret this not as
absence of universal infrastructure on Pythia-160M but as the
combination of (i) Pythia's documented architectural and
training-distribution differences from GPT-2 shifting routing
computation downstream, and (ii) the heavier gradient-noise floor
on this model making subspace projection a less discriminating
probe at the broad Universal scale.

\noindent\textbf{Family- and task-specific circuits with graded fidelity.}
We highlight five findings, each cross-validated against ablation
lift in Table~\ref{tab:cross-task-lift-alt}.

\textit{(i) Numeric Comparison is broad and MLP-mediated, in
sharp contrast to GPT-2's narrow attention circuit.}
Where GPT-2 Small concentrated GT computation in two attention
heads, Pythia-160M distributes the same family across five
ablation-validated GT-specific attention heads
(\texttt{L5H0}, \texttt{L5H9}, \texttt{L7H6}, \texttt{L8H6},
\texttt{L8H9}) plus an MLP component
(\texttt{GT\textsubscript{broad-mlp}} at L10/L11 down).
The combined TS\textsubscript{GT} set yields the strongest in-family
GT signal in the table ($+1.01^{\dagger}$ on GT) with mild positive
spillover onto IOI ($+0.37^{*}$). The MLP unit alone produces
substantial lift on the GT family ($+0.55 / +0.35$ on GT/GT\$)
\emph{and} on the entire SVA family ($+0.57 / +0.52 / +0.50$),
identifying it as a shared agreement/comparison MLP rather than
a GT-specialized one. This MLP-mediated, cross-family pattern is
absent on GPT-2 Small and represents the largest qualitative
divergence between the two models in our analysis.

\textit{(ii) Factual Recall localizes to a single highly specific
attention head.} The TS\textsubscript{PC} entry (\texttt{L9H4})
yields ablation lift of $+0.30^{\ddagger}$ on \textsc{CC} and
$+0.39^{\ddagger}$ on \textsc{PC}, with $|z|\!<\!1$ on every other
task. This is the cleanest task-specific reader in the Pythia
table and mirrors the extreme localization of Factual Recall
observed on GPT-2 Small.

\textit{(iii) Layer 11 attention hosts a clean family-level
readout.} Six L11 attention heads carry family or task labels:
two Numeric Comparison (\texttt{L11H0}, \texttt{L11H3}), one
Factual Recall (\texttt{L11H1}), one IOI task-specific
(\texttt{L11H8}), and two Subject--Verb Agreement
(\texttt{L11H7}, \texttt{L11H11}). Combined with the L10/L11
MLPs carrying SVA (up) and Numeric Comparison (down) signal,
L11 functions as an interpretable family-level integration stage
that has no equally clean counterpart in our GPT-2 Small
visualization---suggesting Pythia-160M consolidates
family-typed computation into a narrower readout band.

\textit{(iv) Agreement/comparison subspaces are anti-aligned with
gender tasks.} Every row in Table~\ref{tab:cross-task-lift-alt}
that produces in-family positive lift on GT or SVA also produces
the largest \emph{negative} lifts in the table on gender tasks:
GT\textsubscript{broad-4} (GB $-1.27$, GP $-0.61$),
GT\textsubscript{broad-mlp} (GB $-0.87$, GP $-0.38$),
SVA\textsubscript{broad} (GB $-1.64$, GP $-0.72$). This pattern
is consistent across attention- and MLP-based projections and
appears only weakly on GPT-2 Small, suggesting that on
Pythia-160M the agreement/comparison subspace shares a parameter
direction whose suppression actively amplifies gender-token
prediction.

\textit{(v) Two task-specific cohorts produce null lift, marking
a boundary of the discovery procedure.}
TS\textsubscript{IOI} and TS\textsubscript{CO} both produce
$|$lift$|\!<\!0.1$ across all twelve tasks including their target,
despite the underlying heads passing the discovery importance
threshold. These are the same two singleton tasks where the
cross-task similarity matrix shows the weakest within-task
structure (Figure~\ref{fig:pythia-fam}). Combined with the
recovery-score behavior in Appendix~\ref{app:faith-pythia},
this indicates that on Pythia-160M these two tasks have
weakly localized circuits at the granularity of unit projections:
the discovery procedure flags candidate units, but no compact
projection direction recovers their causal effect under
ablation.

\noindent\textbf{Summary.}
Cross-task analysis on Pythia-160M reproduces the qualitative consistent
structure observed on GPT-2 Small---family decomposition,
existence of universal infrastructure, single-head localization
of Factual Recall, ablation validity of recovered GT-specific
circuits---while revealing four model-specific properties absent
on GPT-2 Small: late-layer rather than early-layer Universal
placement, distributed rather than narrow GT computation,
MLP-mediated cross-family sharing between agreement and
numeric comparison, and a graded fidelity that degrades on the
most weakly localized singleton tasks under the heavier
gradient-noise regime.

\subsection{Sensitivity Analysis}
\subsubsection{The number of Paraphrase $M$}
\label{app:m}

\begin{table}[t]
\centering
\caption{Sensitivity of SCA to the paraphrase count $M$ on the 
canonical IOI circuit. Reported values are mean $\pm$ std over 
three seeds. Best per-column results are highlighted in 
\textcolor{red}{red}; the selected operating point $M{=}5$ is 
shaded. $K{=}26$ is the matched-budget point where 
$\mathrm{P@}K \equiv \mathrm{R@}K$.}
\label{tab:m_ablation}
\setlength{\tabcolsep}{3pt}
\resizebox{\textwidth}{!}{%
\begin{tabular}{c c cc cc cc cc cc}
\toprule
$M$ & mAP & P@10 & R@10 & P@20 & R@20 & P@26 & R@26 & P@30 & R@30 & P@50 & R@50 \\
\midrule
2  & \textcolor{red}{$0.6770{\scriptstyle\pm0.0037}$} & \textcolor{red}{$0.9667{\scriptstyle\pm0.0577}$} & \textcolor{red}{$0.3718{\scriptstyle\pm0.0222}$} & \textcolor{red}{$0.6833{\scriptstyle\pm0.0289}$} & \textcolor{red}{$0.5256{\scriptstyle\pm0.0222}$} & $0.5769{\scriptstyle\pm0.0000}$ & $0.5769{\scriptstyle\pm0.0000}$ & $0.5444{\scriptstyle\pm0.0192}$ & $0.6282{\scriptstyle\pm0.0222}$ & $0.3400{\scriptstyle\pm0.0000}$ & $0.6538{\scriptstyle\pm0.0000}$ \\
3  & $0.6689{\scriptstyle\pm0.0193}$ & \textcolor{red}{$0.9667{\scriptstyle\pm0.0577}$} & \textcolor{red}{$0.3718{\scriptstyle\pm0.0222}$} & $0.6500{\scriptstyle\pm0.0500}$ & $0.5000{\scriptstyle\pm0.0385}$ & $0.5897{\scriptstyle\pm0.0222}$ & $0.5897{\scriptstyle\pm0.0222}$ & $0.5444{\scriptstyle\pm0.0192}$ & $0.6282{\scriptstyle\pm0.0222}$ & \textcolor{red}{$0.3533{\scriptstyle\pm0.0231}$} & \textcolor{red}{$0.6795{\scriptstyle\pm0.0444}$} \\
\rowcolor{gray!15}
5  & $0.6601{\scriptstyle\pm0.0327}$ & $0.9333{\scriptstyle\pm0.0577}$ & $0.3590{\scriptstyle\pm0.0222}$ & $0.6500{\scriptstyle\pm0.0500}$ & $0.5000{\scriptstyle\pm0.0385}$ & \textcolor{red}{$0.6154{\scriptstyle\pm0.0000}$} & \textcolor{red}{$0.6154{\scriptstyle\pm0.0000}$} & $0.5444{\scriptstyle\pm0.0192}$ & $0.6282{\scriptstyle\pm0.0222}$ & \textcolor{red}{$0.3533{\scriptstyle\pm0.0231}$} & \textcolor{red}{$0.6795{\scriptstyle\pm0.0444}$} \\
8  & $0.6670{\scriptstyle\pm0.0151}$ & $0.9000{\scriptstyle\pm0.0000}$ & $0.3462{\scriptstyle\pm0.0000}$ & $0.6667{\scriptstyle\pm0.0289}$ & $0.5128{\scriptstyle\pm0.0222}$ & $0.6026{\scriptstyle\pm0.0444}$ & $0.6026{\scriptstyle\pm0.0444}$ & $0.5444{\scriptstyle\pm0.0192}$ & $0.6282{\scriptstyle\pm0.0222}$ & \textcolor{red}{$0.3533{\scriptstyle\pm0.0231}$} & \textcolor{red}{$0.6795{\scriptstyle\pm0.0444}$} \\
10 & $0.6529{\scriptstyle\pm0.0085}$ & $0.9000{\scriptstyle\pm0.0000}$ & $0.3462{\scriptstyle\pm0.0000}$ & $0.6500{\scriptstyle\pm0.0000}$ & $0.5000{\scriptstyle\pm0.0000}$ & $0.5769{\scriptstyle\pm0.0000}$ & $0.5769{\scriptstyle\pm0.0000}$ & $0.5444{\scriptstyle\pm0.0192}$ & $0.6282{\scriptstyle\pm0.0222}$ & $0.3467{\scriptstyle\pm0.0115}$ & $0.6667{\scriptstyle\pm0.0222}$ \\
12 & $0.6579{\scriptstyle\pm0.0104}$ & $0.9000{\scriptstyle\pm0.0000}$ & $0.3462{\scriptstyle\pm0.0000}$ & $0.6667{\scriptstyle\pm0.0289}$ & $0.5128{\scriptstyle\pm0.0222}$ & $0.5769{\scriptstyle\pm0.0000}$ & $0.5769{\scriptstyle\pm0.0000}$ & $0.5444{\scriptstyle\pm0.0192}$ & $0.6282{\scriptstyle\pm0.0222}$ & \textcolor{red}{$0.3533{\scriptstyle\pm0.0231}$} & \textcolor{red}{$0.6795{\scriptstyle\pm0.0444}$} \\
15 & $0.6506{\scriptstyle\pm0.0141}$ & $0.8667{\scriptstyle\pm0.0577}$ & $0.3333{\scriptstyle\pm0.0222}$ & $0.6500{\scriptstyle\pm0.0000}$ & $0.5000{\scriptstyle\pm0.0000}$ & $0.5769{\scriptstyle\pm0.0000}$ & $0.5769{\scriptstyle\pm0.0000}$ & $0.5444{\scriptstyle\pm0.0192}$ & $0.6282{\scriptstyle\pm0.0222}$ & \textcolor{red}{$0.3533{\scriptstyle\pm0.0231}$} & \textcolor{red}{$0.6795{\scriptstyle\pm0.0444}$} \\
\bottomrule
\end{tabular}%
}
\end{table}

\paragraph{Validation on canonical IOI circuit.}
We validate SCA against the canonical IOI circuit characterized 
by~\citet{wang2022interpretability} in GPT-2 Small. The circuit 
specifies $|\mathrm{GT}|{=}26$ canonical attention heads---spanning 
name-mover, S-inhibition, duplicate-token, induction, and 
previous-token roles---out of $144$ total heads ($12$ layers 
$\times$ $12$ heads). We score every head with its SCA importance 
$I(u)$ (Algorithm.~\ref{alg:sca}), rank all $144$ heads, and report 
Precision@$K$ and Recall@$K$ against the ground-truth set, 
together with mean Average Precision (mAP). Note that $K{=}26$ is 
the \emph{matched-budget} operating point: when $K$ equals 
$|\mathrm{GT}|$, $\mathrm{P@}K \equiv \mathrm{R@}K$ holds 
identically and the metric reflects overall retrieval quality 
without favoring either side of the precision--recall trade-off.

\paragraph{Sensitivity to the paraphrase count $M$.}
The paraphrase count $M$ controls how aggressively per-instance 
lexical noise is suppressed in $\bar g_i = \frac{1}{M}\sum_j 
g_{i,j}$ (Eq.~\ref{eq:method-grad}); the residual noise decays 
as $\mathcal{O}(M^{-1/2})$ while the task-invariant component is 
preserved. Table~\ref{tab:m_ablation} reports mAP and 
Precision/Recall at $K \in \{10, 20, 26, 30, 50\}$ for 
$M \in \{2, 3, 5, 8, 10, 12, 15\}$, with mean $\pm$ standard 
deviation over three seeds. The pipeline is robust across the 
full sweep: mAP varies only within $[0.6506, 0.6770]$, and the 
three smallest settings ($M{=}2, 3, 5$) are statistically 
indistinguishable on mAP given the observed seed-level standard 
deviations. Larger $M$ values exhibit a mild but consistent 
degradation, suggesting that beyond a small number of 
paraphrases, additional surface-form variants no longer add 
task-relevant gradient diversity.

\paragraph{Choice of $M{=}5$.}
We adopt $M{=}5$ as the operating point for all subsequent 
experiments. The choice is justified along three axes. 
\textbf{(i) Matched-budget retrieval:} $M{=}5$ attains the 
highest $\mathrm{P@}26 = \mathrm{R@}26 = 0.6154$ across the 
sweep, outperforming $M{=}2$ ($0.5769$) by roughly four 
percentage points at the operating point that best summarizes 
retrieval quality. \textbf{(ii) Negligible loss elsewhere:} the 
mAP gap to $M{=}2$ (the global mAP maximum) is $0.0169$, well 
within one standard deviation of $M{=}5$ ($\sigma = 0.0327$), 
and $M{=}5$ ties for the best $\mathrm{P@}50$ and 
$\mathrm{R@}50$. \textbf{(iii) Gradient sample budget:} with 
$N{=}100$ instances, $M{=}5$ produces $N \cdot M = 500$ 
per-paraphrase gradient vectors, $2.5{\times}$ the supply 
available at $M{=}2$. This larger sample budget directly 
benefits the downstream block-wise SVD 
(Eq.~\ref{eq:rho}), where the effective rank 
$\rho_k$ and the quality of the recovered semantic subspace 
$U_{\mathcal{B}_k}$ both improve with the number of independent 
gradient observations, and is essential for the cross-task 
subspace overlap analysis in Section~\ref{sec:cross-task} where 
sample-size limitations would otherwise compound across tasks. 
Settings $M{=}2, 3$ would yield comparable IOI retrieval 
quality but underprovision the spectral analysis; settings 
$M \geq 8$ trade retrieval quality for a sample budget that the 
block-wise SVD cannot meaningfully exploit (since 
$\mathrm{rank}(G_{\mathcal{B}_k}) \leq N$ bounds the usable 
spectral dimension regardless of $M$). $M{=}5$ thus sits at the 
Pareto frontier of retrieval accuracy and gradient sample 
budget.

\subsubsection{Thresholds in Unit-classification}
\label{app:threshold}


\begin{figure}[t]
  \centering
  \begin{subfigure}{0.32\linewidth}
    \includegraphics[width=\linewidth]{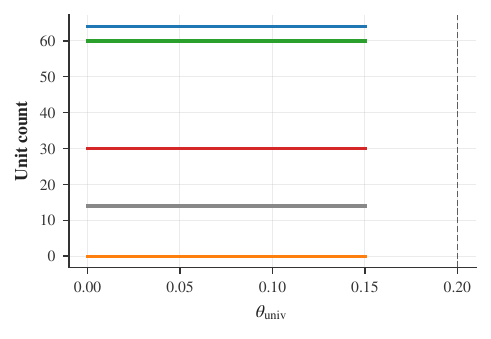}
    \caption{$\theta_{\mathrm{univ}}$}
  \end{subfigure}\hfill
  \begin{subfigure}{0.32\linewidth}
    \includegraphics[width=\linewidth]{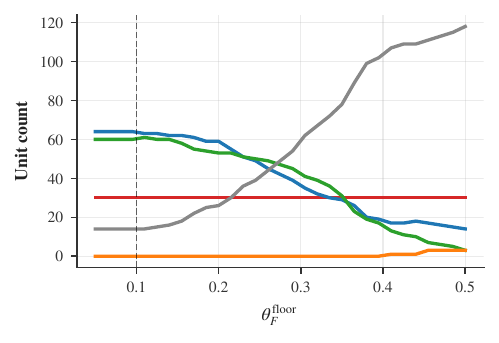}
    \caption{$\theta_F^{\mathrm{floor}}$}
  \end{subfigure}\hfill
  \begin{subfigure}{0.32\linewidth}
    \includegraphics[width=\linewidth]{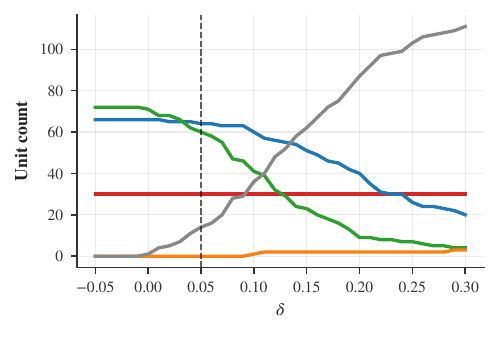}
    \caption{$\delta$}
  \end{subfigure}
  
  \hfill
  \begin{subfigure}{0.32\linewidth}
    \includegraphics[width=\linewidth]{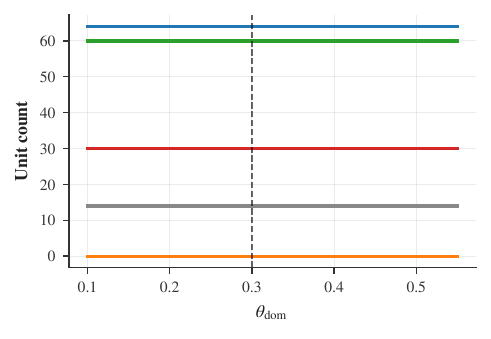}
    \caption{$\theta_{\mathrm{dom}}$}
  \end{subfigure}\hfill
  \begin{subfigure}{0.32\linewidth}
    \includegraphics[width=\linewidth]{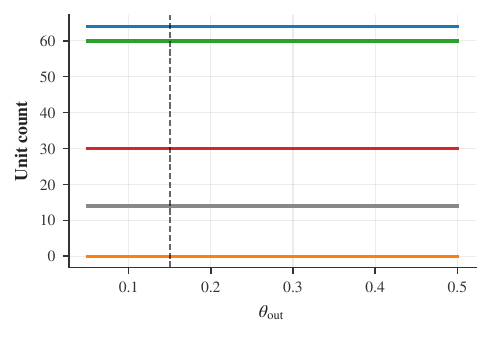}
    \caption{$\theta_{\mathrm{out}}$}
  \end{subfigure}
  \hfill\null
  
  \includegraphics[width=0.80\linewidth]{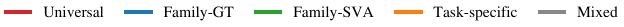}\\[-2pt]
  \caption{\textbf{One-dimensional sensitivity sweeps of the five
  classification thresholds.}
  Each panel sweeps a single threshold across its data-relevant range
  while holding the other four at their defaults; the dashed vertical
  line marks the default value used in the main results.
  Curves give the unit count assigned to each of the five classification
  outcomes (Universal, Family-GT, Family-SVA, Task-specific, Mixed)
  under the \emph{mean} criterion.
  All five thresholds yield piecewise-monotone, plateau-rich curves at
  their defaults, with $\theta_{\mathrm{dom}}$ and $\theta_{\mathrm{out}}$
  being essentially flat across the swept range.}
  \label{fig:A1}
\end{figure}

\begin{figure}[t]
  \centering
  \begin{subfigure}{0.24\linewidth}
    \includegraphics[width=\linewidth]{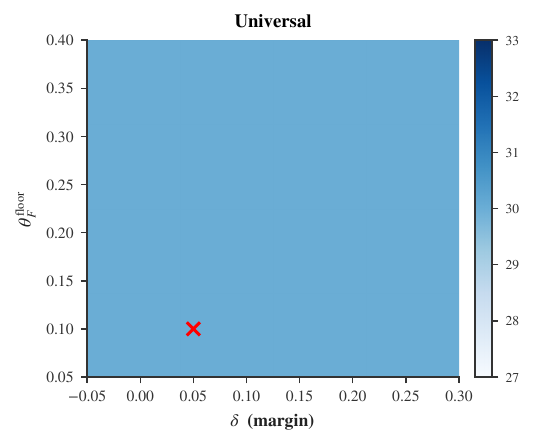}
    \caption{Universal}
  \end{subfigure}\hfill
  \begin{subfigure}{0.24\linewidth}
    \includegraphics[width=\linewidth]{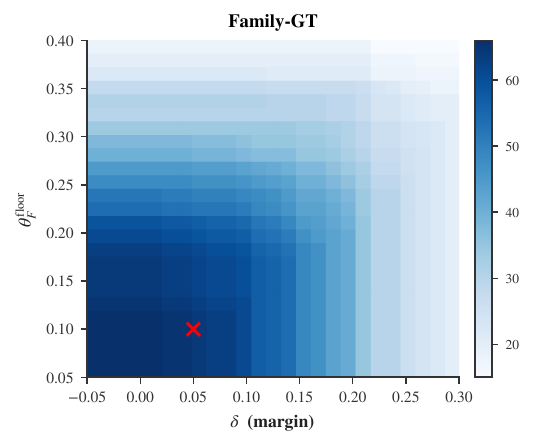}
    \caption{Family-GT}
  \end{subfigure}
  \begin{subfigure}{0.24\linewidth}
    \includegraphics[width=\linewidth]{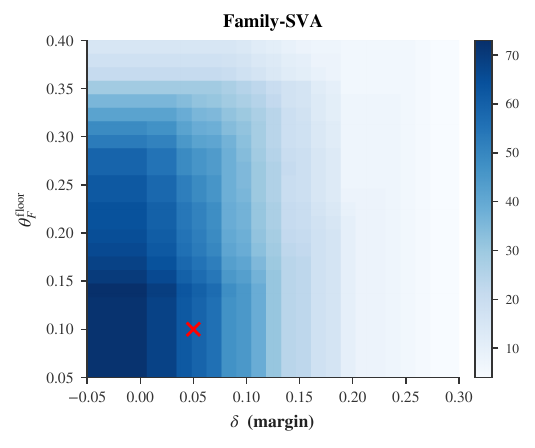}
    \caption{Family-SVA}
  \end{subfigure}\hfill
  \begin{subfigure}{0.24\linewidth}
    \includegraphics[width=\linewidth]{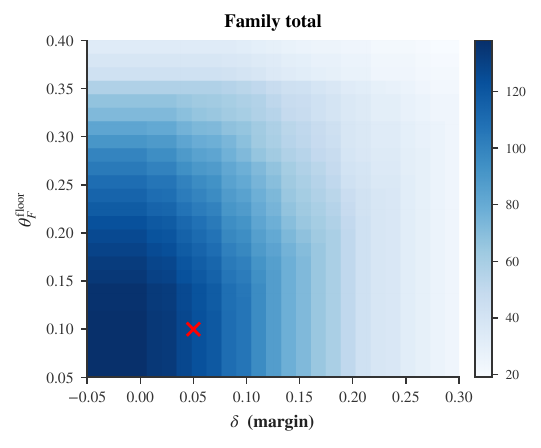}
    \caption{Family total}
  \end{subfigure}
  \caption{%
    \textbf{Joint sensitivity to the family-rule thresholds
    $\theta_F^{\mathrm{floor}}$ and $\delta$.}
    Each panel reports the unit count for one classification outcome
    over a $25\times25$ grid of the family within-floor and the
    margin parameter, with all other thresholds held at default. The
    red \texttt{$\times$} marks the default
    $(\theta_F^{\mathrm{floor}}, \delta) = (0.10, 0.05)$.
    Smooth diagonal contours and the absence of cliffs around the default
    indicate that the family classification is robust to small joint
    perturbations of these two thresholds; the default sits on a high
    plateau for Family-GT and a moderate plateau for Family-SVA.
    Criterion: \emph{mean}; importance: \texttt{consistency\_share}.
  }
  \label{fig:A2}
\end{figure}

\begin{figure}[t]
  \centering
  \begin{subfigure}{0.24\linewidth}
    \includegraphics[width=\linewidth]{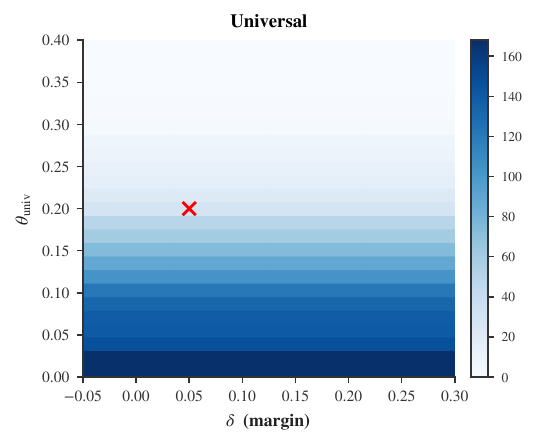}
    \caption{Universal}
  \end{subfigure}\hfill
  \begin{subfigure}{0.24\linewidth}
    \includegraphics[width=\linewidth]{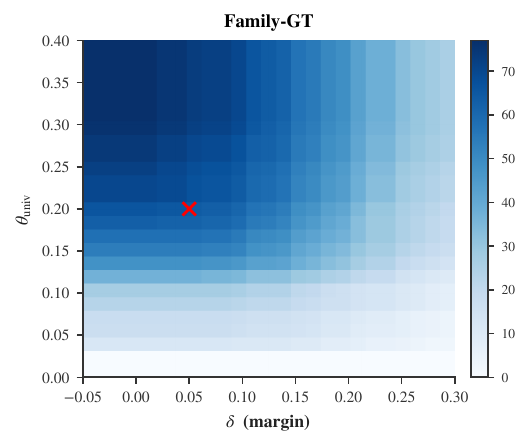}
    \caption{Family-GT}
  \end{subfigure}
  \begin{subfigure}{0.24\linewidth}
    \includegraphics[width=\linewidth]{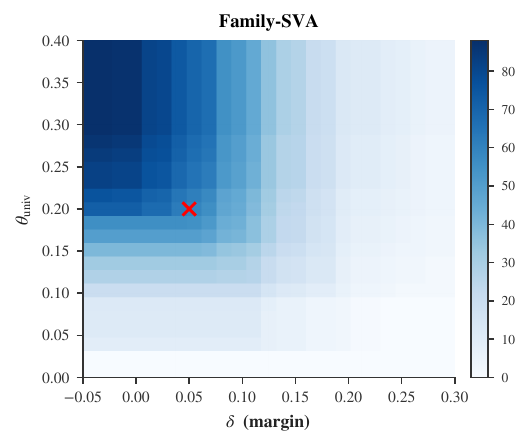}
    \caption{Family-SVA}
  \end{subfigure}\hfill
  \begin{subfigure}{0.24\linewidth}
    \includegraphics[width=\linewidth]{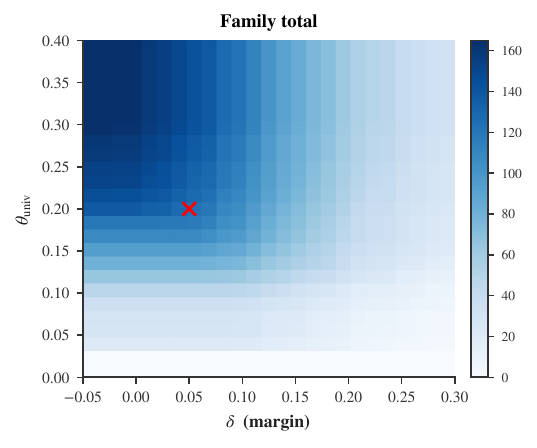}
    \caption{Family total}
  \end{subfigure}
  \caption{%
    \textbf{Cross-rule coupling between $\theta_{\mathrm{univ}}$ and
    $\delta$.}
    Joint sweep of the universal threshold (Rule 1) and the family
    margin (Rule 2) shows near-orthogonal dependence: the universal
    panel exhibits horizontal bands governed primarily by
    $\theta_{\mathrm{univ}}$, whereas the family panels exhibit
    vertical bands governed primarily by $\delta$. The two thresholds
    therefore operate near-independently across the cascade. Red
    \texttt{$\times$} marks the default
    $(\theta_{\mathrm{univ}}, \delta) = (0.20, 0.05)$.
    Criterion: \emph{mean}; importance: \texttt{consistency\_share}.
  }
  \label{fig:A3}
\end{figure}

\begin{table}[t]
\centering
\caption{Local sensitivity of the unit classification to $\pm 20\%$ 
perturbations of each threshold. Entries are the number of unit-level 
label flips relative to the default classification, out of $168$ units. 
Bold marks the only threshold with non-trivial sensitivity under the 
$\mathrm{mean}$ criterion adopted as reference.}
\vspace{0.2em}
\label{tab:A1}
\small
\begin{tabular}{l c c rr rr}
\toprule
& & & \multicolumn{2}{c}{criterion = mean} & \multicolumn{2}{c}{criterion = min} \\
\cmidrule(lr){4-5} \cmidrule(lr){6-7}
Threshold & Default & $\pm 20\%$ Range & $-20\%$ & $+20\%$ & $-20\%$ & $+20\%$ \\
\midrule
$\theta_{\mathrm{univ}}$       & 0.20 & [0.16, 0.24] & \textbf{36} & \textbf{16} & 9 & 4 \\
$\theta_F^{\mathrm{floor}}$    & 0.10 & [0.08, 0.12] & 0  & 1 & 2 & 5 \\
$\delta$                       & 0.05 & [0.04, 0.06] & 3  & 2 & 2 & 2 \\
$\theta_{\mathrm{dom}}$        & 0.30 & [0.24, 0.36] & 0  & 0 & 1 & 0 \\
$\theta_{\mathrm{out}}$        & 0.15 & [0.12, 0.18] & 0  & 0 & 0 & 0 \\
\bottomrule
\end{tabular}
\end{table}

\paragraph{Methodology.}
We assess the robustness of the unit-classification rule of Section~\ref{sec:method-cross-task}
to its five thresholds
$\{\theta_{\mathrm{univ}},\,\theta_F^{\mathrm{floor}},\,\delta,\,\theta_{\mathrm{dom}},\,\theta_{\mathrm{out}}\}$
through three complementary analyses. The study uses the
\texttt{consistency\_share} importance variant on the
$168$-unit $\times\,12$-task cross-task tensor in GPT-2 Small, with family layout
$\mathcal{F} = \{\textsc{GT}, \textsc{SVA}\}$ where
$\textsc{GT} = \{\texttt{gt},\texttt{gt\_price}\}$ and
$\textsc{SVA} = \{\texttt{sva},\texttt{sva\_binary},\texttt{sva\_blimp}\}$;
the remaining seven tasks contribute to $\mathrm{score}(u)$ and to
$\arg\max_t \widehat{I}^{\,t}(u)$ but are not eligible for family-specific labels.
All sweeps are evaluated under both the strict ($\min$) and lenient
($\mathrm{mean}$) within-pair aggregation criteria. 

First, a
\emph{one-dimensional sweep} varies each threshold across a $30$--$40$
point grid covering its data-relevant range 
($\theta_{\mathrm{univ}} \in [0.00,0.30]$,
$\theta_F^{\mathrm{floor}} \in [0.05,0.50]$,
$\delta \in [-0.05,0.30]$,
$\theta_{\mathrm{dom}} \in [0.10,0.55]$,
$\theta_{\mathrm{out}} \in [0.05,0.50]$)—while
holding the other four at their defaults
$(0.20,\,0.10,\,0.05,\,0.30,\,0.15)$. Second, two
\emph{two-dimensional sweeps} on $25 \times 25$ grids probe joint
dependencies: $\theta_F^{\mathrm{floor}} \times \delta$ characterises
the conjunctive admission region inside the family rule, and
$\theta_{\mathrm{univ}} \times \delta$ tests cross-rule coupling
between the universal and family stages of the cascade. Third, a
\emph{local perturbation analysis} applies $\pm 10\%$ and $\pm 20\%$
shifts to each threshold individually and records both per-category
count changes and the number of unit-level label flips relative to
the default classification.

\paragraph{Baseline classification and per-threshold sensitivity.}
At the default thresholds, the $\mathrm{mean}$ criterion produces a
balanced partition
(\textsc{Universal}: $30$, \textsc{Family-GT}: $64$,
\textsc{Family-SVA}: $60$, \textsc{Task-specific}: $0$,
\textsc{Mixed}: $14$), whereas the $\min$ criterion is dominated by
\textsc{Family-GT} ($97$) and starves \textsc{Family-SVA} ($8$), reflecting
that within-pair overlaps in \textsc{GT} are geometrically larger than
in \textsc{SVA} ($q_{75} = 0.36$ versus $0.29$); the $\mathrm{mean}$
partition is adopted as the reference. Under $\pm 20\%$ perturbation on 
the $\mathrm{mean}$ criterion, four of the five thresholds 
($\theta_F^{\mathrm{floor}}$, $\delta$, $\theta_{\mathrm{dom}}$, 
$\theta_{\mathrm{out}}$) shift at most $3$ of $168$ units (1.8\%); the 
universal threshold $\theta_{\mathrm{univ}}$ is the sole exception, with 
a $\pm 20\%$ perturbation flipping $36$ units (21\%) primarily between 
\textsc{Universal} and the family/mixed categories (Table~\ref{tab:A1}). 
The $\min$ criterion concentrates its residual sensitivity in 
$\theta_{\mathrm{univ}}$ ($9$ flips at $-20\%$) and 
$\theta_F^{\mathrm{floor}}$ ($5$ flips at $+20\%$), both at first-order 
universal/family boundaries. 
The remaining thresholds---$\delta$, $\theta_{\mathrm{dom}}$, and 
$\theta_{\mathrm{out}}$---act as essentially binary gates whose chosen 
values lie far from any unit's decision boundary; in particular, 
$\theta_{\mathrm{out}}$ is inactive on this dataset under both criteria, 
indicating that task-specific assignment is bottlenecked by 
$\theta_{\mathrm{dom}}$ and the stability of 
$\arg\max_t \widehat{I}^{\,t}(u)$ rather than by the outside-overlap cap.

\paragraph{Plateau structure and robust ranges.}
The one-dimensional curves (Fig.~\ref{fig:A1}) split the five thresholds 
into two regimes. Four of the five---$\theta_F^{\mathrm{floor}}$, $\delta$, $\theta_{\mathrm{dom}}$, 
and $\theta_{\mathrm{out}}$---contribute negligibly to classification 
sensitivity within the $\pm 20\%$ neighborhood of their defaults:
the $\theta_{\mathrm{dom}}$ and $\theta_{\mathrm{out}}$ 
panels are flat across the swept range, and $\pm 20\%$ perturbations of 
$\theta_F^{\mathrm{floor}}$ or $\delta$ shift at most $3$ of $168$ units 
(Table~\ref{tab:A1}). The universal threshold $\theta_{\mathrm{univ}}$ is 
the sole exception and exhibits genuine sensitivity: as it increases from 
$0$ to $0.30$, unit assignments smoothly migrate from \textsc{Universal} 
toward \textsc{Family-GT}, \textsc{Family-SVA}, and \textsc{Mixed}. 
Crucially, this migration is structured: \textsc{Family-GT} and 
\textsc{Family-SVA} each retain at least $57$ and $45$ units, respectively, 
across the $\pm 20\%$ range $[0.16, 0.24]$ around the default, and the 
relative ordering $\textsc{Family-GT} \geq \textsc{Family-SVA}$ holds 
throughout this neighborhood. The two-dimensional sweeps 
(Figs.~\ref{fig:A2},~\ref{fig:A3}) corroborate this picture. The 
$\theta_F^{\mathrm{floor}} \times \delta$ heatmap (Fig.~\ref{fig:A2}) 
exhibits a smooth diagonal ridge along which family counts decay 
gradually, with the default $(0.10, 0.05)$ on a high plateau for 
\textsc{Family-GT} and a moderate plateau for \textsc{Family-SVA}. The 
$\theta_{\mathrm{univ}} \times \delta$ heatmap (Fig.~\ref{fig:A3}) shows 
that universal counts vary primarily along the $\theta_{\mathrm{univ}}$ 
axis (horizontal bands) and family counts primarily along $\delta$ 
(vertical bands), so the two thresholds operate near-independently. The 
five-dimensional threshold space therefore decomposes into a 
family-identification subspace, on which the classification is highly 
stable, and a universal-vs-family axis along which $\theta_{\mathrm{univ}}$ 
smoothly modulates the universal count without disrupting the family 
structure. Since the central claim of this analysis concerns the family 
decomposition rather than the absolute size of the \textsc{Universal} set, 
the reported classification is robust in the relevant sense.

\section{Computational Resources.}
\label{app:comp}
We report the hardware, per-run wall-clock time, and the
asymptotic complexity of each stage of the pipeline so that the
resource footprint can be estimated for any choice of model, scope,
or dataset size. Throughout, adopted from main text, let $N$ denote the number of instances,
$M$ the number of paraphrases per instance, $D$ the number of
parameters in the analysed scope (e.g.\ all attention output
projections, or attention plus MLP), $K$ the number of discovered
parameter blocks, $d_k$ the size of block $k$ (with
$\sum_{k} d_{k} = D$), $d_u$ the size of a single attention or MLP
unit, and $\rho_{B} \le N$ the effective rank of block $B$.

\begin{itemize}
  \item \textbf{Hardware.} A single workstation node with six NVIDIA
    RTX~A6000 GPUs (48~GB VRAM each), two AMD EPYC~7543 CPUs
    (64 physical cores in total), and 503~GiB of system memory.
    Each run uses a single GPU; the remaining GPUs are used for
    embarrassingly parallel sweeps over tasks.

  \item \textbf{Device split.} Only the per-instance forward and
  backward passes used to extract gradients run on the GPU; all
  downstream linear algebra (gradient accumulation, Gram
  construction, block discovery, and SVD) runs on the host CPU.
  This split is dictated by memory rather than compute: the
  averaged gradient buffer alone reaches tens of gigabytes, exceeding single-GPU memory; and for the small
  instance counts considered here ($N \le 100$), the Gram and SVD
  stages are bandwidth-bound, so host execution incurs no
  meaningful slowdown relative to a GPU implementation.

    \item \textbf{Headline configurations and wall-clock time.}
    We report two settings, both with $N{=}100$, $M{=}5$, sequence
    length 256, and scope = attention output projections plus MLP
    weights:
    \begin{itemize}
      \item \emph{GPT-2~Small} ($D \approx 8.4 \times 10^{7}$).
        End-to-end run: 1.5--3~minutes; peak GPU memory 1--2~GB;
        peak resident host memory $\sim$50~GB, plus $\sim$80~GB of
        memory-mapped scratch on \texttt{tmpfs}.
      \item \emph{Pythia-160M} ($D \approx 1.3 \times 10^{8}$).
        End-to-end run: 2.5--5~minutes; peak GPU memory 2--3~GB;
        peak resident host memory $\sim$75~GB, plus $\sim$130~GB of
        memory-mapped scratch on \texttt{tmpfs}.
    \end{itemize}

  \item \textbf{Gradient collection dominates wall-clock time.}
    Cost is $\mathcal{O}(NM \cdot T_{\mathrm{fb}})$
    forward--backward passes, where $T_{\mathrm{fb}}$ depends on
    model size and sequence length ($\sim$30--50~ms per pass for
    GPT-2~Small at sequence length 256 in \texttt{bf16} on an
    A6000). Persistent CPU memory is dominated by three buffers:
    a per-instance averaged gradient buffer ($4ND$ bytes,
    \texttt{fp32}); a per-instance sign-coherence buffer
    ($ND$ bytes, \texttt{int8}); and the full normalized gradient
    matrix ($2NMD$ bytes, \texttt{fp16}, transparently backed by a
    memory-mapped file on \texttt{tmpfs} when it exceeds 4~GB).
    The dense Gram matrix over paraphrases adds $4(NM)^{2}$ bytes.
    For the headline configuration ($D \approx 8.4 \times 10^{7}$),
    these account for $\sim$34~GB, $\sim$8~GB, $\sim$84~GB
    (memory-mapped), and $\sim$1~MB respectively, all comfortably
    within the host's 503~GiB of RAM.

  \item \textbf{Downstream stages are substantially cheaper.}
    Diagnostics (Noise-to-Signal Ratio and Sign Coherence) cost
    $\mathcal{O}(N^{2} M^{2} + ND)$ on CPU and complete in seconds.
    Block discovery via kernel alignment costs
    $\mathcal{O}(n_{\mathrm{units}}^{2} N^{2} + N^{2} D)$ with peak
    memory $4 \, n_{\mathrm{units}} N^{2}$ bytes for the stacked
    kernel tensor. Block-wise SVD via the Gram trick costs
    $\mathcal{O}(D N^{2} + K N^{3})$ along the fast path that
    reuses the cached averaged gradients. Per-unit weighted SVD
    costs $\mathcal{O}\!\left(\sum_{u} d_{u} \rho_{B}^{2}\right)$
    and is negligible in both time and memory.

  \item \textbf{Scaling to larger models.} Host memory for the
    averaged gradient buffer grows linearly in $D$, and becomes the
    binding constraint well before GPU memory does. For
    Pythia-1B-class models ($D \sim 9 \times 10^{8}$) the buffer
    alone would require $\sim$360~GB and must be disabled; in that
    regime the SVD stage falls back to a slow path that recomputes
    $KMN$ additional backward passes, trading wall-clock time for
    host memory.

  \item \textbf{Knobs for reducing the resource footprint.}
    (i)~Narrowing the analysed scope (e.g.\ restricting to
    attention output projections excludes MLP parameters and
    reduces $D$ by roughly an order of magnitude on GPT-2~Small);
    (ii)~chunking the Gram accumulation to trade peak host memory
    against per-step overhead without changing the asymptotic
    cost.
\end{itemize}

\begin{table}[h]
\centering
\begin{threeparttable}
\caption{Per-stage compute budget for the two headline configurations.
Wall-clock figures are measured
on a single NVIDIA RTX~A6000.}
\vspace{0.2em}
\label{tab:compute-budget}
\begin{tabular}{lllrr}
\toprule
Stage & Asymptotic cost & Device \\
\midrule
Gradient collection
  & $\mathcal{O}(NM \cdot T_{\mathrm{fb}})$
  & GPU + CPU\\
Gram assembly
  & $\mathcal{O}(N M^{2} D)$
  & CPU\\
Diagnostics
  & $\mathcal{O}(N^{2} M^{2} + N D)$
  & CPU\\
Block discovery
  & $\mathcal{O}(n_{\mathrm{units}}^{2} N^{2} + N^{2} D)$
  & CPU\\
Block-wise SVD (fast path)
  & $\mathcal{O}(D N^{2} + K N^{3})$
  & CPU\\
Per-unit weighted SVD
  & $\mathcal{O}\!\left(\sum_{u} d_{u} \rho_{B}^{2}\right)$
  & CPU\\
\midrule
\end{tabular}
\end{threeparttable}
\end{table}

\end{document}